\documentclass{article}

\usepackage{arxiv}

\newif\ifdraft
\newif\ifplots
\plotstrue

\usepackage{url}  % simple URL typesetting
\usepackage{amsmath}
\usepackage{amsfonts}
\usepackage{multirow}
\usepackage{rotating}
\usepackage[table]{xcolor}
\usepackage[normalem]{ulem}
\usepackage{nicefrac}
\usepackage[utf8]{inputenc}
  
\newcommand{\centercell}[1]{\multicolumn{1}{c|}{\textbf{#1}}}

\newcommand{\fsebcomment}[1]{\ifdraft{\leavevmode\color{red}{[FS]: {#1}}}\else{\vspace{0ex}}\fi}

\newcommand{\plot}[1]{\ifplots{{#1}}\else{\vspace{0ex}}\fi}

\newcommand{\fasttext}{\texttt{fastText}}
\newcommand{\wordvec}{\texttt{word2vec}}
\newcommand{\glove}{\texttt{GloVe}}
\newcommand{\glovest}{\texttt{GloVe(static)}}
\newcommand{\glovetr}{\texttt{GloVe(trainable)}}
\newcommand{\glovern}{\texttt{GloVe+Random}}
\newcommand{\glovewce}{\texttt{GloVe+WCEs}}
\newcommand{\glovewcest}{\texttt{GloVe+WCEs(static)}}
\newcommand{\glovewcetr}{\texttt{GloVe+WCEs(trainable)}}
\newcommand{\random}{\texttt{Random}}
\newcommand{\svmtfidf}{\texttt{SVM-tfidf}}
\newcommand{\svmglove}{\texttt{SVM-GloVe}}
\newcommand{\svmglovewce}{\texttt{SVM-GloVe+WCEs}}

\newcommand{\side}[1]{\begin{sideways}{#1}\end{sideways}}

\newcommand{\killpunct}[1]{}

\begin{document}

\title{Word-Class Embeddings \\for Multiclass Text Classification}
% \title{Word-Class Embeddings}

\author{
  Alejandro Moreo, Andrea Esuli, Fabrizio Sebastiani \\
  Istituto di Scienza e Tecnologie dell'Informazione \\
  Consiglio Nazionale delle Ricerche \\
  56124 Pisa, Italy\\
  \texttt{\{firstname.lastname\}@isti.cnr.it} \\
  %% \AND
  %% Coauthor \\
  %% Affiliation \\
  %% Address \\
  %% \texttt{email} \\
  %% \And
  %% Coauthor \\
  %% Affiliation \\
  %% Address \\
  %% \texttt{email} \\
  %% \And
  %% Coauthor \\
  %% Affiliation \\
  %% Address \\
  %% \texttt{email} \\
}

%\editor{}
% For research notes, remove the comment character in the line below.
% \researchnote

\maketitle

% --------------------------------------------------

\begin{abstract}
  \noindent Pre-trained word embeddings encode general word semantics
  and lexical regularities of natural language, and have proven useful
  across many NLP tasks, including word sense disambiguation, machine
  translation, and sentiment analysis, to name a few. In supervised
  tasks such as multiclass text classification (the focus of this
  article) it seems appealing to enhance word representations with
  ad-hoc embeddings that encode task-specific information. We propose
  (supervised) \emph{word-class embeddings} (WCEs), and show that,
  when concatenated to (unsupervised) pre-trained word embeddings,
  they substantially facilitate the training of deep-learning models
  in multiclass classification by topic. We show empirical evidence
  that WCEs yield a consistent improvement in multiclass
  classification accuracy, using four popular neural architectures and
  six widely used and publicly available datasets for multiclass text
  classification. Our code that implements WCEs is publicly available
  at \url{https://github.com/AlexMoreo/word-class-embeddings}.
\end{abstract}

% --------------------------------------------------

\keywords{
  Word-Class Embeddings, Word embeddings, Distributional hypothesis,
  Multiclass text classification
}

% --------------------------------------------------

\section{Introduction}
\label{sec:intro}

\noindent Recent advances in deep learning have led to important
improvements in many NLP tasks that deal with the semantic analysis of
text, including word sense disambiguation, machine translation,
summarization, question answering, and sentiment analysis
(see~\cite{collobert2011natural,lecun2015deep}, for an
overview). %semantic parsing? named entities recognition?
At the heart of the neural approach to the semantics of text lies the
concept of \emph{word embedding} (a.k.a.\ \emph{continuous} or
\emph{distributed representation}
--~\cite{bengio2003neural,mikolov2013distributed}), a dense
representation of a word's meaning in a vector space where the
semantic similarity of words %of terms
is embodied in the notion of distance between vectors. %between points

Word embeddings can either be initialized randomly and allowed to
evolve %as the embedding layer
along the rest of the model parameters, or be initialized from
pre-trained word embeddings obtained offline by scanning massive
amounts of textual data.  This latter approach is generally preferred,
since pre-trained embeddings encode an effective prior
% (that could eventually be fine-tunned for the task at hand)
% ~\cite{erhan2010does}
that embodies our general-purpose knowledge of the semantics of words,
and that can be successfully transferred to (and eventually fine-tuned
for) specific application contexts and downstream
tasks~\cite{erhan2010does}.

Approaches to generate word embeddings typically rely on the
\emph{distributional hypothesis}, according to which words that tend
to occur in similar contexts tend to have similar
meanings~\cite{harris1954distributional}.
% this is achieved by analyzing words context in text
Different realizations of this hypothesis were initially based on
\emph{context-counting}
approaches~\cite{blei2003latent,bullinaria2007extracting,deerwester1990indexing,sahlgren2005introduction} %those are references to LSA, LDA, PPMI, and RI
and later based on \emph{context-predicting}
approaches~\cite{joulin2016bag,mikolov2013distributed,pennington2014glove}.
% those are references to \wordvec, \glove, and \fasttext\.
Context-counting approaches collect frequencies of word co-occurrence
and typically involve some form of matrix factorization to obtain the
final word representations~\cite{deerwester1990indexing}.
Conversely, in context-predicting approaches the word representations
constitute the parameters of a model trained to predict some
distributional property of the data.  As an example, \wordvec 's
\emph{skip-gram with negative sampling} method (SGNS
--~\cite{mikolov2013distributed}) tries to guess the surrounding
words from the observation of the central word in a sliding context
window.

% (e.g., guessing the central word from the observation of the
% surrounding words in a sliding context window
% ~\cite{mikolov2013distributed}, or predicting the co-occurrence
% counts of terms~\cite{pennington2014glove}).
While the relative desirability of one paradigm over the other was
once the subject of debate~\cite{baroni2014don}, it has later been
argued that the two approaches simply embody different ways of
pursuing what is essentially the same
objective~\cite{levy2014neural}, %li2015word},
and that differences in performance are mainly explainable in terms of
hyperparameter settings and design choices~\cite{levy2015improving}.
It has been proven that the optimum of the objective function that
SGNS (a context-predicting method) seeks to optimize can directly be
attained by a context-counting method called \emph{shifted positive
pointwise mutual information} (SPPMI --~\cite{levy2014neural}).
% \fsebcomment{Perché non SPPMI? Nella sezione che segue, per PPMI si
% intende la (non-shifted?) positive pointwise mutual information.}
This seems to suggest that SPPMI (and context-counting approaches in
general) should be preferred to SGNS (and to context-predicting
approaches in general).  However, there are practical reasons why the
opposite is the case.  The main drawback of context-counting
methods %with respect to context-predicting ones
is the fact that they need to work with the entire co-occurrence
matrix, something that becomes impractical when large quantities of
text are involved.  This problem does not harm neural supervised
learning approaches, though, which are inherently incremental when
adopting stochastic optimization (a standard practice nowadays).  For
this reason, the neural approach is currently the dominant one in
modern distributional semantics.
% \fsebcomment{Tutta la discussione di questo paragrafo sembra
% scorrelata da quello che si dice nel paragrafo che segue ...}

% ``Embedding vectors pre-trained on large corpora can serve as the
% embedding layer in neural network models and may significantly
% benefit downstream language tasks because reliable external
% information is transferred from large corpora to those specific
% tasks.''(Enhancing Domain Word Embedding via Latent Semantic
% Imputation)

% ----------------------------------------------------

\subsection{Word-Class Embeddings}
\label{sec:WCEs}

\noindent Through the lens of the downstream task, the pre-trained
word embeddings that all these methods generate are unsupervised,
% \blue{(sometimes called \emph{self-supervised})},
in the sense that they capture how terms are distributed in general
language use, in a way which is completely independent of (and thus
not optimized for) the downstream task.  However, in supervised tasks
such as text classification (the focus of this work), it seems
reasonable to imbue the word representations with supervised
information that is available during
training. %, i.e., by taking into account the class-conditional term distributions.
% The general idea that we explore in this article is that in
% applications of text classification it should be possible to improve
% word representation by making explicit usage of supervised
% information (as opposed to general knowledge).
In this article we propose \emph{word-class embeddings} (WCEs), a form
of supervised embeddings of words specifically designed for multiclass
text classification, that directly model the interactions between
terms and class labels. %in a dedicated dense vector space.
% \alexcomment{Ricordo che avevamo deciso di giustificare perchè lo
% chiamiamo word-class anziché word-label...} \fsebcomment{Secondo me
% non è necessario.}

% When word embeddings are used in a supervised learning process they
% can be considered to be parameter of the learning model, thus
% allowing the learning process to implicitly model in the embeddings
% the relations between input words and the output labels.

% In this article, we propose word-class embeddings, a form of
% supervised embeddings that directly model the interactions between
% terms and labels in a dedicated dense vector space.
A related intuition has been explored before in the context of text
classification~\cite{bojanowski2017enriching,joulin2016bag,tang2015pte,wang2018joint}
by jointly modelling word embeddings and \emph{label embeddings} in a
common vector space as part of the optimization procedure.
% \fsebcomment{Forse bisognerebbe accennare a cosa sono i label
% embeddings.}
Arguably, the best-known among the methods based on this intuition is
\fasttext~\cite{bojanowski2017enriching,joulin2016bag}, a variant of
\wordvec 's \emph{continuous bag-of-words} method (CBOW -- see Section
\ref{sec:relwork-word}) that substitutes the target central word that
CBOW seeks to predict, with a token representing one of the document's
labels.  The result is a method that jointly models
words\footnote{\fasttext\ can consider not only unigrams but also
n-grams and subwords as the surface forms of input.} and labels as
vectors, by recasting labels as new terms and simply applying the
distributional hypothesis anew.
% \alexcomment{Ho spostato la footnote accanto a \emph{words}} that
% embeds words\ and class labels in the same vector space.  To do so,
% \fasttext\ relies on the skip-gram with negative-sampling (SGNS)
% model by substituting the role of the target central word (within a
% sliding window) that SGNS seeks predict with a token representing
% one of the document's label.

We follow a different approach from those explored before by confining
the supervised embeddings in a dedicated vector space, so that they
can then be concatenated with any unsupervised pre-trained
representations.  The resulting embedding matrix can be used as the
building block of any neural architecture.  Our method does not
involve any optimization procedure but operates directly on the
co-occurrence counters.  In a way, the method we propose might be
regarded as the context-counting counterpart of (the
context-predicting) \fasttext\ for word-class distributions, just like
SPPMI stands to SGNS for word-word
distributions~\cite{levy2014neural}.  Note that the disadvantage of
context-counting approaches with respect to context-predicting ones
that we have discussed before (i.e., the need to work with the entire,
potentially huge co-occurrence matrix) does not arise here, since the
amount of labelled documents one typically has in text classification
applications is limited, and working with the co-occurrence matrix is
thus unproblematic. %doable?
% The underlying motivation is directly inspired by the distributional
% hypothesis: words with similar label distributions tend to play
% similar roles in classification.

We show empirically that extending the pre-trained unsupervised word
embeddings with our task-specific supervised WCEs substantially
facilitates the training of neural classifiers, and yields consistent
improvements in multiclass classification performance across six
widely used and publicly available text classification datasets, and
four popular neural architectures (including \fasttext).  Experiments
also show that our word-class embeddings can be computed very quickly.

The rest of this article is structured as follows.  In Section
\ref{sec:relwork} we thoroughly review related work.  We explain the
method in Section~\ref{sec:method}, while Section~\ref{sec:exp}
reports the experimental evaluation we have
conducted. Section~\ref{sec:discussion} tackles a few advanced topics
related to WCEs, while Section~\ref{sec:conclusions} concludes,
pointing at possible avenues for future work.

% Word-label embeddings stand to Word embeddings like Supervised Term
% Weighting stands to Term Weighting.  Link to lifelong learning
% (general) vs task specific learning.  Link to gradient-vanish and
% justify the inclusion of a supervised presure from the bottom.
% Sdrop as a way to counter overfitting (specificty vs generalization)

% Possibly point out that: Word-label embeddings stand to \fasttext\
% like PCA(PPMI) stands to \wordvec, i.e., is the ``counting''
% (doc-by-term matrix) version of the ``predicting'' neural (batched)
% approach~\cite{baroni2014don}.  (Andrea suggested that we could
% attempt to prove it).

% ``Embedding vectors pre-trained on large corpora can serve as the
% embedding layer in neural network models and may significantly
% benefit downstream language tasks because reliable external
% information is transferred from large corpora to those specific
% tasks.''(Enhancing Domain Word Embedding via Latent Semantic
% Imputation)

% ----------------------------------------------------

\section{Related Work}
\label{sec:relwork}

\noindent In this section we turn to review relevant related work on
word embeddings (Section~\ref{sec:relwork-word}), label embeddings
(Section~\ref{sec:relwork-label}), and neural approaches to text
classification that exploit either word or label embeddings (Section
\ref{sec:relwork-tc}).

% ----------------------------------------------------

\subsection{Word Embeddings}
\label{sec:relwork-word}

% Likely, the concept of word embedding was first sketched in
% ~\cite{rumelhart1988learning} and the first neural approach dates
% back to~\cite{bengio2003neural} \alexcomment{[check!]}

\noindent Although the term \emph{word embedding} owns its popularity
to the neural approach, the very first attempts to generate
distributed representations arose in the realm of context-counting
approaches.  Arguably, the best-known one is \emph{Latent Semantic
Analysis} (LSA --~\cite{deerwester1990indexing}), a method that
obtains $r$-dimensional representations of words by factoring (via
singular value decomposition -- SVD) a term-by-context co-occurrence
matrix, and retaining the $r$ eigenvectors with the highest
eigenvalue. \emph{Positive Pointwise Mutual Information} (PPMI
--~\cite{levy2014neural}) takes the positive part of PMI as applied
to the counters of the matrix, before decomposing
it.\footnote{\label{footnote:PMI}\emph{Pointwise Mutual Information}
(PMI) is defined as
$\mathrm{PMI}(t_{i},c_{j})=\log\frac{\Pr(t_{i},c_{j})}{\Pr(t_{i})\Pr(c_{j})}$,
where $\Pr(t_{i},c_{j})$ is the joint probability of term $t_{i}$ and
context $c_{j}$, and $\Pr(t_{i})$ and $\Pr(c_{j})$ are the marginal
probabilities of the term and context, respectively. PPMI takes the
positive part of PMI, i.e.,
$\mathrm{PPMI}(t_{i},c_{j})=\max\{0,\mathrm{PMI}(t_{i},c_{j})\}$.}  We
explore PPMI as an alternative to our method in Section~\ref{sec:tsr}.

\emph{Latent Dirichlet Allocation} (LDA --~\cite{blei2003latent})
is a generative statistical model that assumes each document to be a
mixture of latent topics.  Although LDA ends up building a matrix that
models the strength of the association of words with topics (something
very similar in spirit to our goal), it does not use supervision;
instead, our WCEs are supervised (i.e., they use the class label
information along with the co-occurrence matrix), and require no
optimization.

Most context-counting approaches suffer from the fact that the
co-occurrence matrix has to be explicitly allocated in memory.  In an
attempt to overcome this limitation, \emph{Random Indexing} (RI
--~\cite{kanerva2000random,sahlgren2005introduction}) iteratively
constructs an approximation of the co-occurrence matrix by
accumulating nearly-orthogonal random indexes for terms and getting
rid of the factorization. However, RI is generally outperformed by
neural methods such as \wordvec\ and \glove\ (explained below) when
large quantities of text are available for
training~\cite{sahlgren2016effects}.

% A. L. Maas, R. E. Daly, P. T. Pham, D. Huang, A. Y. Ng, and
% C. Potts. Learning word vectors for sentiment analysis.  In ACL-HLT,
% pages 142–150, 2011.  "Latent Dirichlet Allocation (LDA; (Blei et
% al., 2003)) is a probabilistic document model that assumes each
% document is a mixture of latent topics. For each latent topic T, the
% model learns a conditional distribution p(w|T) for the probability
% that word w occurs in T. One can obtain a kdimensional vector
% representation of words by first training a k-topic model and then
% filling the matrix with the p(w|T) values (normalized to unit
% length).  The result is a word–topic matrix in which the rows are
% taken to represent word meanings. However, because the emphasis in
% LDA is on modelling topics, not word meanings, there is no guarantee
% that the row (word) vectors are sensible as points in a
% k-dimensional space. Indeed, we show in section 4 that using LDA in
% this way does not deliver robust word vectors. The semantic
% component of our model shares its probabilistic foundation with LDA,
% but is factored in a manner designed to discover word vectors rather
% than latent topics. Some recent work introduces extensions of LDA to
% capture sentiment in addition to topical information (Li et al.,
% 2010; Lin and He, 2009; Boyd-Graber and Resnik, 2010)."  Equation 1
% resembles the probabilistic model of LDA (Blei et al., 2003), which
% models documents as mixtures of latent topics. One could view the
% entries of a word vector φ as that word’s association strength with
% respect to each latent topic dimension.

The neural approach to distributional semantics started
with~\cite{bengio2003neural}, and gathered momentum with
\wordvec~\cite{mikolov2013distributed}, a method based on a two-layer
neural network trained to predict the words in the context of a
central word (\emph{skip-gram} -- SG) or the center word from the
words in a (sliding) context window (\emph{continuous bag-of-words} --
CBOW). Input and output terms are represented as one-hot vectors, and
the first layer acts as a lookup table indexing the word embeddings
(the layer parameters). \wordvec\ owns part of its success to
hierarchical softmax and negative
sampling~\cite{mikolov2013efficient}, that permitted to dramatically
speed up their computation, thus allowing the method to scale to
massive amounts of textual data.  \glove\ is another popular method
for generating embeddings, that has proven superior to \wordvec\ in
various tasks~\cite{pennington2014glove}.  \glove\ learns the word
vectors that better reconstruct the probabilities of co-occurrence
between pairs of terms as estimated via their dot product.  Both
\wordvec\ and \glove\ have been used to generate large sets of
embeddings that have later been made publicly available.  We use both
sets of pre-trained vectors in the experiments of Section
\ref{sec:exp}.

\subsection{Label Embeddings}
\label{sec:relwork-label}

% \alexcomment{short intro of the idea}

\noindent A primitive form of label embedding is to be found in
\emph{distributed output codes} for single-label multiclass
classification problems.\footnote{Given a set of classes
$\mathcal{C}=\{c_{1},\ldots,c_{m}\}$, a classification problem is said
to be \emph{multiclass} if $m>2$; it is said to be \emph{single-label}
if each item always belongs to exactly one class; it is said to be
\emph{multilabel} if each item can belong to any number (i.e., 0, 1,
or more than 1) of classes in $\mathcal{C}$.}  The idea is to assign a
different (binary) ``codeword'' of length $m'$ to each of the $m$
classes, with $m'<m$, and then learn $m'$ binary predictors (one for
each binary bit in the codeframe), instead of $m$ predictors (one for
each class).  The actual class to be returned is the class with the
closest codeword (in terms of Hamming distance) to the predicted
binary string.  These distributed codes were said to be ``meaningful''
if each bit was meant to encode one particular aspect of the
label~\cite{sejnowski1987parallel}, e.g., when guessing the correct
digit represented in an image of a handwritten digit, the meaning of
one bit in the codeframe could be ``the image contains a horizontal
line'', while the meaning of another could be ``the image contains a
closed curve''.
% \alexcomment{label attributes}
\cite{dietterich1994solving} obtained improved results by redefining
codewords as \emph{error-correcting codes}, a type of distributed
output codes with constraints.  Error-correcting codes allow the
classifier to recover the correct label in spite of some binary
misclassifications, though at the expense of involving more binary
learners and sacrificing the meaning of each of them.  Note that,
although these early attempts reflect some semantic structure of the
labels, they do so to a very limited extent.

Label embeddings have later proven useful in large sparse multilabel
classification (sometimes called \emph{extreme multilabel
classification}), i.e., in classification scenarios characterized by
large sets of classes and where each item typically belongs to only a
few
classes~\cite{bengio2010label,bhatia2015sparse,hsu2009multi,lin2014multi,weston2010large,yu2014large}.
The general idea is to reduce the high-dimensional and sparse label
matrix by means of an \emph{encoding function}, so that a learner can
be then trained to predict low-dimensional label vectors, that are
ultimately mapped back to the original sparse label space via a
\emph{decoding} function.  In this context,~\cite{hsu2009multi} used
compressed sensing as the encoding function, and applied sparse
reconstruction techniques for the decoding function.  The encoding
function was defined as a random projection, and was thus not learned
from data. ~\cite{bengio2010label} and~\cite{weston2010large}
proposed instead to learn the parameters of the projection matrix,
while other approaches, such as the one of~\cite{lin2014multi},
assumed the encoding function to be implicit, and directly attempted
to learn a code matrix and a decoding function.  Other approaches,
such as that of~\cite{yu2014large}, frame the problem as a (linear)
low-rank approximation, or as a (non-linear) manifold learning of the
sparse label matrix~\cite{bhatia2015sparse}.  The ultimate goal that
all these works pursue is to bypass the cost of training a classifier
for each binary label.
% \alexcomment{This could be trimmed}

A considerable body of research on label embeddings has then emerged
which was aimed at dealing with the difficulty of obtaining labeled
data for specific domains, especially in the realm of ``zero-shot''
classification of
images~\cite{akata2015label,akata2015evaluation,norouzi2013zero}.
% ConvSE~\cite{norouzi2013zero} applies SG to Wikipedia in order to
% model the label embeddings as the word embedding for the term naming
% each of the $m$ classes. A convolutional classifier pre-trained on
% ImageNet's classes (with $m'<<m$ classes), is used to obtain, for an
% input image, the vector of posterior probabilities for each of the
% original $m'$ classes. Finally, it returns the label whose embedding
% better matches the convex linear combination of the label embeddings
% of the ImageNet's classes, as weighted by the posterior
% probabilities.
For instance, ConvSE~\cite{norouzi2013zero} maps image embeddings, as
produced by a convolutional classifier trained on ImageNet, to the
word embedding space representing the class labels (as produced by
SGNS trained on Wikipedia).~\cite{akata2015evaluation} propose
\emph{Structured Joint Embeddings}, investigating different mechanisms
to produce label embeddings from supervised attributes (hand-coded
vectors accounting for aspects describing the label -- much like the
``meaningful'' distributed output codes mentioned above), hierarchical
dependencies between labels (as mined from ontologies such as
WordNet), and unsupervised sources of text data (such as
Wikipedia). Labels are embedded into complementary vector spaces, and
a compatibility function, within the structured support vector machine
(SSVM) framework, is then trained to maximize the matching with
embedded images from the same class.

For the unsupervised representation of labels, bag-of-words and other
distributional semantic models (\glove\ and \wordvec) have been
investigated.~\cite{rodriguez2013label} tackle the problem of text
recognition, and similarly rely on SSVM. The goal is to identify the
correct word from an image containing it. Words are thus taken as
labels, and these labels are embedded in a vector space governed by
lexical (instead of semantic) similarity.\footnote{While
\cite{rodriguez2013label} dub ``word-label embeddings'' their word
representations, this is just to emphasize that words are the target
variables, which indicates that the meaning they attribute to the term
``word-label embedding'' is very different from our notion of WCE.}

% \alexcomment{This could be trimmed}

% ----------------------------------------------------

\subsection{Neural Text Classification}
\label{sec:relwork-tc}

\noindent Text classification (TC) is a supervised learning task in
which a model is trained to predict labels for unseen documents from
the observation of labelled documents.  Unlike traditional machine
learning approaches to TC which represented documents thorough sparse
vectors of lexical
features~\cite{joachims1998text,wang2012baselines}, the neural
approach to TC builds on top of distributed representations for words
and documents. %(embeddings).
% Word embeddings might either be pre-trained on external corpora
% (unsupervised), learnt for the classification task (supervised), or
% a mixture of both (initialized as pre-trained vectors that are then
% fine-tuned for the task).
Popular architectures routinely adopted in neural TC include
convolutional neural networks (CNNs
--~\cite{collobert2011natural,kim2014convolutional,le2018convolutional}),
recurrent neural networks (RNNs
--~\cite{hochreiter1997long,lai2015recurrent,rumelhart1986learning}),
recursive deep models (RDMs --~\cite{socher2013recursive}), and
attention models (ATTNs
--~\cite{luong2015effective,vaswani2017attention}). %,ELMo,BERT,XLNet}).
The training strategy is common across all these architectures: they
first generate a document representation
% (whichever the degree of sophistication)
and they then connect it directly to the labels during training; words
in documents are only connected to labels indirectly.

Different models have been proposed that instead leverage the training
labels directly at the word
level~\cite{bojanowski2017enriching,joulin2016bag,tang2015pte,wang2018joint}.
\cite{tang2015pte} proposed \emph{Predictive Text Embeddings}
(PTEs), %as a way of combining the information which is available in large amounts of unlabeled examples and in (typically) limited amounts of labeled examples to devise word representations that have strong predictive power for classification.
% PTE is designed as
a type of embeddings that rely on a heterogeneous text network
consisting of three bipartite graphs, each of which models one
particular type of co-occurrence: word-word, word-document, and
word-label. An embedding is then generated for each vertex in the
graph, and documents are represented by averaging the embeddings of
the vertices corresponding to the words they contain. However, the
embeddings of the vertices corresponding to the labels are not used
directly by the classifier, but only concur in the generation of the
word embeddings.

\cite{joulin2016bag} and~\cite{bojanowski2017enriching} proposed
\fasttext, a variant of the CBOW architecture for text classification
that generates both word embeddings and label embeddings.  \fasttext\
seeks to predict one of the document's labels (instead of the central
word) and incorporates further tricks (e.g., n-gram features, sub-word
information) to further improve efficiency.  As a variant of CBOW, and
similarly to PTEs, \fasttext\ represents a document as a simple
average of word embeddings.~\cite{wang2018joint} proposed a model
called \emph{Label Embedding Attentive Model} (LEAM), which jointly
embeds words and labels in the same latent space.  Text
representations in LEAM are conditioned on the compatibility between
words and labels according to an attention model.  The label-embedding
attentive model allows LEAM to go beyond simply averaging word
embeddings (as, e.g., PTE and \fasttext\ do), and to weight
differently the contribution of the word embeddings in a non-linear
fashion.  Once words and labels are embedded in a common vector space,
word-label compatibility is measured via cosine similarity.  Our
method instead models these compatibilities directly, without
generating intermediate embeddings for words or labels.
% They represents text through an attentin model, i.e., by differently
% weighting the contribution of each word embedding as attended by the
% label embedding
% signal %or: as quantified via a label-attentive score
% instead of simply averaging as PTE and \fasttext\ do) and
% considering non-linear interactions between phrases and labels.

A differentiating aspect of our method is that it keeps the modelling
of word-class interactions separate from the original word embedding.
Word-class correlations are confined in a dedicated vector space,
whose vectors enhance (by concatenation) the unsupervised
representations.  The net effect is an embedding matrix that is better
suited to classification, and imposes no restriction to the network
architecture using
it. %\alexcomment{Forse qui ci vorrebbe il chiarimento -class vs -label}

\section{Method}
\label{sec:method}

\noindent Let $\mathcal{C}=\{c_{1},\ldots,c_{m}\}$ be the
classification scheme (a.k.a.\ \emph{codeframe}).
% \blue{Text classification can be formalized as the task of
% approximating an unknown target function
% $\Phi:\mathcal{D}\rightarrow \{0,1\}^{m}$, mapping documents from
% the domain $\mathcal{D}$ into class labels assignments\footnote{The
% label vector would contain a single 1 in single-label multiclass
% problems ($m>2$) and binary classification ($m=2$), and any
% combination of 0's and 1's in multilabel multiclass settings.}, by
% means of a \emph{classifier}
% $\hat\Phi:\mathcal{D}\rightarrow \{0,1\}^{m}$.  } \fsebcomment{Non è
% del tutto vero.}
We consider multiclass classifiers
$h:\mathcal{D}\rightarrow \{0,1\}^{m}$, mapping documents from a
domain $\mathcal{D}$ into vectors of $m$ binary class labels. (The
label vector contains a single 1 in single-label multiclass
classification ($m>2$) and in binary classification ($m=2$), and any
combination of 0's and 1's in multilabel multiclass classification.)
We are interested in equipping the classifiers with continuous
distributed representations of words, i.e., with an embedding function
${E:V\rightarrow \mathbb{R}^{r}}$ (sometimes called the \emph{lookup
table}, and often presented simply as a matrix $\mathbf{E}$) mapping
terms in the vocabulary $V$ (the set containing any desired textual
feature, e.g., words, stems, word n-grams, or any such surface form of
interest) into a dense $r$-dimensional vector space.  Most methods
based on distributional semantics learn the mapping $E$ on the basis
of how terms are distributed in an external corpus $\mathcal{D}'$ of
textual data (sometimes huge, and sometimes unrelated to the domain
$\mathcal{D}$).  We instead investigate task-specific mappings, i.e.,
mappings specific to the domain $\mathcal{D}$ and codeframe
$\mathcal{C}$, based on how terms are distributed across classes.
% \fsebcomment{Dobbiamo migliorare la formulazione in modo da rendere
% la funzione $\eta$ e la standardizzazione indipendenti.}

We define the word-class embedding $E(t_{i})\in\mathbb{R}^{r}$ of term
$t_{i}$ with respect to codeframe $\mathcal{C}$ as
\begin{equation}
  \label{eq:emb}
  E(t_{i})=\psi(\eta(t_{i},c_{1}),\ldots,\eta(t_{i},c_{m})) \in \mathbb{R}^{r}
\end{equation}
\noindent where $\eta: V\times \mathcal{C} \rightarrow \mathbb R$ is a
real-valued function that quantifies the
% a priori task-specific belief about the
correlation between term $t_{i}\in V$ and class
$c_{j}\in \mathcal{C}$, and where
$\psi: \mathbb{R}^{m} \rightarrow \mathbb{R}^{r}$ is any projection
function mapping vectors of class-conditional priors into an
$r$-dimensional embedding space. (More details on both $\eta$ and
$\psi$ later on.)
% \blue{The role of $\psi$ is to control the length of the embedding
% when the number of classes is high ($\psi$ can be, e.g., a PCA
% process).}
%
% The functional $\eta$ can be thought of as computing the degree of
% correspondence between words and classes.  Different interpretations
% of what ``correspondence'' is, open the doors
% to % / allow for / permit /
% different mathematical tools (e.g., correlation measures,
% statistical dependency tests, measures of entropy reduction, etc.)
% to intervene in the definition of word-class embeddings.  The true
% $\eta$ is certainly unknown, but can be approximated empirically by
% inspecting the correspondence (whichever the formalization) between
% terms and labels made known through the realizations present in a
% finite sample of $n$ labelled documents
The value of $\eta(t_{i},c_{j})$ can be estimated from a training set
of labelled documents $L=\{(x_{k},\mathbf{y}_{k})\}_{k=1}^{n}$, with
$x_{k}$ the $k$-th training document and
$\mathbf{y}_{k}\in \{0,1\}^{m}$ the binary vector indicating the class
labels attributed to $x_{k}$.
% \andscomment{Il pezzo qui sotto lo leverei.}  \strikeout{The
% training set thus consists of pairs of instances $(x_{k},y_{k})$
% % with %$x_{k}=[t_{1}, t_{2}, \ldots, t_{l_{k}}]$
% where document $x_{k}$ establishes the context in which its
% constituent terms relate to the labels in $y_{k}$.}
We make the default assumption that
% (although this is not necessary \alexcomment{strictly necessary?
% mandatory? letto cosi sembra che lo facciamo senza motivo}) that
the same training set $L$ is also used by the supervised learning
algorithm for generating the classifier $h$.
% \andscomment{Fino a qui.} \fsebcomment{Io leverei la prima frase ma
% non la seconda.}  \alexcomment{ Io lascerei tutti e due.}

We now detail the embedding generation process using matrix notation.
% Our WCEs are computed as follows.
First, we map set $L$ into a matrix
$\mathbf{X}\in\mathbb{R}^{n\times v}$, where
% $v$
$v=|V|$ is the vocabulary length, and where $\mathbf{X}$ consists of
the vectorial representations of the documents in $L$
% (\blue{contexts))
according to a weighted (e.g., tfidf, or BM25) ``bag-of-words''
feature model.
% \strikeout{The vocabulary consists of any desired textual features
% (e.g., words, stems, word n-grams, or any such surface form of
% interest).} \alexcomment{Anticipato}
Note that the step of mapping
% $\{(x_{i},y_{i})\}_{i=1}^{n}$ \blue{the documents}
$L$ into $\mathbf{X}$ is specific to the WCE generation procedure, and
imposes no restrictions on the supervised learning method to be used,
which may instead rely on a different mechanism for representing
documents.
% \andscomment{(for example, the method against which we compare in
% Section \ref{sec:results} use different formulations of
% $\mathbf{X}$)}.
Similarly, we create a document-class binary matrix
$\mathbf{Y}\in\{0,1\}^{n\times m}$, consisting of the $n$ binary
vectors $\mathbf{y}_{k}\in \{0,1\}^{m}$.  We generate a word-class
matrix $\mathbf{A}\in\mathbb{R}^{v\times m}$ as
\begin{equation}
  \label{eq:A}
  \mathbf{A}=\mathbf{X}_{1}^\top \mathbf{Y}
\end{equation}
\noindent where $\mathbf{X}_{1}$ denotes the matrix $\mathbf{X}$ with
L1-normalized columns.
% \footnote{A vector $\mathbf{v}\in\mathbb{R}^{n}$ is said to be
% L1-normalized if $\sum_{x=1}^{n} v_{x} = 1$. The L1-normalized
% version of an arbitrary non-zero vector
% $\mathbf{v}\in\mathbb{R}^{n}$ can be computed as
% $\mathbf{v}_{1}=\mathbf{v}/\sum_{x=1}^{n} v_{x}$.}
Element $a_{ij}$ of matrix $\mathbf{A}$ thus represents the
correlation between the $i$-th feature and the $j$-th class across the
$n$ labelled documents, as quantified by the dot product.\footnote{It
is worth recalling that the bag-of-words model tends to produce
matrices that are highly sparse. Many software packages take advantage
of this sparsity in order to compute matrix multiplication
efficiently, at a cost that, in practice, falls far below the
asymptotic bound $O(vnm)$.  We discuss empirical computational
complexity issues in Section~\ref{sec:times}.}
% \fsebcomment{Qui ci vorrebbe una presenza esplicita della funzione
% $\eta$, di cui il dot product sia istanza.}  \fsebcomment{Dovremmo
% introdurre nell'equazione qui sopra una funzione $\beta$ che sia un
% placeholder per il dot product o, in alternativa, per information
% gain, chi quadro, etc.} \alexcomment{Ho usato $\eta$ per consistenza
% con la correspondence function di DCI}

We may expect a randomly chosen term to show no a priori significant
correlation with a randomly chosen class label.  We thus want to
choose as our $\eta$ function one that is centered at the expected
correlation value (i.e., the value of correlation that may simply be
explained by chance).  In other words, we want $\eta$ to return
positive (resp., negative) values whenever the presence of $t_{i}$
brings stronger (resp., weaker) evidence that the document is in
$c_{j}$ than this expected value, and close to 0 when the presence of
$t_{i}$ brings no significant evidence about the presence of $c_{j}$.
% and negative values when the presence of $t_{i}$ is
% instead %an indication against the presence of $c_{j}$.
A natural way to fulfill this requirement is through
\emph{standardizing}.
%
% \strikeout{$\mathbf{Q}_{ij}$ happens to be inevitably biased towards
% the frequency of term
%% $f_{i}$ (the $i$-th column of $\mathbf{X}$)
% \blue{$t_{i}$} and the prevalence of class
%% $c_{j}$ (the $j$-th column of $\mathbf{Y}$).
% \blue{$c_{j}$}. \fsebcomment{Potrebbe valer la pena introdurre un
% paio di frasi di motivazione: perché il bias dovuto alla prevalence
% e alla frequency non è cosa buona? qual è l'intuizione su cui questa
% normalizzazione si basa?}  It is thus necessary to factor out the
% cumulative aggregation that is simply explained by chance.
%% , i.e., $\mathbb{E}[f_{i}]\cdot\mathbb{E}[c_{j}]$.
%%
% At this point, and for reasons that will become clear shortly, we
% only factor out the prevalence bias introduced by each feature
% $f_{i}$, i.e., $\sum_{k=1}^{n} \mathbf{X}_{ki}$ (which corresponds
% to the L1-norm $||f_{i}||_{1}$) by defining
% $\mathbf{R}_{ij} \leftarrow \frac{\mathbf{Q}_{ij}}{\sum_{k=1}^{n}
% \mathbf{X}_{ki}}$ }
%
%\begin{equation}
%  \mathbf{R}_{ij} \leftarrow \frac{\mathbf{Q}_{ij}}{\sum_{k=1}^{n} \mathbf{X}_{ki}}
%  \label{eq:l1norm}
%\end{equation}
%
%and finally
We thus (independently) standardize each of the $m$ dimensions of
$\mathbf{A}$ so that the resulting matrix $\mathbf{S}$ is such that
the distribution of the elements in its columns has
% its elements $s_{ij}$ has
zero mean and unit variance, i.e.,
\begin{equation}
  s_{ij} \leftarrow z_{j}(a_{ij})=\frac{a_{ij} - \overline{\mu}_{j}}{\overline{\sigma}_{j}}
  \label{eq:finalmatrix}
\end{equation}
\noindent where $z_{j}$ denotes the function that returns standard
scores (a.k.a.\ $z$-scores) for column $j$ (i.e., for the random
variable %$\mathbf{A}_{*j}$
% (note the $*$)
which takes on values $\{a_{1j},\ldots,a_{vj}\}$), with sample mean
\begin{equation}
  \overline{\mu}_{j}=\frac{1}{v}\sum_{i=1}^{v} a_{ij}
\end{equation}
%
% \noindent is the sample mean of the $j$-th column and
\noindent and sample standard deviation
\begin{equation}
  \overline{\sigma}_{j}=\sqrt{\frac{1}{v-1}\sum_{i=1}^{v}(a_{ij}-\overline{\mu}_{j})^2}
\end{equation}
%
% \noindent is the sample standard deviation of the $j$-th dimension
% in $\mathbf{R}$.  \blue{$\mathbf{A}$}.
%
\noindent
% \alexcomment{Forse spiegare il calcolo della media e standard
% deviation risulta troppo elementare per un journal tipo JMLR?}
Note that the resulting random variable which takes on values
$\{s_{1j},\ldots,s_{vj}\}$ is unbiased with respect to the feature
prevalence of $t_{i}$ and the prevalence of class $c_{j}$.  The reason
is that the feature prevalence has been factored out after the L1
normalization of the columns of $\mathbf{X}$, while the class
prevalence has become a constant factor for each column in
$\mathbf{A}$, and is thus implicitly factored out during
standardizing.
% The bias introduced by the prevalence of each feature has explicitly
% been factored out after the L1 normalization of the columns in
% $\mathbf{X}$.}  Note that there is no need to explicitly factor out
% the prevalence bias for class $c_{j}$ because it becomes a constant
% factor for each column \blue{in $\mathbf{A}$} that is implicitly
% factored out during standardizing.  \strikeout{Had we normalized, in
% Equation~\ref{eq:l1norm}, also by the class prevalence, instead of
% applying standardizing, the whole procedure would be reminiscent of
% the Pointwise Mutual Information (PMI) measure. We discuss (and
% experiment with) PMI and other well-known metrics borrowed from
% information theory as \blue{alternative implementations of the
% $\eta$ function} alternatives to our method in
% Section~\ref{sec:tsr}.}

Function $\eta$ is thus
\begin{equation}
  \eta(t_{i},c_{j})=z_j(\mathbf{t}_{i}^\top \mathbf{c}_{j})
  \label{eq:eta}
\end{equation}
\noindent where $\mathbf{t}_{i}\in \mathbb{R}^{n}$ is the
L1-normalized column vector of weighted values for term $t_{i}$ in
$\mathbf{X}$ and $\mathbf{c}_{j}\in \mathbb{R}^{n}$ is the binary
column vector of class $c_{j}$ in $\mathbf{Y}$.  In
Section~\ref{sec:tsr} we experiment with functions alternative to the
dot product as the instantiation of function $\eta$, including ones
that, unlike the dot product, pay equal attention to positive and
negative correlation.
% do also cater for negative correlation (and not only positive one).
% A feature-class matrix $\mathbf{Q}\in\mathbb{R}^{v\times m}$ is then
% generated as
%
% \begin{equation}
%  \label{eq:A}
%  \mathbf{A}=\mathbf{X}_{1}^\top \mathbf{Y}
%\end{equation}
%
%\noindent \blue{where $\mathbf{X}_{1}$ denotes the matrix
% $\mathbf{X}$ with L1-normalized columns, and}
%% $\mathbf{Q}_{ij}$
% \blue{$a_{ij}=\eta(t_{i},c_{j})$} represents the (positive)
% correlation between the $i$-th feature and the $j$-th class across
% the $n$ labelled documents, as quantified by the dot product.
% \fsebcomment{Dovremmo introdurre nell'equazione qui sopra una
% funzione $\beta$ che sia un placeholder per il dot product o, in
% alternativa, per information gain, chi quadro, etc.}\alexcomment{Ho
% usato $\eta$ per consistenza con la correspondence function di DCI}
% It is worth recalling that the bag-of-words model tends to produce
% matrices that are highly sparse. Many software packages exist which
% take advantage of this sparsity in order to compute matrix
% multiplications efficiently, at a cost that, in practice, falls far
% below the asymptotic bound $O(vnm)$ (we discuss empirical
% computational complexity issues in Section~\ref{sec:times}).
 
There are additional motivations behind the use of standardizing.  On
one hand, the zero-mean property establishes the zero-vector as a
natural choice for any possible future term not encountered at
training time, since the zero-vector would indicate that the term
shows no a priori correlation to any of the classes.  (Further
considerations regarding the treatment of out-of-vocabulary terms are
discussed in Section~\ref{sec:predict}.)  On the other hand, unit
variance guarantees that all classes contribute approximately equally
to the representation, which reinforces the possibility that the
downstream classifier performs well on all classes.

For the moment being, let us simply define the projector $\psi$ in
Equation~\ref{eq:emb} to be the identity function (thus forcing $r$ to
be equal to $m$; we will come back to this in
Section~\ref{sec:method:codeframes}); then $\mathbf{S}$ is the
resulting WCE matrix. Arranged in rows are the WCEs, that encode how
each word is distributed across the classes in the codeframe.
% How it distributes across the classes in the codeframe.  The WCEs,
% arranged in rows, are the distributional fingerprint For any term,
% its WCE explicitly encodes how it distributes across the classes in
% the codeframe.  There is an additional motivation behind the use of
% standardizing. Since the WCEs are obtained from the labelled
% collection, no representation can be found for out-of-vocabulary
% terms that might eventually arise during test. In WCEs, representing
% those terms as zero vectors turns out to be a natural choice thanks
% to standardizing, thus indicating the unseen term shows no a-priori
% correlation to any of the classes. The use of pre-trained word
% embeddings could mitigate this problem: since they are obtained
% offline from vast collections of texts, it is fairly likely that a
% representation for a unseen term in a test document exist in the
% pre-trained set after all (unless it is a domain-specific term).  In
% such cases, we explore to predict the supervised representation from
% the pre-trained representation (see Section~\refexp:predict}).
The WCE matrix $\mathbf{S}$ can finally be concatenated with any other
pre-trained word embedding matrix $\mathbf{U}$ (as those produced by,
e.g., \glove\ or \wordvec) to define the embedding matrix
$\mathbf{E}$.

% ----------------------------------------------------

\subsection{Large Codeframes}
\label{sec:method:codeframes}

\noindent The necessity of dealing with large codeframes could easily
cause the optimization of neural models relying on WCEs to become
intractable. % due to memory constraints.
The reason is that, in many applications of text classification,
hundreds of thousands of features are generated, and the newly added
WCEs lie on a (dense) vector space with as many dimensions as classes
in the codeframe.  In such cases we might want $\psi$ to implement a
dimensionality reduction technique, thus mapping $m$-dimensional
vectors into an $r$-dimensional space, with $r<m$.
% \alexcomment{Non neccessariamente $r\ll m$, e.g., in WIPO-gamma,
% $m=300$, $r=613$.} %$r\ll m$.

% \strikeout{In such cases, we replace $\mathbf{S}$ with a low-rank
% approximation obtained by principal component analysis (PCA).}
In this work we assume $\psi$ to be implemented via principal
component analysis (PCA), in order to replace $\mathbf{S}$ with a
low-rank approximation of it.  In the experiments of
Section~\ref{sec:exp}, when dealing with codeframes with $m>300$ we
choose to retain only the $300$ principal components with the largest
eigenvalues (i.e., those explaining the largest variance); in the
literature, $300$ is indeed a popular choice for the size of word
embeddings.  (Somehow abusing notation, and when clear from context,
we will use symbol $\mathbf{S}$ to either denote $\mathbf{S}$
\textit{or} its low-rank approximation, assuming the application of
PCA to be implicit whenever $m>300$.)
% This idea is not novel, and actually grounds on work related to
% extreme classification~\cite{}.

% Alternative solutions to this problem
Alternative ways for implementing $\psi$ might be found in the class
of \emph{label-embedding} approaches from the \emph{extreme multilabel
text classification}
literature~\cite{bhatia2015sparse,hsu2009multi,yu2014large} already
discussed in Section~\ref{sec:relwork-label}, or more generally in
dimensionality reduction
techniques~\cite{baldi2012autoencoders,tSNE}.

% ----------------------------------------------------

\subsection{Regularization}
\label{sec:method:regularization}

\noindent %WCEs inject a task-specific pressure that might unbalance
% the generalization capability of the classifier.  \blue{
WCEs inject a task-specific pressure into the representation mechanism
that might compromise the data-generating process of training and test
documents, % (and thenceforth, the iid assumption).
since, unlike when using pre-trained embeddings, terms from the
training documents have played a role in the generation of WCEs.
% When using WCEs to represent the input documents this might thus
% drift the distribution of training
% documents %thus introducing some divergency
% with respect to the distribution of unseen documents.  }
% \fsebcomment{Andrebbe detto meglio: tecnicamente non è la iid
% assumption che viene compromessa, quanto la generalità della
% rappresentazione.}
Indeed, during preliminary experiments we observed that models
operating with a concatenation of WCEs and pre-trained word-embeddings
tend to incur a much lower training loss than those using pre-trained
word-embeddings only, but the former tend to perform substantially
worse on unseen data (more details on this in Section
\ref{sec:regularization}).  This case of overfitting makes evident the
need for properly regularizing the model.

In order to perform regularization, we apply a variant of
dropout~\cite{srivastava2014dropout} to the embedding layer. Dropout
consists of zeroing random activations in order to prevent nodes from
co-adapting. Since dropout is only applied in the training phase, the
activation values are scaled by $(1-p)^{-1}$ during training, with $p$
the drop probability, in order to keep the expected activation
consistent with the test phase.

We only apply dropout to the WCEs, and keep the unsupervised
embeddings untouched (there is no reason to believe the unsupervised
embeddings fit more the training documents than the unseen documents).
Let
$\mathbf{E}=[\mathbf{U}\oplus\mathbf{S}]\in\mathbb{R}^{v\times(q+r)}$
represent the entire embedding layer, consisting of the concatenation
(here denoted by the $\oplus$ operator) of the unsupervised
$q$-dimensional matrix $\mathbf{U}$ and the supervised $r$-dimensional
matrix $\mathbf{S}$. In order to bring to bear correct expected
activations during test, we compute the scaling at training time
as\footnote{Since we undertake a stochastic optimization, this
actually applies to \emph{batches} of data.}
\begin{equation}
  D(\mathbf{E})=\frac{[\mathbf{U} \oplus (1-p) \cdot 
  d(\mathbf{S})]}{1 - \frac{pr}{q+r}}
\end{equation}
\noindent where $d$ indicates \emph{dropout} and $D$ indicates
\emph{supervised dropout}.
% The $\xi(\mathbf{E})$ matrix is the final output of our embedding
% generation process. \fsebcomment{Alejandro, controlla la parte qui
% sopra ed eventualmente modifica.}

% p = p_drop drop_from, drop_to = drop_range m = drop_to - drop_from
% #length of the supervised embedding l = input.shape[2] #total
% embedding length corr = (1 - p) input[:, :, drop_from:drop_to] =
% corr * F.dropout(input[:, :, drop_from:drop_to], p=p) input /= (1 -
% (p * m / l))

% Regularization: dropout adjusted. [we might explore the effect of
% dropout in \glove\ as well...]

% -----------------------------------------------------------------

\section{Experiments}
\label{sec:exp}

\noindent In this section we turn to describing the experiments that
we have carried out in order to quantify the contribution of WCEs to
multiclass text classification. In order to make all the experiments
discussed in this paper fully reproducible, we make available at
\url{https://github.com/AlexMoreo/word-class-embeddings} the code that
implements our method and all the baselines used in this work.

% ----------------------------------------------------

\subsection{Datasets}
\label{sec:datasets}

\noindent In our experiments we use the following six publicly
available datasets:

\begin{itemize}
\item \textsc{Reuters-21578} is a popular multilabel dataset which
  consists of a set of 12,902 news stories, partitioned (according to
  the ``ModApt\'e'' split we adopt) into a training set of 9,603
  documents and a test set of 3,299
  documents.\footnote{\url{http://www.daviddlewis.com/resources/testcollections/reuters21578/}}
  In our experiments we restrict our attention to the 115 classes with
  at least one positive training example. This dataset presents cases
  of severe imbalance, with many classes containing fewer than 5
  positive examples.

\item \textsc{20Newsgroups} is a single-label test collection of
  approximately 20,000 posts on Usenet discussion groups, nearly
  evenly partitioned across 20 different newsgroups
  (classes).\footnote{\url{http://qwone.com/~jason/20Newsgroups/}. Note
  that this version of \textsc{20Newsgroups} is indeed single-label:
  while a previous version contained a small set of document with more
  than one label (corresponding to posts that had been cross-posted to
  more than one newsgroup), that set is not present in this version we
  use.}  In this article we use the ``harder'' version of the dataset,
  i.e., the one from which all metadata (headers, footers, and quotes)
  have been removed.\footnote{\label{foot:harder}While some previous
  papers (e.g.,~\cite{tang2015pte}) have reported substantially
  higher scores for this dataset, it is worth noticing that we use a
  harder, more realistic version of the dataset than has been used in
  those papers. Following~\cite{moreo2019learning}, in our version we
  removed all headers, footers, and quotes, since these fields contain
  terms that are highly correlated with the target labels, thus making
  the classification task unrealistically easy; see
  \url{http://scikit-learn.org/stable/datasets/twenty_newsgroups.html}
  for further details. Our results are indeed consistent with other
  papers following the same policy.}

\item \textsc{Ohsumed}~\cite{Hersh94} is a dataset consisting of a
  set of MEDLINE documents spanning the years from 1987 to
  1991.\footnote{\url{http://disi.unitn.it/moschitti/corpora.htm}}
  Each entry consists of summary information relative to a paper
  published on one of 270 medical journals. The available fields are
  title, abstract, MeSH indexing terms, author, source, and
  publication type.  Following~\cite{joachims1998text}, we restrict
  our experiments to the set of 23 cardiovascular disease classes, and
  we use the 34,389 documents of year 1991 that have at least one of
  these 23 classes. Since no standard training/test split has been
  proposed in the literature we randomly partition the set into a part
  used for training (70\% of the documents) and a part used for
  testing (the other 30\%).

\item \textsc{RCV1-v2} is a dataset comprising 804,414 news stories
  published by Reuters from Aug 20, 1996, to Aug 19, 1997; for text
  classification purposes it is traditionally split into a training
  set consisting of the (chronologically) first 23,149 documents (the
  ones written in Aug 1996), and a test set consisting of the last
  781,265 documents (the ones written from Sep 1996
  onwards).\footnote{Available from
  \url{http://www.ai.mit.edu/projects/jmlr/papers/volume5/lewis04a/lyrl2004_rcv1v2_README.htm}}
  In our experiments we use this dataset in its entirety, and stick to
  the standard training/test split described above.  \textsc{RCV1-v2}
  is multilabel, i.e., a document may belong to several classes at the
  same time.  Of the 103 classes of which its ``Topic'' hierarchy
  consists, in our experiments we have restricted our attention to the
  101 classes with at least one positive training example. This
  dataset is the one with the largest test set in our experiments.

\item \textsc{JRC-Acquis} (version 3.0) is a collection of legislative
  texts of European Union law written between the 1950s and
  2006~\cite{Steinberger:2006rz}.\footnote{\url{https://ec.europa.eu/jrc/en/language-technologies/jrc-acquis}}
  \textsc{JRC-Acquis} is publicly available for research purposes, and
  covers 22 official European languages. We restrict our attention to
  the English subset, which consists of 20,370 documents.  For our
  experiments, we consider the 13,137 documents written in the [1950,
  2005] interval as the training set, and leave the remaining 7,233
  documents written in 2006 as the test set. The dataset is multilabel
  and is labelled according to the EuroVoc thesaurus. We focus on the
  2,706 classes with at least one positive element in the training
  set. This dataset is the one with the largest codeframe in our
  experiments.

\item \textsc{WIPO-gamma} is a test collection of patent
  documents.\footnote{\url{https://www.wipo.int/classifications/ipc/en/ITsupport/Categorization/dataset/}}
  Documents are labelled according to the International Patent
  Classification (IPC) taxonomy, covering patents and patent
  applications in all areas of technology.  We focus on the
  single-label version labelled at the subclass level in the IPC
  hierarchy. For our experiments, we extract the abstract field of the
  documents (thus discarding the list of inventors, list of applicant
  companies, claims, and the long description), and follow the
  train/test split made available by the WIPO organization.  The
  dataset contains a total of 1,118,299 documents, of which 896,363
  (80\%) is used as the training set, and the remaining 221,936 (20\%)
  are used for test.  This dataset is the one with the largest
  training set in our experiments.
\end{itemize}

\noindent Details of these datasets are given in Table
\ref{tab:datasets}.  Note that the datasets chosen cover a broad
spectrum of experimental conditions, including single-label and
multilabel scenarios, a number of classes ranging from tens
(\textsc{20Newsgroups}) to thousands (\textsc{JRC-Acquis}), a number
of documents from small (\textsc{Reuters-21578}) to very large
(\textsc{WIPO-gamma}), from well balanced datasets
(\textsc{20Newsgroups}) to severely imbalanced ones (e.g.,
\textsc{RCV1-v2}), etc. Note also that the first four datasets
(\textsc{Reuters-21578}, \textsc{20Newsgroups}, \textsc{Ohsumed},
\textsc{RCV1-v2}) are probably the most popular datasets in text
classification research; together with the fact that they are all
publicly available, this guarantees a high level of comparability (and
interpretability) to our results.
In this work we restrict our attention to text classification by
topic, and leave other dimensions (e.g., classification by sentiment)
for future work (see also the discussion in
Section~\ref{sec:limitations}).

\begin{table}[t]
  \resizebox{\textwidth}{!}{%
  \begin{tabular}{|l||c|r|r|r|r|r|r|r|r|r|r|r|}
    \hline
    \multicolumn{1}{|c||}{\textbf{\side{Dataset}}} & \centercell{\side{Type}} & \centercell{\side{\# Classes}} & \centercell{\side{\# Train docs}}  & \centercell{\side{\# Test docs}}  & \centercell{\side{Total \# of docs\hspace{.5em}\mbox{}}} & \centercell{\side{Vocabulary}} & \centercell{\side{OOV}} & \centercell{\side{\# Words}}  & \centercell{\side{Prev(mean)\hspace{.5em}\mbox{}}} & \centercell{\side{Prev(std)}}  & \centercell{\side{Prev(min)}} & \centercell{\side{Prev(max)}} \\ \hline\hline
    \textsc{Reuters-21578} & ML & 115 & 9,603 & 3,299 & 12,902 & 8,250 & 24,094 & 1.7M & 83.9 & 314.3 & 1 & 2,877  \\ \hline
    \textsc{20Newsgroups} & SL & 20  & 11,314  & 7,532 & 18,846 & 17,184  & 112,594  & 3.5M & 565.7  & 56.8 & 377 & 600 \\ \hline
    \textsc{Ohsumed}  & ML & 23  & 24,061  & 10,328  & 34,389 & 18,238  & 44,062 & 6.2M & 1,734.3 & 1,523.6 & 301 & 6,729  \\ \hline
    \textsc{RCV1-v2} & ML & 101 & 23,149  & 781,265 & 804,414  & 24,816  & 384,327  & 188.1M & 709.9 & 1,417.2 & 1 & 10,282 \\ \hline
    \textsc{JRC-Acquis}  & ML & 2,706  & 13,137  & 7,233 & 20,370 & 21,109  & 141,800  & 28.3M  & 25.7 & 58.6 & 2 & 1,151  \\ \hline
    \textsc{WIPO-gamma} & SL & 613 & 896,363 & 221,936 & 1,118,299 & 114,802 & 395,570  & 417.8M & 1,462.3 & 4,354.6 & 1 & 63,465 \\ \hline
  \end{tabular}%
  }
  \caption{Details of the datasets we use in this research. Column
  ``Type'' indicates whether the classification is multilabel (ML) or
  single-label (SL). Column ``Vocabulary'' shows the number of terms
  occurring at least 5 times in the training set, while Column ``OOV''
  shows the number of terms occurring in fewer than 5 training
  documents or exclusively in test documents. Columns with prefix
  ``Prev'' indicate a few statistics about class prevalence in the
  training set.}
  \label{tab:datasets}
\end{table}

We pre-process text by using the default analyzer available in the
\texttt{scikit-learn}
framework\footnote{\url{http://scikit-learn.org/}} (which applies
lowercasing, stop word removal, punctuation removal), and by masking
numbers with a dedicated token.  For the computation of the WCEs, we
retain all terms (unigrams) appearing at least 5 times in the training
set.  However, in experiments involving pre-trained embeddings we also
consider those out-of-vocabulary (OOV) terms for which a pre-trained
embedding exists; these terms are represented by the zero vector in
the WCE space (see Section~\ref{sec:method}).  This is a major
advantage of also using pre-trained embeddings, which allow neural
models to also make sense (at testing time) of terms unseen at
training time if they have anyway been encountered during the
pre-training phase (see Section~\ref{sec:predict} for more on this).

% As the weighting criterion we use a version of the well-known
% $\mathit{tfidf}$ method, expressed as
% %
% \begin{equation}
% \mathit{tfidf}(f,d)=\log\#(f,d)\times \log \frac{|Tr_{i}|}{|d'\in Tr_{i} : \#(f,d')>0|}
% \label{eq:tfidf}
% \end{equation}
% %
% \noindent where $\#(f,d)$ is the raw number of occurrences of
% feature $f$ in document $d$ and $\lambda_{i}$ is the language $d$ is
% written in; weights are then normalized via cosine normalization, as
% %
% \begin{equation}
% w(f,d)=\frac{\mathit{tfidf}(f,d)}{\sqrt{\sum_{f'\in F_{i}} \mathit{tfidf}(f',d)^2}}
% \label{eq:tfidfnorm}
% \end{equation}

% ----------------------------------------------------

\subsection{Evaluation Measures}

\noindent As the effectiveness measure we use $F_{1}$, the harmonic
mean of precision ($\pi$) and recall ($\rho$), defined as
$F_{1}=(2\pi\rho)/(\pi+\rho)=(2\mathrm{TP})/(2\mathrm{TP}+\mathrm{FP}+\mathrm{FN})$,
where $\mathrm{TP}$, $\mathrm{FP}$, $\mathrm{FN}$, are the numbers of
true positives, false positives, false negatives, from the binary
contingency table.  We take $F_{1}=1$ when
$\mathrm{TP}=\mathrm{FP}=\mathrm{FN}=0$, since the classifier has
correctly classified all examples as negative.

As defined above, $F_{1}$ is a measure for binary classification
only. For multiclass classification, we average $F_{1}$ across all the
classes of a given codeframe by computing both micro-averaged $F_{1}$
(denoted by $F_{1}^{\mu}$) and macro-averaged $F_{1}$ (denoted by
$F_{1}^{M}$).  $F_{1}^{\mu}$ is obtained by (i) computing the
class-specific values $\mathrm{TP}_{j}$, $\mathrm{FP}_{j}$, and
$\mathrm{FN}_{j}$, (ii) obtaining $\mathrm{TP}$ as the sum of the
$\mathrm{TP}_{j}$'s (same for $\mathrm{FP}$ and $\mathrm{FN}$), and
then applying the $F_{1}$ formula.  $F_{1}^{M}$ is obtained by first
computing the class-specific $F_{1}$ values and then averaging them
across the classes.

% -------------------------------------------------

\subsection{Supervised Learners for Classifier Training}
\label{sec:classifiertraining}

\noindent We test the contribution of WCEs to text classification
using a ``traditional'' (i.e., non-neural), high-performance learner
(support vector machines), a well-known, deep learning -based library
for text classification (\fasttext), and three popular architectures
based on deep neural networks (convolutional neural networks,
long-short term memory networks, and attention models).

For each such system we explore different variants, corresponding to
different ways of instantiating the embedding matrix $\mathbf{E}$.  As
the pre-trained embeddings we use the biggest set of \glove\ vectors
made available, and consisting of 2.2M 300-dimensional word embeddings
generated from a text corpus of 840 billion
tokens.\footnote{\url{http://nlp.stanford.edu/data/glove.840B.300d.zip}}
(In Section~\ref{sec:exp:pre-trained} we report results of using
\wordvec\ embeddings instead of \glove\ embeddings.) The variants we
explore are the following:
\begin{description}

\item[\random:] \emph{randomly} initialized \emph{trainable}
  embeddings (their dimensionality is optimized from the range
  $\{50,200,300\}$ on a validation set).
  
\item[\glovest:] \emph{static} pre-trained \glove\ embeddings.
  
\item[\glovetr:] \emph{trainable} vectors initialized with pre-trained
  \glove\ embeddings.
  
\item[\glovern:] \emph{trainable} embeddings initialized as the
  concatenation of pre-trained \glove\ embeddings and \emph{random}
  embeddings. The random embeddings are chosen to have the same
  dimensionality as WCEs. This configuration serves for control
  purposes, in order to ensure that any possible relative improvement
  brought about by the use of WCEs cannot merely be attributed to the
  presence of more parameters in the embedding layer.
  
\item[\glovewcest:] \emph{static} concatenations of pre-trained
  \glove\ embeddings and WCEs.
  
\item[\glovewcetr:] \emph{trainable} embeddings initialized with the
  concatenation of pre-trained \glove\ embeddings and WCEs.
  
\end{description}
\noindent We perform hyperparameter search via grid-search on the
validation set, independently for each combination of type
$\langle$dataset, architecture, variant$\rangle$.\footnote{We generate
the validation set by randomly sampling 20\% of the training set, with
a maximum of 20,000 documents; the rest is taken to be the training
set proper. We keep the training/validation split consistent across
all methods.} The hyperparameters to be optimized are dependent on the
architecture, and are explained in the sections below.

% ----------------------------------------------------

\subsubsection{Support Vector Machines}

\noindent The problem of training a classifier via SVMs with
unsymmetric costs is stated as the empirical risk minimization
problem~\cite{cortes1995support,Morik99a}\footnote{Note that,
consistently with~\cite{cortes1995support,Morik99a}, in this
formulation we assume the class labels $y_k$ to be in $\{-1,+1\}$,
while in Section~\ref{sec:method} we had assumed them to be in
$\{0,1\}$; the difference is, of course, unproblematic.}
\begin{equation*}
  \begin{aligned}
    \text{minimize:} & & \frac{1}{2}||\mathbf{w}||^2 + C_{+}\sum_{k=1}^{n} \xi_{k}[y_k=+1] + C_{-} \sum_{k=1}^{n} \xi_{k}[y_k=-1] \\
    \text{over:} & & \mathbf{w},b,\xi_{1},\ldots,\xi_{n}\\
    \text{subject to:} & & \forall_{k=1}^{n} :
    y_{k}(\mathbf{w}\cdot\mathbf{x}_{k} + b) \geq 1 - \xi_{k} \\
    & & \forall_{k=1}^{n} : \xi_{k} > 0
  \end{aligned}
\end{equation*}
\noindent where $\mathbf{w}$ and $b$ are the parameters (hyperplane
and bias) of the separation functional, $\xi_{k}$ are the slack
variables for the labelled examples $(x_{k},y_{k})$, $[\cdot]$ is the
indicator function that returns 1 if its argument is true and 0 if it
is false, and $C_{+}$ and $C_{-}$ are two hyperparameters that control
the trade-off between training error and margin for positive and
negative examples, respectively \cite{Morik99a}.  It is convenient to
factor $C_{+}$ and $C_{-}$ as $C_{+}=CJ_{+}$ and $C_{-}=CJ_{-}$, so
that the unsymmetric cost factors are confined to two dedicated
hyperparameters, with $J_{+}$ (resp., $J_{-}$) controlling the amount
by which training error on positive examples (resp., negative
examples) outweighs error on the negatives (resp., positives).  We
follow \cite{Morik99a} and set the cost factors so that the ratio
$J_+/J_-$ equals the ratio $N/P$ between the number $N$ of negative
training examples and the number $P$ of positive training
examples.\footnote{In \texttt{scikit-learn} this is achieved by
setting $J_+=n/(mP)$ and $J_-=n/(mN)$, and corresponds to setting the
parameter \texttt{class\_weight} to ``balanced''.}  The trade-off
between training error and margin, now confined to $C$, becomes the
only hyperparameter we tune. Note that, while in other application
fields the kernel to employ is considered an important parameter to
optimize, in text classification it is customary to employ the linear
kernel, since there are theoretical arguments for its optimality in
these contexts~\cite{Joachims:2001ij}; we thus use the linear kernel
without further ado.

Although the application of SVMs to text classification dates back
to~\cite{Dumais98,joachims1998text}, SVMs are still considered among
the strongest baselines for text classification.
% Other traditional methods popularly used as baselines for modern
% approaches include Logistic Regression and
% k-NN. \alexcomment{Discuss the rationale here, or in the related
% work.}
\footnote{Somehow surprisingly, though, several relevant related works
where SVMs are used as baselines (see,
e.g.,~\cite{joulin2016bag,jiang2018text,zhang2015character}) do not
report the details of how, if at all, they tune the SVM
hyperparameters.}  Having set the unsymmetric cost factors, the most
influential hyperparameter for SVMs is $C$.
% and the cost-factors $J_{+}$ by which training errors on positive
% examples outweigh errors on negative.
We choose the best value for $C$ from the set
$\{10^{-3},10^{-2},\ldots, 10^{+3}\}$ by performing 5-fold
cross-validation on the full set of labelled documents (i.e., the
training set proper plus the validation set); we perform the
optimization of $C$ independently for each class.\footnote{Using
$k$-fold cross-validation ($k$-FCV) on the full set of labelled
documents is a more expensive, but stronger, way of doing parameter
optimization than using a single split between a training set and a
validation set, because $k$-FCV performs $k$ such splits. We here use
$k$-FCV for SVMs and single-split optimization for all the other deep
learning -based architectures because it is realistic to do so, i.e.,
because SVMs are computationally cheap enough for us to be able to
afford $k$-FCV, while neural architectures are not.}  In the
experiments of this paper we use the implementation of SVMs available
in \texttt{scikit-learn}.\footnote{This implementation relies on
\texttt{liblinear}. See
\url{https://scikit-learn.org/stable/modules/generated/sklearn.svm.LinearSVC.html}
for further details.}  We leave the rest of the parameters set to
their default values.
% Since many datasets are highly imbalanced, we set the parameter
% \texttt{class\_weight} to ``balance'' \fsebcomment{... dovremmo
% spiegare a che serve ...}, and leave the rest of the parameters set
% to their default values.  \alexcomment{....regarding
% class\_weight=balance: the cost-factor, by which training errors on
% positive examples outweigh errors on negative examples, is defined
% in proportion to the positive and negative class prevalences....}

Since SVMs do not cater for word embedding fine-tuning, we only report
experiments involving sets of static embeddings. Specifically, for
SVMs we report the following experiments:

\begin{description}

\item[\svmtfidf:] a SVM-based classifier trained on the tfidf matrix
  $\mathbf{X}$ defined in Section~\ref{sec:method};

\item[\svmglove:] a SVM-based classifier trained on the projection
  $\mathbf{X} \mathbf{U}$, with $\mathbf{U}$ the pre-trained \glove\
  embeddings;

\item[\svmglovewce:] a SVM-based classifier trained on the projection
  $\mathbf{X} \mathbf{E}$, with
  $\mathbf{\mathbf{E}=[\mathbf{U} \oplus \mathbf{S}]}$ the
  concatenation of pre-trained \glove\ embeddings and WCEs.

\end{description}

% ----------------------------------------------------

\subsubsection{fastText}

\noindent We report experiments for
\fasttext~\cite{bojanowski2017enriching,joulin2016bag}, which we have
already discussed in Sections~\ref{sec:WCEs} and~\ref{sec:relwork-tc}.

We select hyperparameters via grid-search optimization on the
validation set.
Following~\cite{bojanowski2017enriching,joulin2016bag}, we let the
learning rate vary in $\{0.05, 0.1, 0.25, 0.5\}$ and the number of
epochs in $\{5,10\}$.
% and the number of dimensions in $\{50,200,300\}$ \blue{(300 was not
% an dimension tested in the study
% of~\cite{bojanowski2017enriching,joulin2016bag}, but we add it for
% consistency with the size of our randomly trained embeddings)}.
% \alexcomment{In realtà 300 l'ho aggiunto io per consistenza con la
% dimensione GloVe... vogliamo dirlo?}
Some of our datasets (\textsc{20News\-groups}, \textsc{JRC-Acquis},
\textsc{Ohsumed}, and \textsc{Reuters-21578}) are comparatively
smaller than those tested in the original article; in those cases we
let the number of epochs vary on $\{5,50,100,200\}$ (we observed
drastic improvements when using 100 or 200 epochs, but no further
improvement when going beyond 200).

In our experiments we only consider the case in which \fasttext\
operates with unigram features.  While \fasttext\ has been found to
deliver better results when using bigrams, the adoption of bigrams
would likely lead to similar improvements for the rest of the methods
being compared~\cite{wang2012baselines}, and might only blur the
purpose of this comparison.

For \fasttext\ we only report experiments involving sets of trainable
vectors (combinations \random, \glovetr, \glovern, and \glovewcetr),
since \fasttext\ is implicitly a method for learning embeddings, and
thus does not cater for static vectors. In the case of \glovewcetr, we
do not apply supervised dropout, since we stick to the official
implementation.\footnote{\url{https://fasttext.cc/}}

\subsubsection{Deep-Learning Architectures}

\noindent Since our goal is to quantify the relative improvement (if
any) brought about by concatenating WCEs to pre-trained embeddings, we
adopt the most general and simple formulation for each of our three
deep-learning architectures, and leave the exploration of more
sophisticated models for future research.

Let $(x_k,\mathbf{y}_k)$ be a training instance, with
$x=[t_{1k}, \ldots ,t_{lk}]$ a document consisting of a sequence of
$l$ terms (padded where necessary) labelled with
$\mathbf{y}_k\in\{0,1\}^{m}$. Let
$\mathbf{E}\in\mathbb{R}^{v\times r}$ be the embedding matrix,
containing $v$ $r$-dimensional word embeddings
$\mathbf{e_{i}}\in\mathbb{R}^r$ (where embeddings may either be
initialized randomly, using pre-trained embeddings, or concatenations
of pre-trained and supervised embeddings).  All the tested
architectures use an embedding layer as the first layer on the
network, which transforms every input document $x_k$ into
$\mathbf{x}_k=[E(t_{1k}), \ldots, E(t_{lk})]$,
% \begin{equation}
%   \mathbf{x}=\mathbf{e}_{1} \oplus \mathbf{e}_{2} \oplus \ldots \oplus \mathbf{e}_{l} 
%   \mathbf{X}=\mathbf{e}^\top_{1} \oplus \mathbf{e}^\top_{2} \oplus
%   \ldots \oplus \mathbf{e}^\top_{l}
% \end{equation}
%
% \noindent
where $E(t_{ik})$ is the word embedding in $\mathbf{E}$ for term
$t_{ik}$.
% \strikeout{, and where symbol $\oplus$ denotes
% concatenation. }\alexcomment{Detto in 3.2.}
Different models implement different transformations
\begin{equation}
  \label{eq:docembedding} \mathbf{o}_k=N(\mathbf{x}_k;\Theta)
\end{equation}
\noindent of the input (as defined below) parameterized by $\Theta$,
with $\mathbf{o}_k$ denoting the document embedding of $x_k$.
Finally, $\mathbf{o}_k$ is mapped into the space of label outputs by
\begin{equation}
  \hat{\mathbf{y}}_k=f(\mathbf{o}_k\cdot \mathbf{W} + \mathbf{b})
\end{equation}
\noindent where $\mathbf{W}$ and $\mathbf{b}$ are the parameters
(weight and bias) of an affine transformation, $f$ is a non-linear
function, instantiated as the softmax function for single-label
problems or as the sigmoid function for multilabel problems, and
$\hat{\mathbf{y}}_k\in[0,1]^{m}$ is a vector of $m$ predicted
posterior probabilities, one for each class.  The full set of model
parameters to optimize is thus
$\Theta'=[\Theta;\mathbf{W},\mathbf{b}]$.

We initialize the model parameters using the \emph{Xavier uniform}
method described in~\cite{glorot2010understanding}.  We then train
the model by backprogating the errors, where error is computed as the
\emph{cross-entropy} loss (in the single-label case) or as the
\emph{binary cross-entropy} loss (in the multilabel case).  We carry
out optimization via stochastic gradient descent with the Adam update
rule~\cite{kingma2014adam}.  We set the learning rate to 1e-3 and the
batch size to 100 documents, dynamically padding the sequences to
$l=\min{\{500,l_{max}\}}$, where $l_{max}$ is the length of the
longest document within the batch.

We train the models for a maximum of 200 epochs, but we apply early
stopping whenever 10 consecutive training epochs do not yield any
improvement in the validation set in terms of $F_{1}^{M}$.  An epoch
consists of a full pass over all the training documents.  Since
\textsc{WIPO-gamma} is one order of magnitude larger than the other
datasets, we consider an epoch to be over after 30,000 documents (300
batches) have been processed.  We dump the model parameters whenever
the value of $F_{1}^{M}$ on the validation set improves.  When the
training epochs are over, we restore the best model parameters and
perform one final training epoch on the validation set.
% \alexcomment{drop-prob}

We consider the following network architectures as alternative
implementations of the transformation $N$ of Equation
\ref{eq:docembedding}.

% ----------------------------------------------------

\paragraph{Convolutional Neural Networks.}

\noindent Convolutional Neural Networks (CNN) are a special type of
neural models particularly suited for computer vision, that apply
convolved filters which are robust to position-invariant patterns.  In
text-related
applications~\cite{collobert2011natural,kim2014convolutional} a
convolution is the result of the application of a linear filter to a
matrix consisting of the $w$ word embeddings corresponding to the
words that appear in a sliding window of length $w$, in order to
produce a feature map.

A convolutional layer typically contains and applies many filters,
each of which is followed by a non-linear activation function
(typically: the \emph{rectified linear unit} $ReLU(x)=\max\{0,x\}$,
which is the one we use here) and a max-pooling operation that takes
the maximum value for each filter.  The result is a thus a vector with
as many features as there are filters.

We consider one single convolutional layer~\cite{le2018convolutional}
with $\gamma$ output channels for each window length $w\in\{3,5,7\}$
(i.e., $3\gamma$ output channels in total), where $\gamma$ is a
hyperparameter to be optimized on a validation set. We let $\gamma$
vary in the range $\{64, 128, 256, 512\}$.  The final representation
is a vector $\mathbf{o}_k\in\mathbb{R}^{3\gamma}$, which concatenates
all convolved outputs, followed by the application of a dropout
operator.

\paragraph{Long-Short Term Memory Networks.} Recurrent Neural Networks
(RNNs --~\cite{lai2015recurrent,rumelhart1986learning}) are a
family of network architectures specially devised for processing
sequential data. RNNs apply the same computation to each input in the
sequence. The internal state $\mathbf{h}_{ik}$ at time $i$ is defined
recursively as
$\mathbf{h}_{ik}=f(\mathbf{h}_{(i-1)k},E(t_{ik});\Theta)$, with
$E(t_{ik})$ the embedding of term $t_{ik}$ and $\Theta$ parameterizing
the recurrent function. The model is trained by \emph{Backprogragation
Through Time} (BPTT) via unfolding the recursive computation and
sharing the parameters $\Theta$ across all time steps.  In this work
we adopt the well-known \emph{Long-Short Term Memory}
(LSTM)~\cite{hochreiter1997long} as the recurrent cell.
% , and take the last state $\mathbf{h}_{n}$ to represent the entire
% document.
We apply gradient clipping at $\pm0.1$ in order to avoid exploding
gradients.
% \alexcomment{Say something about not using bi-directional?}
The size $\gamma$ of the hidden state is a hyperparameter of the
model, to be optimized on a validation set from the range
$\{256, 512, 1024, 2048\}$.  The output
$\mathbf{o}_k\in \mathbb{R}^\gamma$ is the final state
$\mathbf{h}_{lk}$ produced by the LSTM.

% ----------------------------------------------------

\paragraph{Attention Models.} 
Attention models (ATTNs) implement criteria that enable the model to
weight differently (i.e., to pay different attention to) the
contribution of intermediate factors in certain computations.
Although attention mechanisms by their
own~\cite{vaswani2017attention} constitute nowadays an entire family
of deep neural models, called
\emph{transformers}~\cite{BERT,peters2018deep,XLNet}, we focus on a
simpler formulation, called \emph{soft-scaled dot-product attention
mechanism}~\cite{luong2015effective}.

This attention mechanism takes all hidden states
$\mathbf{H}_k=[\mathbf{h}_{1k},\ldots,\mathbf{h}_{lk}]$ produced by a
RNN (we use the LSTM here as well) and the document embedding
$\mathbf{o}_k=\mathbf{h}_{lk}$, and computes a vector of attention
weights over all intermediate states, i.e.,
\begin{equation}
  \mathbf{a}_k=\mathrm{softmax}(\mathbf{o}_k^\top\mathbf{H}_k)
\end{equation}
and produces a new output $\mathbf{o}'\in \mathbb{R}^\gamma$ as a
weighted sum
$$\mathbf{o}'_k=\sum_{a_{ik} \in \mathbf{a}_k} a_{ik} \mathbf{h}_{ik}$$ As
for LSTM, for the hidden layer $\gamma$ we choose the size from the
range $\{256, 512, 1024, 2048\}$ that performs best on the validation
set.
% \alexcomment{I think the version we have is wrong}

% ----------------------------------------------------------------------

\subsection{Results}
\label{sec:results}

\noindent Tables~\ref{tab:final-te-macro-F1} and
\ref{tab:final-te-micro-F1} report the $F_{1}^{M}$ and $F_{1}^{\mu}$
results we have obtained.  Since neural architectures use a random
inizialization of the parameters, our results for them are averages
across 10 runs.  (SVMs are deterministic and thus excluded from the
test of statistical significance, which requires the repetition of
random trials.)

% \begin{minipage}
% --------------------------------------------
% MACRO-F1
% --------------------------------------------

\begin{table}[t]
\centering
\resizebox{\textwidth}{!}{%
\begin{tabular}{|l|l||r|r|r|r|r|r|}
\hline
model     & variant &              \centercell{\rotatebox{90}{\textsc{20Newsgroups}\hspace{1ex}\mbox{}}} &                 \centercell{\rotatebox{90}{JRC-Acquis}} &                    \centercell{\rotatebox{90}{\textsc{Ohsumed}}} &                    \centercell{\rotatebox{90}{\textsc{RCV1-v2}}} &              \centercell{\rotatebox{90}{\textsc{Reuters-21578}\hspace{1ex}\mbox{}}} &                 \centercell{\rotatebox{90}{\textsc{WIPO-gamma}\hspace{1ex}\mbox{}}} \\
\hline\hline
\multirow{3}{*}{SVM} & \texttt{tfidf} &                         .670\phantom{$^{\dagger\dagger}\pm$.000} &             \textbf{.397}\phantom{$^{\dagger\dagger}\pm$.000} &                   .673\phantom{$^{\dagger\dagger}\pm$.000} &          \textbf{.584}\phantom{$^{\dagger\dagger}\pm$.000} &                \textbf{.636}\phantom{$^{\dagger\dagger}\pm$.000} &                        ---\phantom{$^{\dagger\dagger}\pm$.000} \\
%     & WCEs &                         .680\phantom{$^{\dagger\dagger}\pm$.000} &                      .243\phantom{$^{\dagger\dagger}\pm$.000} &                   .581\phantom{$^{\dagger\dagger}\pm$.000} &                   .412\phantom{$^{\dagger\dagger}\pm$.000} &                         .575\phantom{$^{\dagger\dagger}\pm$.000} &                        ---\phantom{$^{\dagger\dagger}\pm$.000} \\
      & \glovest &                         .635\phantom{$^{\dagger\dagger}\pm$.000} &                      .277\phantom{$^{\dagger\dagger}\pm$.000} &                   .505\phantom{$^{\dagger\dagger}\pm$.000} &                   .495\phantom{$^{\dagger\dagger}\pm$.000} &                         .531\phantom{$^{\dagger\dagger}\pm$.000} &                        ---\phantom{$^{\dagger\dagger}\pm$.000} \\
     & \glovewcest &                         .678\phantom{$^{\dagger\dagger}\pm$.000} &                      .291\phantom{$^{\dagger\dagger}\pm$.000} &                   .629\phantom{$^{\dagger\dagger}\pm$.000} &                   .493\phantom{$^{\dagger\dagger}\pm$.000} &                         .556\phantom{$^{\dagger\dagger}\pm$.000} &                        ---\phantom{$^{\dagger\dagger}\pm$.000} \\\hline
     
\multirow{4}{*}{\fasttext} & \random &  .614\phantom{$^{\dagger\dagger}$}$\pm$.003 &  .312\phantom{$^{\dagger\dagger}$}$\pm$.006 &  .609\phantom{$^{\dagger\dagger}$}$\pm$.002 &  .523\phantom{$^{\dagger\dagger}$}$\pm$.001 &  .511\phantom{$^{\dagger\dagger}$}$\pm$.009 &  .513\phantom{$^{\dagger\dagger}$}$\pm$.002 \\
         & \glovetr &  .640\phantom{$^{\dagger\dagger}$}$\pm$.003 &  .330\phantom{$^{\dagger\dagger}$}$\pm$.005 &  .627\phantom{$^{\dagger\dagger}$}$\pm$.002 &  .548$^{\dagger\dagger}$$\pm$.001 &  .529\phantom{$^{\dagger\dagger}$}$\pm$.008 &  .547\phantom{$^{\dagger\dagger}$}$\pm$.003 \\
         & \glovern &  .640\phantom{$^{\dagger\dagger}$}$\pm$.003 &  .331\phantom{$^{\dagger\dagger}$}$\pm$.004 &  .624\phantom{$^{\dagger\dagger}$}$\pm$.002 &  \cellcolor{gray!70}.548$^{\dagger\dagger}$$\pm$.001 &  .528\phantom{$^{\dagger\dagger}$}$\pm$.008 &  .547\phantom{$^{\dagger\dagger}$}$\pm$.002 \\
         & \glovewcetr &  \cellcolor{gray!70}.693\phantom{$^{\dagger\dagger}$}$\pm$.002 &  \cellcolor{gray!70}.348\phantom{$^{\dagger\dagger}$}$\pm$.005 &  \cellcolor{gray!70}.655\phantom{$^{\dagger\dagger}$}$\pm$.002 &  .491\phantom{$^{\dagger\dagger}$}$\pm$.001 &  \cellcolor{gray!70}.569\phantom{$^{\dagger\dagger}$}$\pm$.008 &  \cellcolor{gray!70}\textbf{.571}\phantom{$^{\dagger\dagger}$}$\pm$.001 \\\hline

\multirow{6}{*}{CNN} & \random &  .636\phantom{$^{\dagger\dagger}$}$\pm$.005 &  .318\phantom{$^{\dagger\dagger}$}$\pm$.004 &  .666\phantom{$^{\dagger\dagger}$}$\pm$.004 &  .446\phantom{$^{\dagger\dagger}$}$\pm$.010 &  .518\phantom{$^{\dagger\dagger}$}$\pm$.011 &  .453\phantom{$^{\dagger\dagger}$}$\pm$.013 \\
         & \glovest &  .670\phantom{$^{\dagger\dagger}$}$\pm$.003 &  .318\phantom{$^{\dagger\dagger}$}$\pm$.003 &  .657\phantom{$^{\dagger\dagger}$}$\pm$.006 &  .491\phantom{$^{\dagger\dagger}$}$\pm$.010 &  .505\phantom{$^{\dagger\dagger}$}$\pm$.016 &  .444\phantom{$^{\dagger\dagger}$}$\pm$.013 \\
         & \glovetr &  .691\phantom{$^{\dagger\dagger}$}$\pm$.003 &  .323\phantom{$^{\dagger\dagger}$}$\pm$.005 &  .698\phantom{$^{\dagger\dagger}$}$\pm$.005 &  .515\phantom{$^{\dagger\dagger}$}$\pm$.015 &  .537\phantom{$^{\dagger\dagger}$}$\pm$.013 &  \cellcolor{gray!70}.479\phantom{$^{\dagger\dagger}$}$\pm$.014 \\
         & \glovern &  .696\phantom{$^{\dagger\dagger}$}$\pm$.002 &  .329\phantom{$^{\dagger\dagger}$}$\pm$.004 &  .692\phantom{$^{\dagger\dagger}$}$\pm$.005 &  .510\phantom{$^{\dagger\dagger}$}$\pm$.011 &  .535\phantom{$^{\dagger\dagger}$}$\pm$.013 &  .455\phantom{$^{\dagger\dagger}$}$\pm$.009 \\
         & \glovewcest &  .703$^{\dagger\dagger}$$\pm$.003 &  .331\phantom{$^{\dagger\dagger}$}$\pm$.007 &  .699\phantom{$^{\dagger\dagger}$}$\pm$.004 &  .515\phantom{$^{\dagger\dagger}$}$\pm$.007 &  .600$^{\dagger\dagger}$$\pm$.016 &  .391\phantom{$^{\dagger\dagger}$}$\pm$.007 \\
         & \glovewcetr &  \cellcolor{gray!70}\textbf{.706}\phantom{$^{\dagger\dagger}$}$\pm$.004 &  \cellcolor{gray!70}.346$^{\dagger\dagger}$$\pm$.003 &  \cellcolor{gray!70}\textbf{.706}\phantom{$^{\dagger\dagger}$}$\pm$.005 &  \cellcolor{gray!70}.523\phantom{$^{\dagger\dagger}$}$\pm$.008 &  \cellcolor{gray!70}.611\phantom{$^{\dagger\dagger}$}$\pm$.014 &  .424\phantom{$^{\dagger\dagger}$}$\pm$.013 \\\hline

\multirow{6}{*}{LSTM} & \random &  .426\phantom{$^{\dagger\dagger}$}$\pm$.014 &  .191\phantom{$^{\dagger\dagger}$}$\pm$.012 &  .567\phantom{$^{\dagger\dagger}$}$\pm$.011 &  .349\phantom{$^{\dagger\dagger}$}$\pm$.017 &  .371\phantom{$^{\dagger\dagger}$}$\pm$.023 &  .473\phantom{$^{\dagger\dagger}$}$\pm$.015 \\
         & \glovest &  .624\phantom{$^{\dagger\dagger}$}$\pm$.007 &  .213\phantom{$^{\dagger\dagger}$}$\pm$.007 &  .646\phantom{$^{\dagger\dagger}$}$\pm$.014 &  .509\phantom{$^{\dagger\dagger}$}$\pm$.013 &  .458\phantom{$^{\dagger\dagger}$}$\pm$.034 &  .444\phantom{$^{\dagger\dagger}$}$\pm$.116 \\
         & \glovetr &  .629\phantom{$^{\dagger\dagger}$}$\pm$.008 &  .185\phantom{$^{\dagger\dagger}$}$\pm$.009 &  .655\phantom{$^{\dagger\dagger}$}$\pm$.025 &  .398\phantom{$^{\dagger\dagger}$}$\pm$.140 &  .469\phantom{$^{\dagger\dagger}$}$\pm$.040 &  .512\phantom{$^{\dagger\dagger}$}$\pm$.013 \\
         & \glovern &  .622\phantom{$^{\dagger\dagger}$}$\pm$.007 &  .184\phantom{$^{\dagger\dagger}$}$\pm$.007 &  .662\phantom{$^{\dagger\dagger}$}$\pm$.005 &  .460\phantom{$^{\dagger\dagger}$}$\pm$.021 &  .461\phantom{$^{\dagger\dagger}$}$\pm$.029 &  .524\phantom{$^{\dagger\dagger}$}$\pm$.010 \\
         & \glovewcest &  .652\phantom{$^{\dagger\dagger}$}$\pm$.047 &  \cellcolor{gray!70}.273\phantom{$^{\dagger\dagger}$}$\pm$.008 &  \cellcolor{gray!70}.688\phantom{$^{\dagger\dagger}$}$\pm$.013 &  \cellcolor{gray!70}.555\phantom{$^{\dagger\dagger}$}$\pm$.016 &  \cellcolor{gray!70}.569\phantom{$^{\dagger\dagger}$}$\pm$.018 &  .505\phantom{$^{\dagger\dagger}$}$\pm$.009 \\
         & \glovewcetr &  \cellcolor{gray!70}.660\phantom{$^{\dagger\dagger}$}$\pm$.010 &  .247\phantom{$^{\dagger\dagger}$}$\pm$.007 &  .684\phantom{$^{\dagger\dagger}$}$\pm$.013 &  .500\phantom{$^{\dagger\dagger}$}$\pm$.014 &  .532\phantom{$^{\dagger\dagger}$}$\pm$.057 &  \cellcolor{gray!70}.534\phantom{$^{\dagger\dagger}$}$\pm$.008 \\\hline

\multirow{6}{*}{ATTN} & \random &  .554\phantom{$^{\dagger\dagger}$}$\pm$.004 &  .204\phantom{$^{\dagger\dagger}$}$\pm$.071 &  .572\phantom{$^{\dagger\dagger}$}$\pm$.004 &  .367\phantom{$^{\dagger\dagger}$}$\pm$.011 &  .467\phantom{$^{\dagger\dagger}$}$\pm$.016 &  .491\phantom{$^{\dagger\dagger}$}$\pm$.011 \\
         & \glovest &  .624\phantom{$^{\dagger\dagger}$}$\pm$.004 &  .249\phantom{$^{\dagger\dagger}$}$\pm$.006 &  .649\phantom{$^{\dagger\dagger}$}$\pm$.005 &  \cellcolor{gray!70}.546$^{\dagger\dagger}$$\pm$.006 &  .524\phantom{$^{\dagger\dagger}$}$\pm$.021 &  .498\phantom{$^{\dagger\dagger}$}$\pm$.028 \\
         & \glovetr &  .626\phantom{$^{\dagger\dagger}$}$\pm$.005 &  .240\phantom{$^{\dagger\dagger}$}$\pm$.006 &  .643\phantom{$^{\dagger\dagger}$}$\pm$.012 &  .468\phantom{$^{\dagger\dagger}$}$\pm$.018 &  .495\phantom{$^{\dagger\dagger}$}$\pm$.052 &  .527\phantom{$^{\dagger\dagger}$}$\pm$.020 \\
         & \glovern &  .622\phantom{$^{\dagger\dagger}$}$\pm$.007 &  .240\phantom{$^{\dagger\dagger}$}$\pm$.007 &  .644\phantom{$^{\dagger\dagger}$}$\pm$.007 &  .471\phantom{$^{\dagger\dagger}$}$\pm$.009 &  .504\phantom{$^{\dagger\dagger}$}$\pm$.021 &  .530\phantom{$^{\dagger\dagger}$}$\pm$.019 \\
         & \glovewcest &  \cellcolor{gray!70}.682\phantom{$^{\dagger\dagger}$}$\pm$.004 &  \cellcolor{gray!70}.310\phantom{$^{\dagger\dagger}$}$\pm$.006 &  \cellcolor{gray!70}.681\phantom{$^{\dagger\dagger}$}$\pm$.003 &  .508\phantom{$^{\dagger\dagger}$}$\pm$.010 &  \cellcolor{gray!70}.594$^{\dagger}$\phantom{$^{\dagger}$}$\pm$.020 &  .512\phantom{$^{\dagger\dagger}$}$\pm$.013 \\
         & \glovewcetr &  .659\phantom{$^{\dagger\dagger}$}$\pm$.005 &  .283\phantom{$^{\dagger\dagger}$}$\pm$.009 &  .676\phantom{$^{\dagger\dagger}$}$\pm$.004 &  .527\phantom{$^{\dagger\dagger}$}$\pm$.014 &  .566\phantom{$^{\dagger\dagger}$}$\pm$.016 &  \cellcolor{gray!70}.533\phantom{$^{\dagger\dagger}$}$\pm$.008 \\\hline

\end{tabular}
}
\caption{Classification performance in terms of $F_1^M$. Boldface indicates the best absolute result
for each dataset, and greyed-out cells indicate the best result
locally to a specific neural architecture. Symbols $\dag$ and
$\dag\dag$ indicate the methods, if any, whose performance is
\emph{not} statistically significantly different with respect to the
best result obtained by any neural approach according to a two-tailed
t-test at confidence level $\alpha=0.05$ and $\alpha=0.005$,
respectively.}
\label{tab:final-te-macro-F1}
\end{table}

% ---------------------------------------------
% micro-F1
%----------------------------------------------
\begin{table}[h]
\centering
\resizebox{\textwidth}{!}{%
\begin{tabular}{|l|l||r|r|r|r|r|r|}
\hline
model     & variant &              \centercell{\rotatebox{90}{\textsc{20Newsgroups}\hspace{1ex}\mbox{}}} &                 \centercell{\rotatebox{90}{\textsc{JRC-Acquis}}} &                    \centercell{\rotatebox{90}{\textsc{Ohsumed}}} &                    \centercell{\rotatebox{90}{\textsc{RCV1-v2}}} &              \centercell{\rotatebox{90}{\textsc{Reuters-21578}\hspace{1ex}\mbox{}}} &                 \centercell{\rotatebox{90}{\textsc{WIPO-gamma}\hspace{1ex}\mbox{}}} \\
\hline\hline
\multirow{3}{*}{SVM} & tfidf &                         .679\phantom{$^{\dagger\dagger}\pm$.000} &                      .525\phantom{$^{\dagger\dagger}\pm$.000} &                   .702\phantom{$^{\dagger\dagger}\pm$.000} &                   \textbf{.808}\phantom{$^{\dagger\dagger}\pm$.000} &                \textbf{.881}\phantom{$^{\dagger\dagger}\pm$.000} &                        ---\phantom{$^{\dagger\dagger}\pm$.000} \\
%     & WCEs &                         .689\phantom{$^{\dagger\dagger}\pm$.000} &                      .255\phantom{$^{\dagger\dagger}\pm$.000} &                   .646\phantom{$^{\dagger\dagger}\pm$.000} &                   .702\phantom{$^{\dagger\dagger}\pm$.000} &                         .795\phantom{$^{\dagger\dagger}\pm$.000} &                        ---\phantom{$^{\dagger\dagger}\pm$.000} \\ 
          & GloVe (static) &                         .649\phantom{$^{\dagger\dagger}\pm$.000} &                      .404\phantom{$^{\dagger\dagger}\pm$.000} &                   .544\phantom{$^{\dagger\dagger}\pm$.000} &                   .702\phantom{$^{\dagger\dagger}\pm$.000} &                         .791\phantom{$^{\dagger\dagger}\pm$.000} &                        ---\phantom{$^{\dagger\dagger}\pm$.000} \\
     & GloVe+WCEs (static) &                         .688\phantom{$^{\dagger\dagger}\pm$.000} &                      .385\phantom{$^{\dagger\dagger}\pm$.000} &                   .658\phantom{$^{\dagger\dagger}\pm$.000} &                   .757\phantom{$^{\dagger\dagger}\pm$.000} &                         .834\phantom{$^{\dagger\dagger}\pm$.000} &                        ---\phantom{$^{\dagger\dagger}\pm$.000} \\
\hline

\multirow{4}{*}{\fasttext} & \random &  .622\phantom{$^{\dagger\dagger}$}$\pm$.003 &  .511\phantom{$^{\dagger\dagger}$}$\pm$.002 &  .644\phantom{$^{\dagger\dagger}$}$\pm$.002 &  .772\phantom{$^{\dagger\dagger}$}$\pm$.001 &  .822\phantom{$^{\dagger\dagger}$}$\pm$.004 &  .689\phantom{$^{\dagger\dagger}$}$\pm$.001 \\
         & \glovetr &  .649\phantom{$^{\dagger\dagger}$}$\pm$.003 &  .530\phantom{$^{\dagger\dagger}$}$\pm$.002 &  .662\phantom{$^{\dagger\dagger}$}$\pm$.002 &  .781\phantom{$^{\dagger\dagger}$}$\pm$.001 &  .839\phantom{$^{\dagger\dagger}$}$\pm$.002 &  .696\phantom{$^{\dagger\dagger}$}$\pm$.000 \\
         & \glovern &  .649\phantom{$^{\dagger\dagger}$}$\pm$.003 &  .530\phantom{$^{\dagger\dagger}$}$\pm$.002 &  .651\phantom{$^{\dagger\dagger}$}$\pm$.001 &  \cellcolor{gray!70}.781\phantom{$^{\dagger\dagger}$}$\pm$.001 &  .839\phantom{$^{\dagger\dagger}$}$\pm$.003 &  .696\phantom{$^{\dagger\dagger}$}$\pm$.000 \\
         & \glovewcetr &  \cellcolor{gray!70}.704\phantom{$^{\dagger\dagger}$}$\pm$.002 &  \cellcolor{gray!70}\textbf{.536}\phantom{$^{\dagger\dagger}$}$\pm$.004 &  \cellcolor{gray!70}.688\phantom{$^{\dagger\dagger}$}$\pm$.001 &  .769\phantom{$^{\dagger\dagger}$}$\pm$.001 &  \cellcolor{gray!70}.843\phantom{$^{\dagger\dagger}$}$\pm$.003 &  \cellcolor{gray!70}.701$^{\dagger}$\phantom{$^{\dagger}$}$\pm$.000 \\\hline

\multirow{6}{*}{CNN} & \random &  .644\phantom{$^{\dagger\dagger}$}$\pm$.005 &  .509\phantom{$^{\dagger\dagger}$}$\pm$.007 &  .691\phantom{$^{\dagger\dagger}$}$\pm$.005 &  .746\phantom{$^{\dagger\dagger}$}$\pm$.006 &  .835\phantom{$^{\dagger\dagger}$}$\pm$.004 &  .673\phantom{$^{\dagger\dagger}$}$\pm$.004 \\
         & \glovest &  .684\phantom{$^{\dagger\dagger}$}$\pm$.002 &  .527\phantom{$^{\dagger\dagger}$}$\pm$.004 &  .693\phantom{$^{\dagger\dagger}$}$\pm$.006 &  .799$^{\dagger}$\phantom{$^{\dagger}$}$\pm$.004 &  .848\phantom{$^{\dagger\dagger}$}$\pm$.005 &  .649\phantom{$^{\dagger\dagger}$}$\pm$.005 \\
         & \glovetr &  .703\phantom{$^{\dagger\dagger}$}$\pm$.003 &  .528\phantom{$^{\dagger\dagger}$}$\pm$.005 &  .720\phantom{$^{\dagger\dagger}$}$\pm$.007 &  \cellcolor{gray!70}.803$^{\dagger\dagger}$$\pm$.009 &  .857\phantom{$^{\dagger\dagger}$}$\pm$.003 &  \cellcolor{gray!70}.675\phantom{$^{\dagger\dagger}$}$\pm$.005 \\
         & \glovern &  .707\phantom{$^{\dagger\dagger}$}$\pm$.002 &  \cellcolor{gray!70}.529$^{\dagger}$\phantom{$^{\dagger}$}$\pm$.008 &  .717\phantom{$^{\dagger\dagger}$}$\pm$.006 &  .796\phantom{$^{\dagger\dagger}$}$\pm$.005 &  .851\phantom{$^{\dagger\dagger}$}$\pm$.003 &  .666\phantom{$^{\dagger\dagger}$}$\pm$.003 \\
         & \glovewcest &  .715$^{\dagger\dagger}$$\pm$.004 &  .509\phantom{$^{\dagger\dagger}$}$\pm$.007 &  .723\phantom{$^{\dagger\dagger}$}$\pm$.003 &  .789\phantom{$^{\dagger\dagger}$}$\pm$.002 &  .858\phantom{$^{\dagger\dagger}$}$\pm$.004 &  .623\phantom{$^{\dagger\dagger}$}$\pm$.003 \\
         & \glovewcetr &  \cellcolor{gray!70}\textbf{.717}\phantom{$^{\dagger\dagger}$}$\pm$.003 &  .520\phantom{$^{\dagger\dagger}$}$\pm$.004 &  \cellcolor{gray!70}\textbf{.729}\phantom{$^{\dagger\dagger}$}$\pm$.004 &  .792\phantom{$^{\dagger\dagger}$}$\pm$.002 &  \cellcolor{gray!70}.866\phantom{$^{\dagger\dagger}$}$\pm$.005 &  .649\phantom{$^{\dagger\dagger}$}$\pm$.005 \\\hline

\multirow{6}{*}{LSTM} & \random &  .434\phantom{$^{\dagger\dagger}$}$\pm$.013 &  .226\phantom{$^{\dagger\dagger}$}$\pm$.094 &  .618\phantom{$^{\dagger\dagger}$}$\pm$.009 &  .680\phantom{$^{\dagger\dagger}$}$\pm$.015 &  .699\phantom{$^{\dagger\dagger}$}$\pm$.024 &  .685\phantom{$^{\dagger\dagger}$}$\pm$.005 \\
         & \glovest &  .633\phantom{$^{\dagger\dagger}$}$\pm$.007 &  .360\phantom{$^{\dagger\dagger}$}$\pm$.015 &  .687\phantom{$^{\dagger\dagger}$}$\pm$.011 &  .793\phantom{$^{\dagger\dagger}$}$\pm$.005 &  .823\phantom{$^{\dagger\dagger}$}$\pm$.011 &  .662$^{\dagger}$\phantom{$^{\dagger}$}$\pm$.063 \\
         & \glovetr &  .638\phantom{$^{\dagger\dagger}$}$\pm$.008 &  .193\phantom{$^{\dagger\dagger}$}$\pm$.063 &  .686\phantom{$^{\dagger\dagger}$}$\pm$.020 &  .715$^{\dagger}$\phantom{$^{\dagger}$}$\pm$.100 &  .804\phantom{$^{\dagger\dagger}$}$\pm$.018 &  .701$^{\dagger}$\phantom{$^{\dagger}$}$\pm$.004 \\
         & \glovern &  .631\phantom{$^{\dagger\dagger}$}$\pm$.007 &  .189\phantom{$^{\dagger\dagger}$}$\pm$.068 &  .688\phantom{$^{\dagger\dagger}$}$\pm$.003 &  .751\phantom{$^{\dagger\dagger}$}$\pm$.012 &  .796\phantom{$^{\dagger\dagger}$}$\pm$.010 &  \cellcolor{gray!70}.703$^{\dagger\dagger}$$\pm$.003 \\
         & \glovewcest &  .661\phantom{$^{\dagger\dagger}$}$\pm$.044 &  \cellcolor{gray!70}.449\phantom{$^{\dagger\dagger}$}$\pm$.017 &  \cellcolor{gray!70}.714\phantom{$^{\dagger\dagger}$}$\pm$.009 &  \cellcolor{gray!70}.807\phantom{$^{\dagger\dagger}$}$\pm$.007 &  \cellcolor{gray!70}.860$^{\dagger}$\phantom{$^{\dagger}$}$\pm$.006 &  .686\phantom{$^{\dagger\dagger}$}$\pm$.004 \\
         & \glovewcetr &  \cellcolor{gray!70}.668\phantom{$^{\dagger\dagger}$}$\pm$.010 &  .403\phantom{$^{\dagger\dagger}$}$\pm$.009 &  .709\phantom{$^{\dagger\dagger}$}$\pm$.011 &  .765\phantom{$^{\dagger\dagger}$}$\pm$.008 &  .849\phantom{$^{\dagger\dagger}$}$\pm$.007 &  .703$^{\dagger\dagger}$$\pm$.003 \\\hline

\multirow{6}{*}{ATTN} & \random &  .559\phantom{$^{\dagger\dagger}$}$\pm$.004 &  .336\phantom{$^{\dagger\dagger}$}$\pm$.112 &  .620\phantom{$^{\dagger\dagger}$}$\pm$.006 &  .677\phantom{$^{\dagger\dagger}$}$\pm$.008 &  .775\phantom{$^{\dagger\dagger}$}$\pm$.009 &  .690\phantom{$^{\dagger\dagger}$}$\pm$.005 \\
         & \glovest &  .634\phantom{$^{\dagger\dagger}$}$\pm$.003 &  .421\phantom{$^{\dagger\dagger}$}$\pm$.009 &  .680\phantom{$^{\dagger\dagger}$}$\pm$.006 &  \cellcolor{gray!70}.797\phantom{$^{\dagger\dagger}$}$\pm$.002 &  .818\phantom{$^{\dagger\dagger}$}$\pm$.008 &  .688\phantom{$^{\dagger\dagger}$}$\pm$.012 \\
         & \glovetr &  .634\phantom{$^{\dagger\dagger}$}$\pm$.007 &  .397\phantom{$^{\dagger\dagger}$}$\pm$.013 &  .667\phantom{$^{\dagger\dagger}$}$\pm$.009 &  .744\phantom{$^{\dagger\dagger}$}$\pm$.008 &  .790\phantom{$^{\dagger\dagger}$}$\pm$.014 &  \cellcolor{gray!70}\textbf{.707}\phantom{$^{\dagger\dagger}$}$\pm$.007 \\
         & \glovern &  .631\phantom{$^{\dagger\dagger}$}$\pm$.006 &  .396\phantom{$^{\dagger\dagger}$}$\pm$.012 &  .668\phantom{$^{\dagger\dagger}$}$\pm$.007 &  .745\phantom{$^{\dagger\dagger}$}$\pm$.006 &  .798\phantom{$^{\dagger\dagger}$}$\pm$.009 &  .707$^{\dagger\dagger}$$\pm$.006 \\
         & \glovewcest &  \cellcolor{gray!70}.691\phantom{$^{\dagger\dagger}$}$\pm$.004 &  \cellcolor{gray!70}.489\phantom{$^{\dagger\dagger}$}$\pm$.007 &  \cellcolor{gray!70}.709\phantom{$^{\dagger\dagger}$}$\pm$.003 &  .783\phantom{$^{\dagger\dagger}$}$\pm$.006 &  \cellcolor{gray!70}.858\phantom{$^{\dagger\dagger}$}$\pm$.006 &  .687\phantom{$^{\dagger\dagger}$}$\pm$.006 \\
         & \glovewcetr &  .668\phantom{$^{\dagger\dagger}$}$\pm$.005 &  .459\phantom{$^{\dagger\dagger}$}$\pm$.013 &  .698\phantom{$^{\dagger\dagger}$}$\pm$.005 &  .773\phantom{$^{\dagger\dagger}$}$\pm$.008 &  .841\phantom{$^{\dagger\dagger}$}$\pm$.004 &  .703$^{\dagger\dagger}$$\pm$.002 \\\hline

\end{tabular}
}
\caption{Classification performance in terms of $F_1^\mu$.}
\label{tab:final-te-micro-F1}
\end{table}
% \end{minipage}

Various facts emerge from these results.  First, beating
well-optimized ``traditional'' baselines, such as \svmtfidf, is not
easy (especially in terms of $F_{1}^{M}$).  This has already been
noticed in past literature~\cite{wang2012baselines}, and has recently
stimulated debate~\cite{lin2019neural,yang2019critically}.  For SVMs,
our results for \textsc{20Newsgroups} and \textsc{Reuters-21578} are
in line with those reported for the same datasets by
\cite{Bekkerman03}. These authors found that the adoption of more
sophisticated supervised representations helps SVMs to improve over
simple tfidf features in \textsc{20Newsgroups} but not on
\textsc{Reuters-21578},
% (and WebKB -- a dataset we have not included in our experiments),
and argued that in this latter case a considerably high accuracy is
achievable by simply using a handful of highly correlated terms for
each class (something that is already well represented in a
bag-of-words model).
% The addition of supervised embeddings in \textsc{20Newsgroups}
% helped SVM to improve over SVM-tfidf.  This is in line with the
% findings of \cite{Bekkerman03}
Notwithstanding this, we should observe that SVMs, in their standard
formulation, do not scale to very large training sets; as a result, we
were unable to train them on \textsc{WIPO-gamma}.
% \alexcomment{We could report results for online SGD with Hinge
% loss...}

Concerning neural approaches, a method equipped with WCEs either turns
out to be the best performer, or is comparable (in a statistically
significant sense) to the best performer, both in terms of $F_{1}^{M}$
and $F_{1}^{\mu}$, and for all datasets.
% Locally to each network architecture, the variants using supervised
% embeddings improve, with few exceptions, over the rest.
Concatenating \glove\ embeddings and WCEs almost always yields
superior performance with respect to only using \glove\ embeddings,
both for static and trainable pre-trained embeddings, across all
models (this applies also to SVMs).  On average across all six
datasets and neural models, \glovewcest\ shows a relative improvement
of
% +8.54\% and +4.51\% <-- without fasttext
+7.42\% and +3.30\% over \glovest\ in terms of $F_{1}^{M}$ and
$F_{1}^{\mu}$, respectively.  In a similar way, the relative
improvement of \glovewcetr\ with respect to \glovetr\ amounts to
% +7.32\% and +7.40\%
+6.66\% and +4.42\%
% \alexcomment{(added fasttext in the calculations)}
on average in terms of $F_{1}^{M}$ and $F_{1}^{\mu}$, respectively.
In all cases except $F_{1}^{\mu}$ for the static case, the differences
in performance, as averaged across methods and datasets, are
statistically significant at $\alpha=0.05$.
% \alexcomment{Quel +3.30\% non è significativo...}

The superiority of the models also equipped with WCEs cannot be merely
explained by the higher number of parameters in their embedding
layer. By inspecting the execution logs, we found out that
approximately 50\% of the times the best hyperparameters chosen by the
variants with WCEs were consistent with those chosen by the same
variant without them. In such cases, the \glovewcetr\ setting contains
exactly the same number of trainable parameters as the corresponding
\glovern\ variants, yet it performs better (with relative average
improvements of
% +5.80\% and +8.70\%
+6.19\% and +4.40\% in $F_{1}^{M}$ and $F_{1}^{\mu}$, respectively,
and with statistical significance at $\alpha=0.05$). For other
approximately 40\% of the times, the model selected when using WCEs
happened to be comparatively smaller than when not using them, and
only in the remaining 10\% of the cases the variants using WCEs turned
out to be larger.

% The variants involving WCEs generally outperform \fasttext\ across
% most datasets, and for both evaluation measures.
Fine-tuning embeddings (\glovetr\ and \glovewcetr) has proven
consistently superior to keeping them static (\glovest\ and
\glovewcest) when using the CNN
architecture, %; similar patterns do not emerge for
but not necessarily for LSTM or ATTN.

% justify the improvement comes not just because of the increase in
% the number of parameters, since some baselines (e.g., \glove\ w/o
% supervised) sometimes takes (according to the hyperparam selection)
% a configuration that has many more parameters

% Say we have not try to counter the imbalance problem (many methods
% to do so exist --over/under-sampling, sampling selection, class
% weight balance, etc.-- that go beyond the scope of this research.

% ----------------------------------------------------------------------

\subsection{Learning Curves}
\label{sec:learningcurves}

\noindent In this section we look at the learning curves for the three
deep learning architectures when equipped with different types of
embeddings. For the sake of clarity, we choose to plot only three
representative variants: \random, \glovest\ and \glovewcest\ (that,
for simplicity, we here simply denote by \glove\ and \glovewce).
Unless specified differently, these plots and all the subsequent ones
are generated with the same, fixed hyperparameters for all variants,
i.e., 250 channels for CNN and 512 hidden nodes for LSTM and ATTN, a
supervised dropout probability of 0.5 for \glovewce, and 200
dimensions for random embeddings; we run 100 training epochs and
deactivate early stop.  For these experiments we choose
\textsc{RCV1-v2}, since it is arguably the most widely adopted
benchmark in the literature of text classification by topic. (In
similar experiments that we have run on other datasets we have
verified similar trends.)

Figure~\ref{fig:trend:rcv1} shows a grid of plots that visualize
learning curves as a function of the number of epochs for
\textsc{RCV1-v2}. The first and second rows display (the logarithm of)
the training and validation loss, respectively, while the third and
fourth rows display the values of $F_{1}^{M}$ and $F_{1}^{\mu}$,
respectively, on the validation set.  Columns correspond, from left to
right, to the CNN, LSTM, and ATTN architectures.
% \blue{We here show the evolution of the evaluation metrics as
% tracked in the validation set %(instead of in the test set)
% in order to display some information which is available for a
% practitioner endeavoured to deploy a neural classifier.
%% simulate a realistic scenario in which a practitioner endeavours to
%% train a neural model.  information which is accessible to the
%% practitioner in real scenarios.
% } \fsebcomment{Non ho capito ...}

There are a few observations that we can make from these plots.
Networks with random embeddings have more parameters to tune than
networks which rely on static pre-trained embeddings (since the latter
are fixed, and thus are not trainable parameters), and thus lower the
training loss faster than the rest.  Notwithstanding this, the
validation loss is always higher for them than that of \glove\ and
\glovewce, which shows that the presence of random embeddings brings
about a tendency to overfit the training data.  This tendency to
generate overfitting is well countered by the variants that use static
pre-trained embeddings.  The knowledge incorporated in \glove\
embeddings is generic and thus consistent for validation documents as
well.

\def \plotdir {training_trend_plots} \def \captiontext {Learning
curves on dataset } \def \displayeval {va}

% \def \datasetplot {\textsc{20Newsgroups}} \def \datasetname
% {\textsc{20Newsgroups}} \input{plotgrid}

% \def \datasetplot {jrcall} \def \datasetname {\textsc{JRC-Acquis}}
% \input{plotgrid}

% \def \datasetplot {\textsc{Ohsumed}} \def \datasetname
% {\textsc{Ohsumed}} \input{plotgrid}

\def \datasetplot {rcv1} \def \datasetname {\textsc{RCV1-v2}} \def
\plotlabel {fig:trend:rcv1} \plot{\begin{figure}[h!]
\centering

\begin{minipage}{.08\textwidth}
\center \scriptsize 
\end{minipage}
\begin{minipage}{.29\textwidth}
\center \scriptsize \hspace{1.6cm} CNN
\end{minipage}
\begin{minipage}{.29\textwidth}
\center \scriptsize \hspace{1.6cm} LSTM
\end{minipage}
\begin{minipage}{.29\textwidth}
\center \scriptsize \hspace{1.6cm} ATTN
\end{minipage}

\begin{minipage}{.08\textwidth}
\center \scriptsize \rotatebox{90}{Training error}
\end{minipage}
\begin{minipage}{.29\textwidth}
  \includegraphics[width=\linewidth]{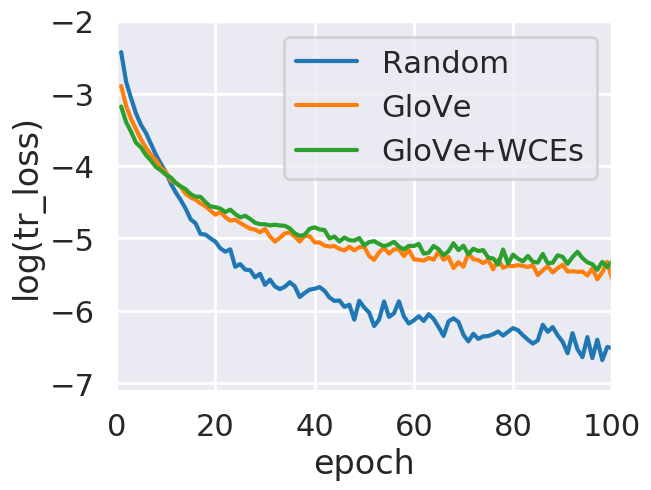}
\end{minipage}
\begin{minipage}{.29\textwidth}
  \includegraphics[width=\linewidth]{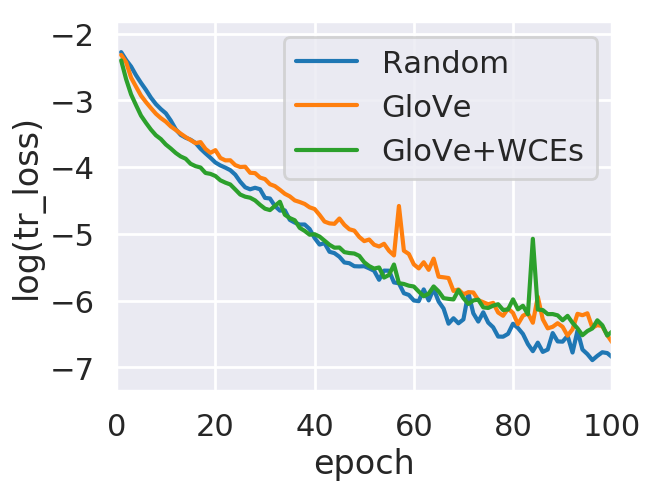}
\end{minipage}
\begin{minipage}{.29\textwidth}
  \includegraphics[width=\linewidth]{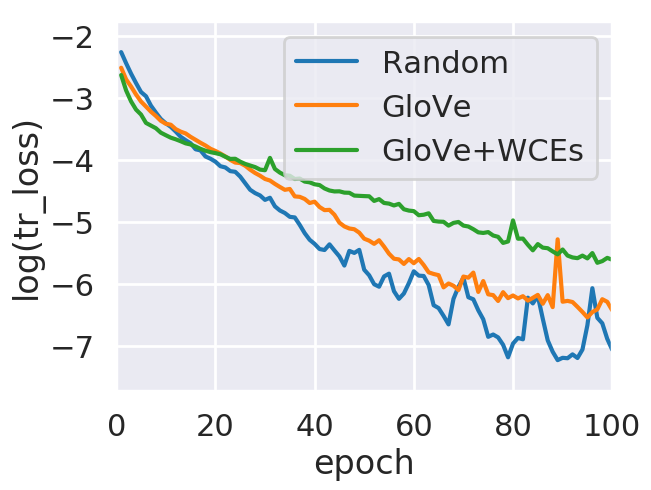}
\end{minipage}

\begin{minipage}{.08\textwidth}
\center \scriptsize \rotatebox{90}{Validation error}
\end{minipage}
\begin{minipage}{.29\textwidth}
  \includegraphics[width=\linewidth]{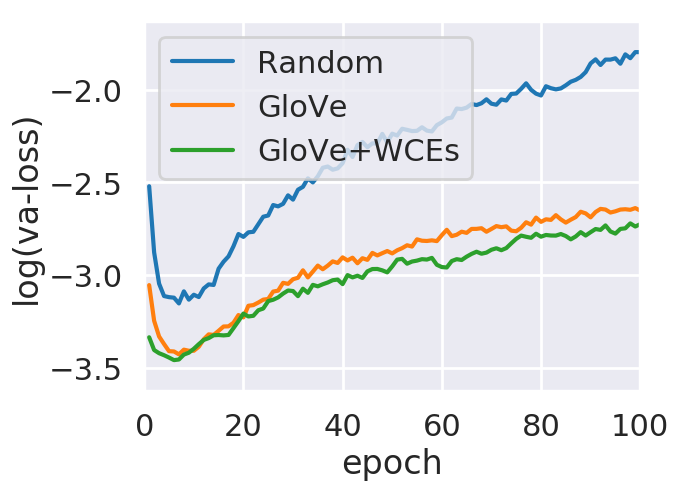}
\end{minipage}
\begin{minipage}{.29\textwidth}
  \includegraphics[width=\linewidth]{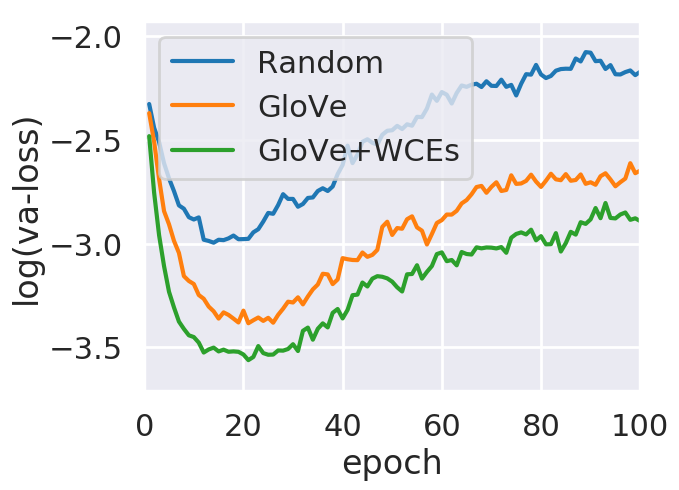}
\end{minipage}
\begin{minipage}{.29\textwidth}
  \includegraphics[width=\linewidth]{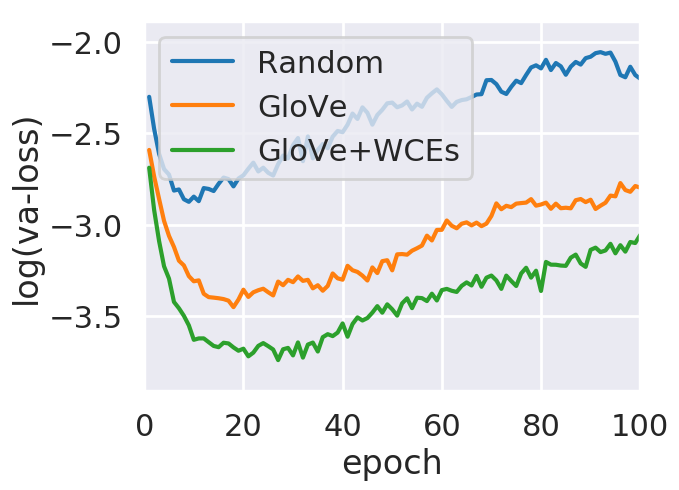}
\end{minipage}

\begin{minipage}{.08\textwidth}
\center \scriptsize \rotatebox{90}{$F_1^M$(\displayeval)}
\end{minipage}
\begin{minipage}{.29\textwidth}
  \includegraphics[width=\linewidth]{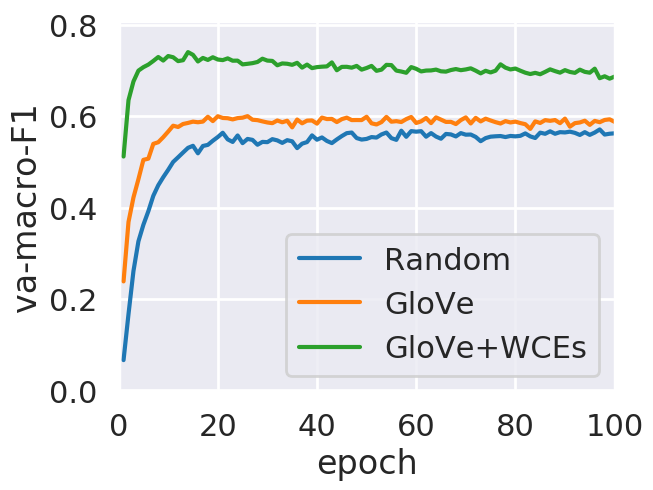}
\end{minipage}
\begin{minipage}{.29\textwidth}
  \includegraphics[width=\linewidth]{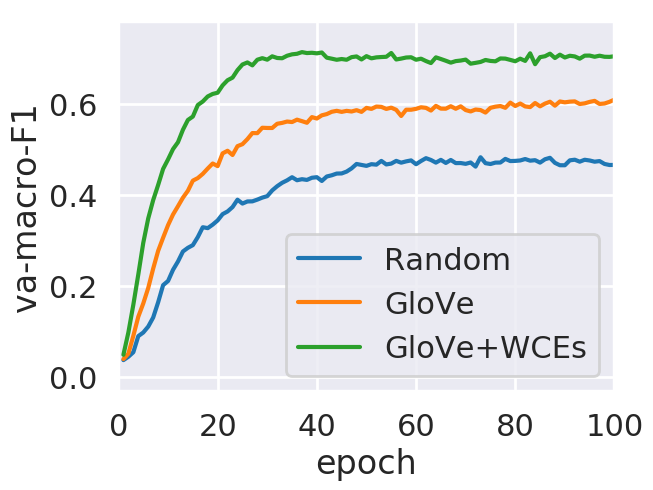}
\end{minipage}
\begin{minipage}{.29\textwidth}
  \includegraphics[width=\linewidth]{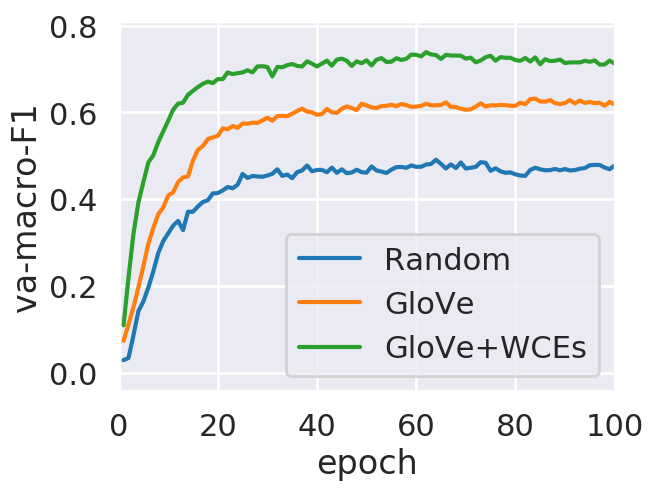}
\end{minipage}

\begin{minipage}{.08\textwidth}
\center \scriptsize \rotatebox{90}{$F_1^\mu$(\displayeval)}
\end{minipage}
\begin{minipage}{.29\textwidth}
  \includegraphics[width=\linewidth]{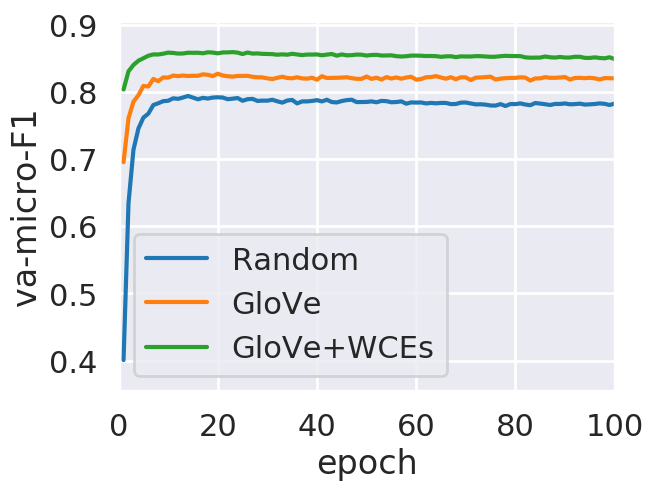}
\end{minipage}
\begin{minipage}{.29\textwidth}
  \includegraphics[width=\linewidth]{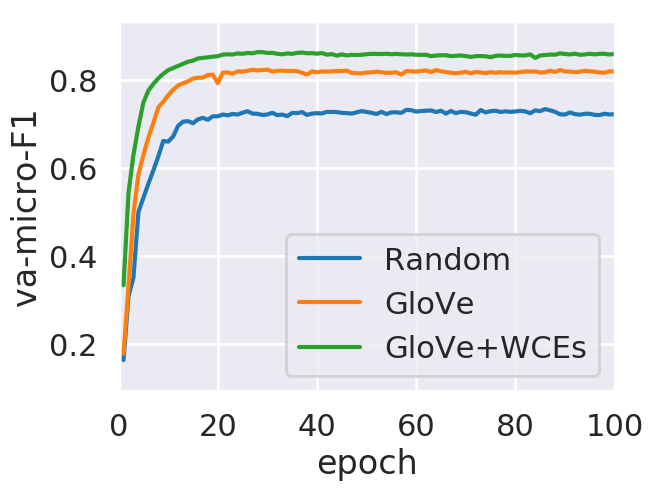}
\end{minipage}
\begin{minipage}{.29\textwidth}
  \includegraphics[width=\linewidth]{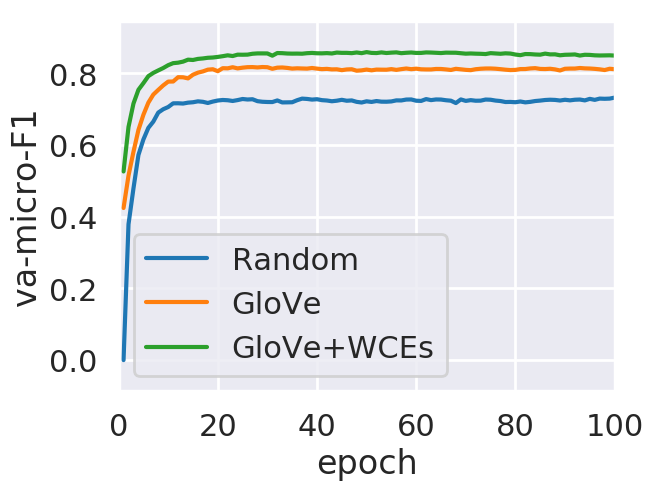}
\end{minipage}

\caption{\captiontext \datasetname}
\label{\plotlabel}

\end{figure}}

% \def \datasetplot {reuters21578} \def \datasetname
% {\textsc{Reuters-21578}} \input{plotgrid}

% \def \datasetplot {wipo-sl-sc} \def \datasetname
% {\textsc{WIPO-gamma}} \input{plotgrid}

We can also observe that the use of WCEs eases parameter optimization
across the three architectures tested.  That is, models equipped with
WCEs reach promising regions of the parameter space faster (i.e., in
fewer epochs) than those not using them.

However, since models relying exclusively on pre-trained embeddings
are anyway connected to class labels through the loss function, it is
legitimate to wonder what is the real contribution of WCEs to the
supervised learning task.  Clearly, WCEs are not bringing any new
information to the model, since they are computed using the very same
amount of information the learner has when training a classifier (this
is in contrast to pre-trained embeddings, which are learnt from
external data).
% \fsebcomment{Non mi sembra un buon argomento, sappiamo tutti che
% diversi learners sono diversamente bravi nello sfruttare la medesima
% informazione, e.g., SVMs e NB.}  But, what is the contribution of
% WCEs to the supervised learning task?  We believe the WCEs establish
% a bias that
We conjecture that this has to do with the way WCEs inject supervised
information from the bottom (word level) and the top (document level),
instead of only from the top, which may favour the gradient flow.
% The gradient flow undergoes many dificulties through the learning
% process (e.g., gradient vanishing or explosion) that might be
% alleviated this way.
In other words, WCEs may not really be adding any new information, but
are handling the available class label information in a more efficient
way.
% \fsebcomment{Sì, ma non abbiamo dato una risposta alla domanda che
% ci siamo posti ...}

It is worth noting that WCEs model the correlations between words and
labels globally (at the dataset level) and not locally (at the batch
level).  Global word-class dependencies remain reachable to batched
optimization only in the long term.  The same principle, according to
which the word-class distributions are mined globally beforehand, and
later serve the purpose of a model prior, was already used
in~\cite{moreo2019learning} (more details are given in Section
\ref{sec:connections}).

% ----------------------------------------------------------------------

\subsection{The Importance of Regularization}
\label{sec:regularization}

\noindent WCEs directly inject into the model information from the
label distribution as available in the training set.  This might
somehow compromise the generalization capability of the classifier
when dealing with future unseen data, for the reasons discussed at the
beginning of Section~\ref{sec:method:regularization}.  During
preliminary experiments we observed that this is indeed the case, and
that there is thus a need for properly regularizing the model.

Figure~\ref{fig:reg:rcv1} shows the effect of supervised dropout
(which, as explained in Section~\ref{sec:method:regularization}, is
the device we use performing regularization) at varying drop
probability rates for the \glovewce\ configuration. In the last two
rows of this figure we plot $F_{1}^{M}$ and $F_{1}^{\mu}$ as directly
computed on the test set of \textsc{RCV1-v2} every 10 epochs (last two
columns).  The rationale behind showing the values of effectiveness on
the test set (instead of on the validation set, as in
Figure~\ref{fig:trend:rcv1}) is to better illustrate the effect of
regularization when dealing with unseen data: as observed in
Section~\ref{sec:datasets}, news stories in the RCV1-v2 test set are
from a disjoint time window with respect to those in the training set,
which is not true for validation documents, which are randomly drawn
from the original training set.

We can observe that applying no regularization at all ($p=0.0$) yields
a faster minimization of the training loss, but also results in poorer
generalization on unseen data.  The model generalizes better for
$p>0.0$; setting $p=0.5$ actually yields the best results.  Of course,
in general the optimal value for $p$ has to be explored on a
validation set in a case-by-case fashion, but we have observed the
setting $p=0.5$ to generally entail, across all methods and datasets,
a good tradeoff between loss minimization in training and performance
on validation/test data (this is indeed in agreement with general
considerations regarding dropout as reported in the literature).

% Indeed, setting $p=0.5$ generally entailed a good trade-off between
% loss minimization in training and performance in validation/test
% across all methods and datasets (omitted for brevity). Though the
% optimal value is to be explored in a validation set, $p=0.5$
% represents (in line with common practices) a reasonable default
% choice for supervised
% dropout. % (this is in agreement with general considerations regarding dropout in literature).

\def \plotdir {regularization} \def \captiontext {Effect of supervised
dropout regularization in } \def \displayeval {te}

% \def \datasetplot {\textsc{20Newsgroups}} \def \datasetname
% {\textsc{20Newsgroups}} \input{plotgrid}

% \def \datasetplot {jrcall} \def \datasetname {\textsc{JRC-Acquis}}
% \input{plotgrid}

% \def \datasetplot {\textsc{Ohsumed}} \def \datasetname
% {\textsc{Ohsumed}} \input{plotgrid}

\def \datasetplot {rcv1} \def \datasetname {\textsc{RCV1-v2}} \def
\plotlabel {fig:reg:rcv1} \plot{}
% \input{plotgrid}

% \def \displayeval {va} \input{plotgrid}

% \def \datasetplot {reuters21578} \def \datasetname
% {\textsc{Reuters-21578}} \input{plotgrid}

% \def \datasetplot {wipo-sl-sc} \def \datasetname
% {\textsc{WIPO-gamma}} \input{plotgrid}

% \def \captiontext {Regularization} \input{plot_testgrid}

% ----------------------------------------------------------------------

\subsection{Execution Times}
\label{sec:times}

\noindent Table~\ref{tab:timing} reports the average time each method
takes to complete a full training and test run. All timings are
recorded on the same machine, equipped with a 12-core processor Intel
Core i7-4930K at 3.40GHz with 32 GB of RAM and an Nvidia GeForce GTX
1080, under Ubuntu 16.04 (LTS).  Note that SVMs
% \footnote{We use the implementation available in
% \texttt{scikit-learn}, see
% \url{https://scikit-learn.org/stable/modules/generated/sklearn.svm.LinearSVC.html}}
and \fasttext\
% \footnote{We use the original implementation as available from
% \url{https://fasttext.cc/}}
run on CPU, while CNN, LSTM, and ATTN (in our own implementation based
on PyTorch\footnote{\url{https://pytorch.org/}}) run on GPU.  We do
not clock model selection, i.e., the times reported correspond to the
execution time of the model with already optimized
hyperparameters.\footnote{The total time of computation to execute all
10 runs for all variants amounts to 37d 14h 52m (this is without
considering model selection, which by itself requires 15d 10h 22m).}
In the interest of brevity, for each neural architecture we only
report times for
% \blue{Random}, \glove\ (trainable), and \glove+WCEs (trainable)
% variants.
\glovetr, and \glovewcetr\ variants.
% \alexcomment{Ho generato pure i tempi per Random; sono tra commenti
% nella tabella.}

\begin{table}[t]
  \center \resizebox{\textwidth}{!}{%
  \begin{tabular}{|l|l||r|r|r|r|r|r|}
    \hline
    \multicolumn{1}{|c}{\side{Model}}  & \multicolumn{1}{|c||}{\side{Variant}} & \centercell{\side{\textsc{20Newsgroups}\hspace{1ex}\mbox{}}} & \centercell{\side{\textsc{Ohsumed}}} & \centercell{\side{\textsc{RCV1-v2}}} & \centercell{\side{\textsc{JRC-Acquis}}} & \centercell{\side{Reuters21578}} & \centercell{\side{\textsc{WIPO-gamma}\hspace{1ex}\mbox{}}} \\\hline\hline

    \multirow{3}{*}{SVM} & \texttt{tfidf} & \textbf{2s} & \textbf{4s} & \textbf{1m \phantom{0}4s} & \textbf{8m \phantom{0}9s}  & \textbf{3s} & -- \\
    % & WCEs & \textbf{1s} & 10s & 3m 22s & 16h 26m 42s & 1m 19s &--
    % \\\hline
                                       & \glove & 1m 55s & 2m 50s & 5m \phantom{0}1s & 11m 18s & 20s & --\\
                                       & \glove+WCEs & 22s & 2m \phantom{0}4s & 4m 47s & 13m 14s & 20s & --\\\hline
    
    \multirow{2}{*}{\fasttext}     
    % \multirow{3}{*}{\fasttext} & Random & 46s & 1m \phantom{0}5s &
    % \textbf{18s} & 18m 32s & 19s & \textbf{18m 57s} \\
                                       & \glove  &  48s & 48s & 2m 30s & 18m 39s & 20s & \textbf{19m 15s }\\
                                       & \glove+WCEs & 3s & 5s & 3m 18s & 36m 17s & 52s & 36m 43s \\\hline

    \multirow{2}{*}{CNN}      
    % \multirow{3}{*}{CNN} & Random & 5m 26s & 16m 49s & 54m 20s & 42m
    % 18s & 10m 13s & 42m 34s \\
                                       & \glove\  & 5m 58s & 19m 16s & 44m 53s  & 26m 41s & 5m 46s & 2h 28m 57s \\
                                       & \glove+WCEs  & 4m  \phantom{0}2s & 8m  \phantom{0}7s  & 25m 31s  & 43m 38s & 7m 45s & 4h 11m 33s \\\hline

    \multirow{2}{*}{LSTM} 
    % \multirow{3}{*}{LSTM} & Random & 7m 51s & 22m 44s & 37m 18s & 1h
    % \phantom{0}6m 35s & 17m 11s & 1h 31m 38s \\
                                       & \glove\ & 6m 51s & 35m 59s & 1h  \phantom{0}6m 53s & 20m 14s & 10m  \phantom{0}9s  & 3h 10m 28s \\
                                       & \glove+WCEs  & 16m 10s  & 20m 55s & 1h 19m 37s & 41m 33s & 9m 19s & 1h 35m  \phantom{0}9s \\\hline

    \multirow{2}{*}{ATTN}                               
    % \multirow{3}{*}{ATTN} & Random & 5m 12s & 11m 33s & 25m 34s & 1h
    % 13m 38s & 11m 39s & 47m 23s \\
                                       & \glove\ & 4m 54s & 32m  \phantom{0}3s & 1h 10m 22s & 25m 32s & 18m 41s  & 3h  \phantom{0}3m 52s \\
                                       & \glove+WCEs  & 4m 22s & 5m 52s  & 32m  \phantom{0}7s  & 35m  \phantom{0}7s & 8m 28s & 1h 40m 48s \\\hline
		
  \end{tabular}
  }
  \caption{Average run time for dataset; \textbf{boldface} indicates
  the fastest method for each
  dataset.}%, ordered by increasing number of classes}
  \label{tab:timing}
\end{table}

SVMs with tfidf features typically rank as the fastest method.  When
learning from dense representations SVMs become slower, though.
% \strikeout{(with the sole exception of \textsc{20Newsgroups}, where
% the dense space has only 20 dimensions).}\alexcomment{Questo
% riguardava SVM+WCEs (che l'ho tolto)}
Since SVMs run one optimization problem for each class, execution
times drastically grow in \textsc{JRC-Acquis} (whose codeframe
consists of 2,706 classes).  The combination of these two factors
(dense representations and large codeframes)
may %\strikeout{lead SVMs to become impractical,}
further penalize execution times for SVMs, as shown in the experiment
running \svmglove\ and \svmglovewce\ on \textsc{JRC-Acquis}.  The rest
of the models instead run one simple optimization for all classes at
once.  No clear pattern emerges as to which among \glove\ and
\glovewce\ is faster.  This means that the benefits in classification
accuracy brought about by enlarging the dimensionality of the
embedding space with word-class features does not necessarily entail a
penalization in terms of running times; convergence times remain in
any case governed by the stochastic nature of backpropagation.
Finally, even if \fasttext\ does not implement any early-stopping
policy, it still stands out as the fastest neural approach by a large
margin, and is sometimes even comparable to SVMs.  In particular, the
two running times that stand out are those regarding
\fasttext-\glovewce\ on \textsc{20Newsgroups} and \textsc{Ohsumed}.
In both cases, these low training times are due to the fact that, as
resulting from hyperparameter optimization, only 5 training epochs are
required, instead of the 200 epochs that are required for
\fasttext-\glove.

Table~\ref{tab:suptiming} reports the time it takes to create the
WCEs, which can be broken down into two components, i.e., (i) the time
needed to generate matrix $\mathbf{X}$ (which encodes the bag-of-words
model with tfidf weighting), and (ii) the time needed to subsequently
generate matrix $\mathbf{S}$ (which contains the WCEs).
% The total time is broken down in the time required to create the
% tfidf matrix and the time to obtain the supervised embedding matrix.
Most of the total time is accounted for by the generation of matrix
$\mathbf{X}$ (1st row of Table~\ref{tab:suptiming}); the computational
cost of generating the WCEs from matrix $\mathbf{X}$ is almost
negligible (2nd row of Table~\ref{tab:suptiming}), and typically
represents less than 10\% of the total time.  The higher times clocked
for \textsc{JRC-Acquis} and \textsc{WIPO-gamma} are due to the
application of PCA, which is not necessary for the other datasets (see
Section~\ref{sec:method:codeframes}). The total times (last row of
Table~\ref{tab:suptiming}) are, in all cases, much smaller than the
times needed for optimizing the models, that are shown in
Table~\ref{tab:timing}.

% Complexity is $O(d f c)$
\begin{table}[t]
  \center \resizebox{\textwidth}{!}{%
  \begin{tabular}{|c||r|r|r|r|r|r|}
    \hline
    \multicolumn{1}{|c||}{\textbf{\side{Computation}}} & \centercell{\side{\textsc{20Newsgroups}\hspace{.5ex}\mbox{}}} & \centercell{\side{\textsc{Ohsumed}}} & \centercell{\side{\textsc{RCV1-v2}}} & \centercell{\side{\textsc{JRC-Acquis}}} & \centercell{\side{Reuters21578}} & \centercell{\side{\textsc{WIPO-gamma}\hspace{1ex}\mbox{}}} \\\hline\hline
    $\mathbf{X}$ & 2.10s (99.16\%) & 3.85s (98.95\%) & 4.74s (97.34\%) & 14.00s (56.00\%) & 1.11s (96.78\%) &  4m 39.00s (91.56\%) \\
    $\mathbf{S}$ & 0.02s \phantom{0}(0.84\%)  & 0.04s \phantom{0}(1.05\%)  & 0.13s \phantom{0}(2.66\%)  & 11.00s (44.00\%) & 0.04s \phantom{0}(3.22\%)  & 25.79s \phantom{0}(8.44\%) \\\hline\hline
    Total & 2.12s \phantom{(00.00\%)} & 3.89s \phantom{(00.00\%)} & 4.87s \phantom{(00.00\%)} & 25.00s \phantom{(00.00\%)} & 1.15s \phantom{(00.00\%)} & 5m  \phantom{0}5.00s \phantom{(00.00\%)} \\\hline

\end{tabular}
}
\caption{Total time required by the computation of WCEs ($\mathbf{X}$
indicates the ``weighted bag-of-words'' matrix mentioned at the
beginning of Section~\ref{sec:method}, while $\mathbf{S}$ indicates
the WCE matrix of Equation~\ref{eq:finalmatrix}).}
\label{tab:suptiming}
\end{table}

% The time it takes to generate a classifier with linear SVMs largely
% depends on the number of classes, since SVMs need to solve one
% optimization problem for each of them.  Indeed, \textsc{JRC-Acquis}
% is the dataset with the largest codeframe (2,706 classes) and the
% dataset for which SVMs require more time to train.  Training times
% also increase when operating with dense representations (SVM-S)
% instead of sparse representations (SVM-tfidf).  Finally, SVMs are
% not incremental, and thus require the entire training set to be
% allocated in memory. This limitation actually prevented SVMs to be
% able to generate a classifier for \textsc{WIPO-gamma}, a dataset
% containing nearly 900K training examples.  SupTotal indicates the
% time it takes to generare the WCEs. Note these records are
% negligible when compared to the training times, which means the
% benefits brought about by the use of WCEs come at almost no
% cost. BoW and SupEmb report SupTotal times broken down in the time
% to generate the Tfidf matrix of the training corpus and the time it
% takes to compute the embeddings afterwards. Interestingly, BoW
% accounts for most of the computation time in EmbTotal (typically,
% most of 90\% of it). This means models already relying on the BoW
% model could benefit from the use of WCEs almost for free. Add: the
% only cases in which SupEmb took some time are \textsc{WIPO-gamma}
% (PCA) and \textsc{JRC-Acquis} (PCA and multilabel).

% ---------------------------------------------------------

\subsection{Other Measures of Correlation}
\label{sec:tsr}

\noindent In this section we explore other correlation measures as
alternative ways for computing the WCEs. In other words, we explore
alternatives to the use of the dot product %for generating matrix
% $\mathrm{Q}$ in Equation~\ref{eq:A}. \fsebcomment{Dovremmo definire
% una generica funzione $f$ (di cui il dot product è un esempio) e
% riscrivere l'Equation~\ref{eq:A} in forma matriciale in modo che usi
% $f$.}
for instantiating the $\eta$ function of Equation~\ref{eq:eta}.
%
% Also in these cases we apply standardizing to the resulting matrix
% $\mathbf{A}$ (Equation~\ref{eq:A}), since we have observed its
% application to be beneficial irrespectively of the function used to
% instantiate $\eta$. %of the $\eta$ function of choice.
%% The rest of the procedure to generate the WCEs () remains
%% untouched.
% \fsebcomment{Hmmm ...}
For this, we use well-known functions from information theory or
statistics that have been routinely used for feature selection
purposes in text classification, including Positive Pointwise Mutual
Information (PPMI -- see~Footnote~\ref{footnote:PMI}), Information
Gain (IG), and Chi-square ($\chi^2$).
% (see, e.g.,~\cite{forman2003extensive} for an overview).
% \andscomment{List them here. Comment if and how for any of them the
% X matrix is defined differently from our method.}
In preliminary experiments we had carried out using these functions we
had indeed found that standardizing the resulting matrix $\mathbf{A}$
(Equation~\ref{eq:A}) improves accuracy.  We thus report the results
of using as the $\eta$ function one that also performs standardizing,
e.g., $\eta_{\chi^2}(t_i,c_j)=z_{j}(\chi^2(t_i,c_j))$ (similarly for
PPMI and IG).
% \strikeout{, which thus defines stronger baselines against which to
% compare our method.}} \fsebcomment{Non direi che queste siano delle
% baselines.}  \blue{(This also ensures that any improvement in
% performance brought about by the method we propose can not merely be
% attributed to standardizing.)}

\def \captiontext {Classification accuracy (in terms of $F^{M}_{1}$)
resulting from the use of measures of correlation alternative to the
one (``Dot'') that we use for generating WCEs.}  \def \plotdir
{supervised_functions} \def \plotlabel {fig:tsr}
\plot{\def \datasetA {20newsgroups}
\def \datasetB {jrcall}
\def \datasetC {ohsumed}
\def \datasetD {rcv1}
\def \datasetE {reuters21578}
\def \datasetF {wipo-sl-sc}

\def \datasetnameA {20 Newsgroups}
\def \datasetnameB {JRC-Acquis}
\def \datasetnameC {Ohsumed}
\def \datasetnameD {RCV1-v2}
\def \datasetnameE {Reuters-21578}
\def \datasetnameF {WIPO-gamma}

\begin{figure}
\centering

\begin{minipage}{.08\textwidth}
\center \scriptsize 
\end{minipage}
\begin{minipage}{.29\textwidth}
\center \scriptsize \hspace{1.6cm} CNN
\end{minipage}
\begin{minipage}{.29\textwidth}
\center \scriptsize \hspace{1.6cm} LSTM
\end{minipage}
\begin{minipage}{.29\textwidth}
\center \scriptsize \hspace{1.6cm} ATTN
\end{minipage}

\begin{minipage}{.08\textwidth}
\center \scriptsize \rotatebox{90}{\datasetA}
\end{minipage}
\begin{minipage}{.29\textwidth}
\center \scriptsize \includegraphics[width=\linewidth]{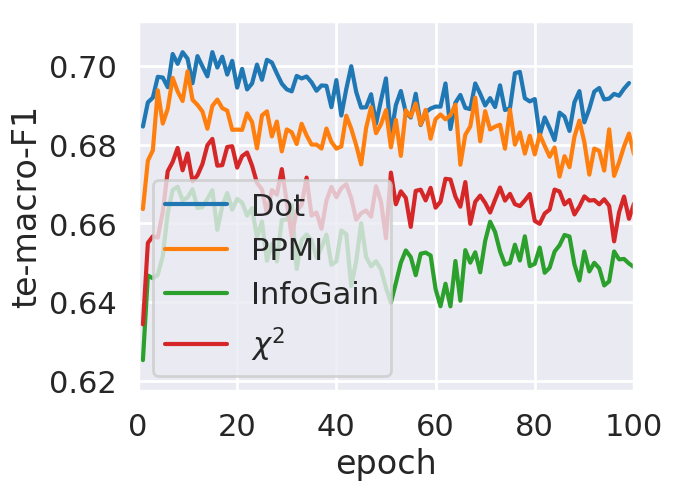}
\end{minipage}
\begin{minipage}{.29\textwidth}
\center \scriptsize \includegraphics[width=\linewidth]{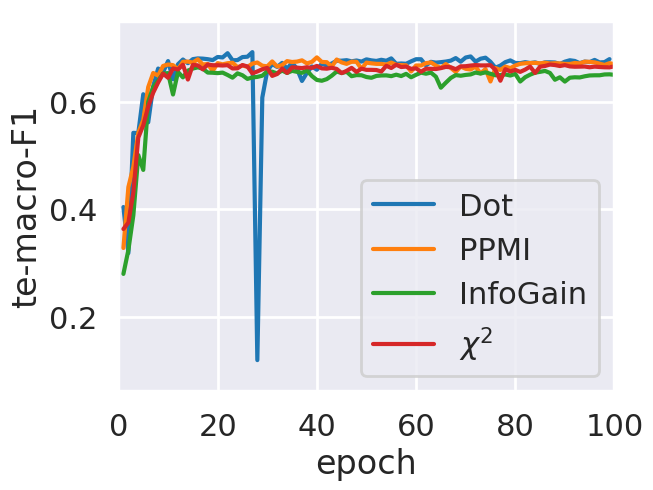}
\end{minipage}
\begin{minipage}{.29\textwidth}
\center \scriptsize \includegraphics[width=\linewidth]{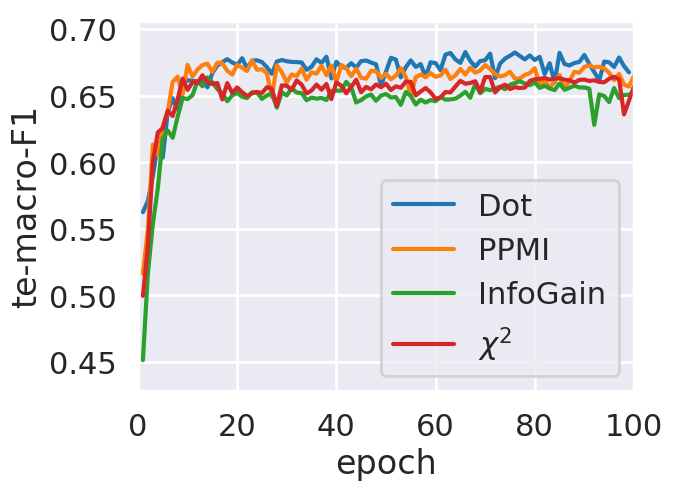}
\end{minipage}

\begin{minipage}{.08\textwidth}
\center \scriptsize \rotatebox{90}{\datasetnameB}
\end{minipage}
\begin{minipage}{.29\textwidth}
\center \scriptsize \includegraphics[width=\linewidth]{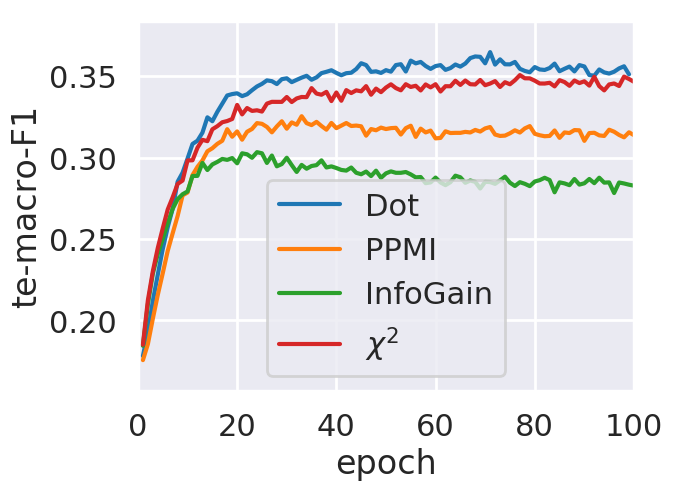}
\end{minipage}
\begin{minipage}{.29\textwidth}
\center \scriptsize \includegraphics[width=\linewidth]{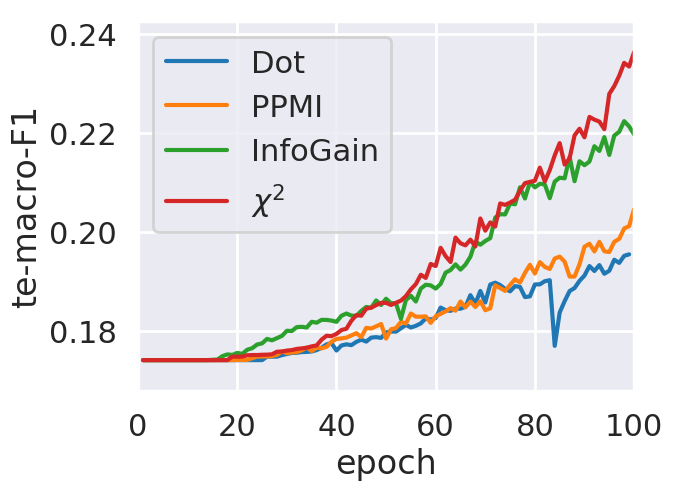}
\end{minipage}
\begin{minipage}{.29\textwidth}
\center \scriptsize \includegraphics[width=\linewidth]{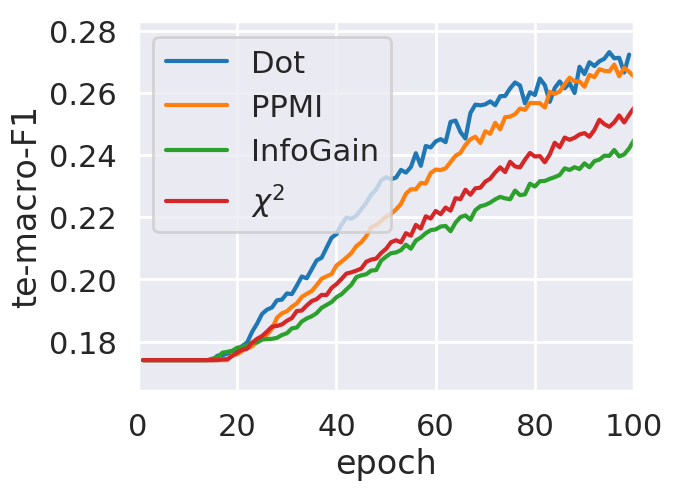}
\end{minipage}

\begin{minipage}{.08\textwidth}
\center \scriptsize \rotatebox{90}{\datasetnameC}
\end{minipage}
\begin{minipage}{.29\textwidth}
\center \scriptsize \includegraphics[width=\linewidth]{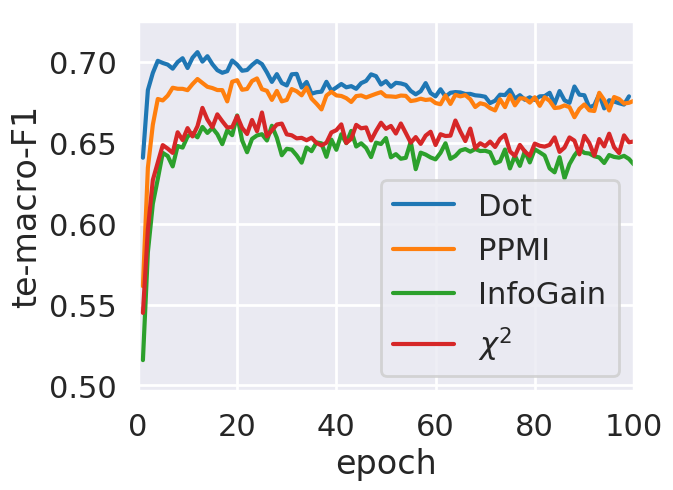}
\end{minipage}
\begin{minipage}{.29\textwidth}
\center \scriptsize \includegraphics[width=\linewidth]{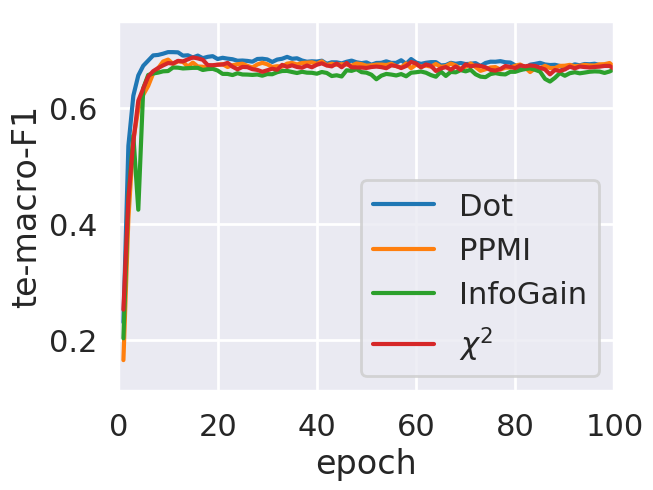}
\end{minipage}
\begin{minipage}{.29\textwidth}
\center \scriptsize \includegraphics[width=\linewidth]{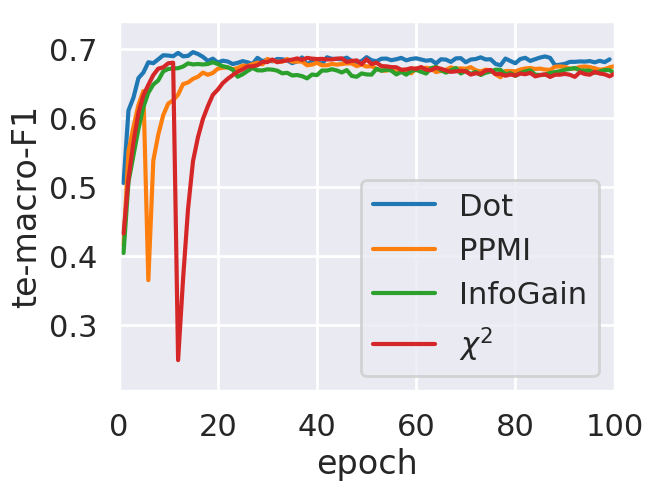}
\end{minipage}

\begin{minipage}{.08\textwidth}
\center \scriptsize \rotatebox{90}{\datasetnameD}
\end{minipage}
\begin{minipage}{.29\textwidth}
\center \scriptsize \includegraphics[width=\linewidth]{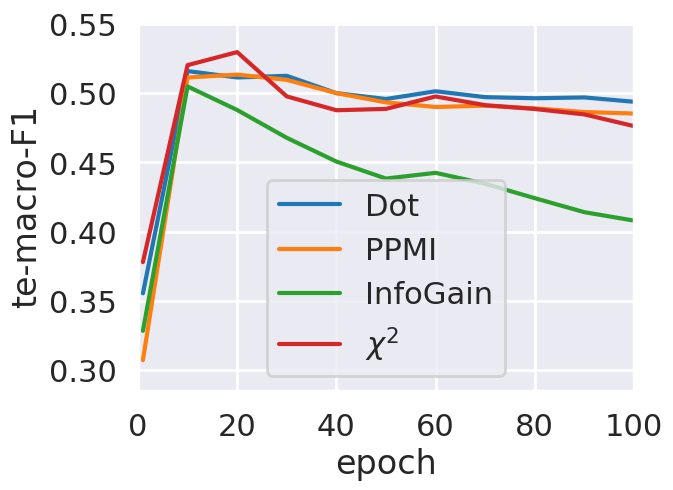}
\end{minipage}
\begin{minipage}{.29\textwidth}
\center \scriptsize \includegraphics[width=\linewidth]{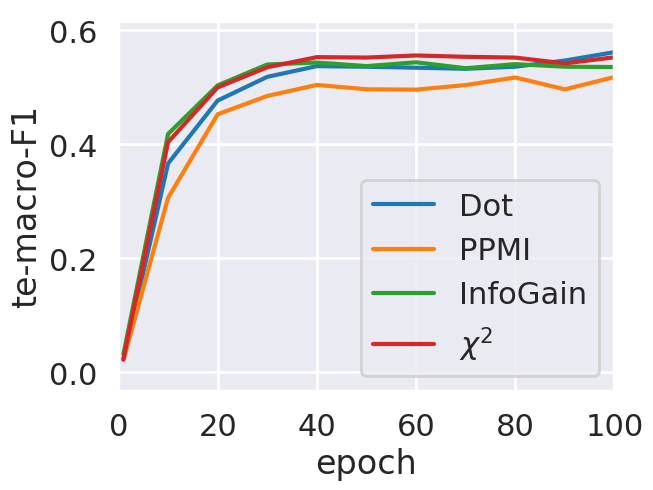}
\end{minipage}
\begin{minipage}{.29\textwidth}
\center \scriptsize \includegraphics[width=\linewidth]{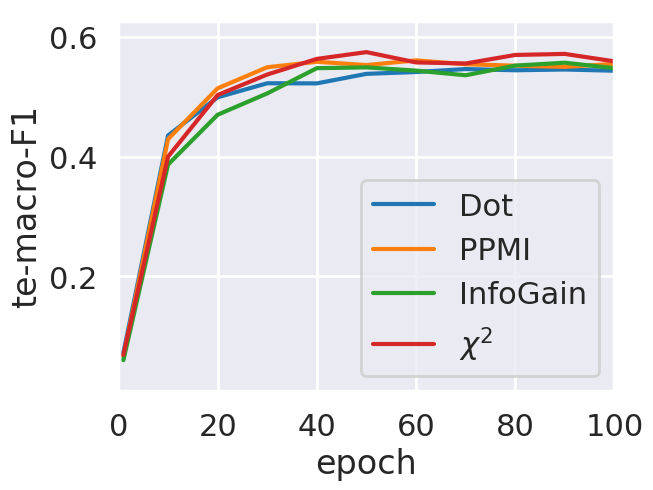}
\end{minipage}

\begin{minipage}{.08\textwidth}
\center \scriptsize \rotatebox{90}{\datasetnameE}
\end{minipage}
\begin{minipage}{.29\textwidth}
\center \scriptsize \includegraphics[width=\linewidth]{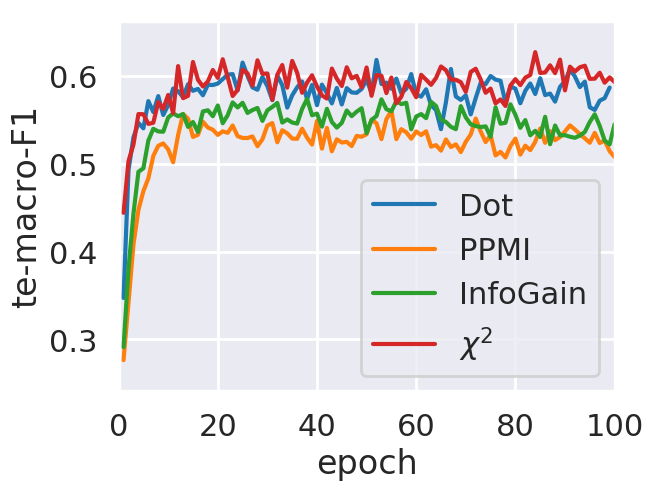}
\end{minipage}
\begin{minipage}{.29\textwidth}
\center \scriptsize \includegraphics[width=\linewidth]{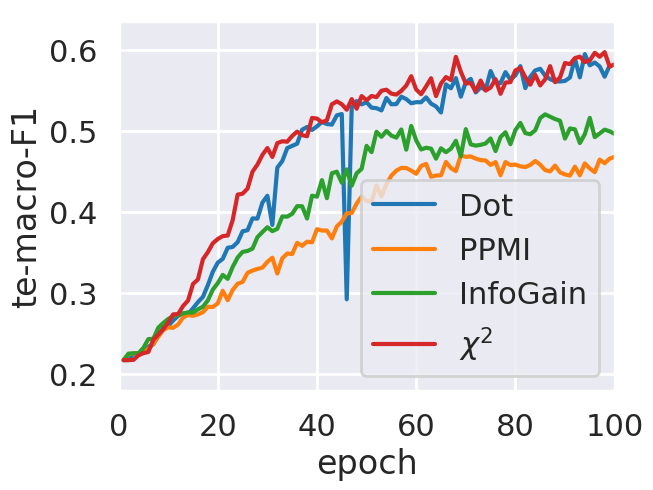}
\end{minipage}
\begin{minipage}{.29\textwidth}
\center \scriptsize \includegraphics[width=\linewidth]{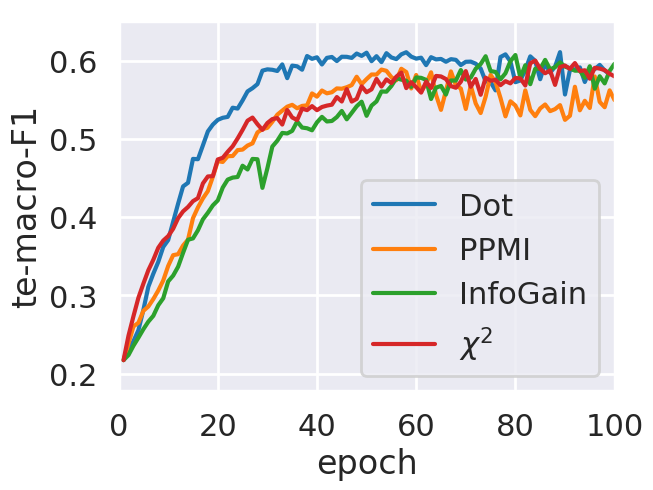}
\end{minipage}

\begin{minipage}{.08\textwidth}
\center \scriptsize \rotatebox{90}{\datasetnameF}
\end{minipage}
\begin{minipage}{.29\textwidth}
\center \scriptsize \includegraphics[width=\linewidth]{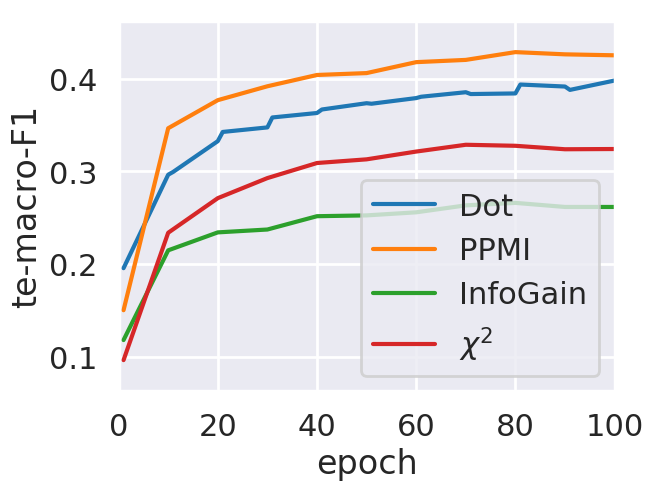}
\end{minipage}
\begin{minipage}{.29\textwidth}
\center \scriptsize \includegraphics[width=\linewidth]{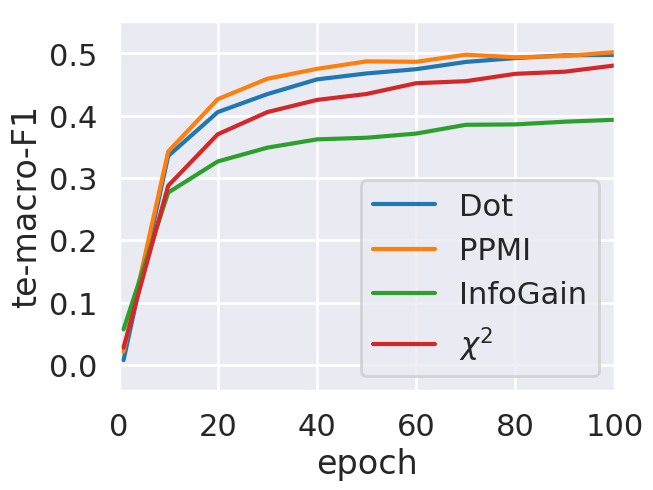}
\end{minipage}
\begin{minipage}{.29\textwidth}
\center \scriptsize \includegraphics[width=\linewidth]{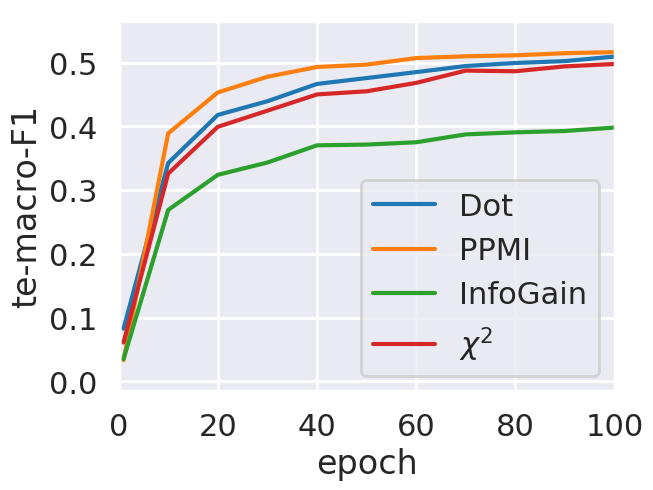}
\end{minipage}

\caption{\captiontext}
\label{\plotlabel}
\end{figure}}

Figure~\ref{fig:tsr} compares the classification performance, in terms
of $F_{1}^{M}$, of the dot product (as originally used in
Equation~\ref{eq:A}, and here abbreviated as ``Dot'') against PPMI,
IG, and $\chi^2$.
% Positive Pointwise Mutual Information (PPMI --
% see~Footnote~\ref{footnote:PMI}), Information Gain (IG), and
% Chi-square ($\chi^2$).  , and GSS (a positive-only correlation
% measure discussed in~\cite{}).  [Maybe is better to replace PMI
% with PPMI and GSS with ConfWeight or TFRF]
%
As can be observed from Figure~\ref{fig:tsr}, ``Dot'' is almost always
superior to all other functions, or at least comparable to the
best-performing function, across all datasets and network
architectures.  Other functions behave irregularly across experiments;
for example, PPMI seems to be the most competitive method for
\textsc{WIPO-gamma}, but is a weak one in \textsc{Reuters-21578} (this
applies to all architectures) and \textsc{JRC-Acquis} (in CNN and
LSTM).

We conjecture that the superiority of the dot product partly depends
on its ability to take non-binary (in our case: tfidf) weights into
account, while PPMI, IG, and $\chi^2$ only consider binary
presence/absence indicators.\footnote{Though most traditional
functions used for feature selection can only use presence/absence,
other metrics exist that work with weighted scores, e.g., the
\emph{Fisher score}. In initial experiments not described in this
paper we have indeed tried to use the Fisher score, but we have
eventually given up, due to the fact that (a) its computation is very
slow, and (b) the classification accuracy that we have observed is not
much different from what can be obtained with the other functions
mentioned above, and is often intermediate between the best and the
worst recorded values.} Additionally, the dot product only leverages
positive correlations (this is also true for PPMI), and is thus much
faster than other methods that compute negative correlations at
intermediate steps, as is the case of IG and $\chi^2$.  The reason is
that our tfidf matrix $\mathbf{X}$ and the label matrix $\mathbf{Y}$
(see Equation~\ref{eq:A}) are highly sparse; many machine learning
software packages (including the ones we use) do cater for the
presence of sparse matrices, and thus do not explicitly represent zero
values, causing any operation involving non-zero values
% (e.g., to compute $\Pr(t_{i},\overline{c_{j}})$ one needs to count
% on $\mathbf{X}$ and $\mathbf{Y}$ how many documents containing term
% $t_{i}$ are \emph{not} labelled with class $c_{j}$)
to substantially increase computation times.  For example, it takes
20s to compute matrix $\mathbf{S}$ for \textsc{Reuters-21578} when
using $\chi^2$, while the same only takes 0.04s when using the dot
product (see Table~\ref{tab:suptiming}).
% It might be worth investigating the \alexcomment{[Might it be
% interesting to test our technique as a feature selection function in
% TC?]}

% ----------------------------------------------------------------------

\subsection{Different Pre-trained Embeddings}
\label{sec:exp:pre-trained}

\noindent Up to now, he have tested the performance of WCEs as an
extension of \glove\ vectors. It might be interesting to check how
well WCEs could perform when concatenated to pre-trained vectors other
than those generated by \glove.

\def \captiontext {Classification accuracy resulting from the use of
\glove\ embeddings or \wordvec\ embeddings, with or without
concatenated WCEs.}  \def \plotdir {glove_vs_word2vec} \def \plotlabel
{fig:pre-trained} \plot{}

Table~\ref{fig:pre-trained} shows the results of experiments in which
WCEs are concatenated to \wordvec\ pre-trained
embeddings~\cite{mikolov2013distributed}.\footnote{Available at
\url{https://code.google.com/archive/p/word2vec/}} The results show
that the improvements brought about by WCEs for \wordvec\ are similar
to those reported for \glove. It thus seems reasonable to believe that
the use of WCEs is beneficial in general, and independently of the
particular set of pre-trained embeddings under consideration.
% In principle, it would be possible to use random indexing-like
% vectors as well, inasmuch as some dimensionality reduction (e.g.,
% PCA) is applied in advance to them.

One question that still remains open is how much effort it might
require to pair WCEs with differently characterized embeddings, such
as contextualized vector
representations~\cite{BERT,mccann2017learned,peters2018deep,XLNet},
subword representations~\cite{bojanowski2017enriching},
character-based
embeddings~\cite{kim2016character,zhang2015character}, or
high-dimensional sparse
representations~\cite{sahlgren2005introduction}.  We plan to answer
these questions in future research.

% ----------------------------------------------------------------------

\subsection{Visualizing WCEs}
\label{sec:exploration}

\noindent In this section we try to gain an understanding on how terms
represented by WCEs are topologically distributed in the embedding
space, and how WCEs alter the distribution of pre-trained embeddings
once they are concatenated to them.  To do so we use \texttt{Embedding
Projector}, a publicly available tool for data visualization based on
the \emph{t-distributed Stochastic Neighbor Embedding} (t-SNE)
technique~\cite{tSNE}\footnote{\url{https://projector.tensorflow.org/}},
and which allows to map word embeddings onto a 2-dimensional space.
We use its default parameters (perplexity=18 and learning rate=10) and
perform 1,000 iterations.  We only represent 5,000 terms, in order to
get a clearer visualization; the terms we select are the most
predictive ones (as quantified via information gain) for each class,
following a ``round robin'' policy which selects the same number of
highly predictive terms for each class~\cite{forman2004pitfall}.
% \footnote{Round Robin perform round passes over classes, picking the
% most predictive feature ranked for that class, until the desired
% number of features has been reached.}
We assign different colours to the classes, and colour each embedding
according to the class for which it was selected.  For this experiment
we choose \textsc{20Newsgroups} (the dataset with fewest classes),
with the aim of keeping the colour coding simple enough, thus
maximizing visual clarity.% to be inspected.

Figure~\ref{fig:glove} shows the distribution of \glove\ vectors. The
top part shows that, to some extent, some word clusters seem to
correlate with some classes. This was to be expected, since terms
relevant to a given class tend to be semantically related to each
other. The enlarged region (bottom) shows how \glove\ succeeds at
producing meaningful local structures containing smaller clusters of
semantically interrelated words, e.g., \{``diet'', ``dietary'',
``vitamin''\}, or \{``doctor'', ``medical'', ``hospital''\}, within
class \texttt{sci.med}.

\def \imagewidth {0.53\textwidth}

\plot{
\begin{figure}[t]
  \begin{minipage}{\textwidth}
    \center \includegraphics[width=\imagewidth]{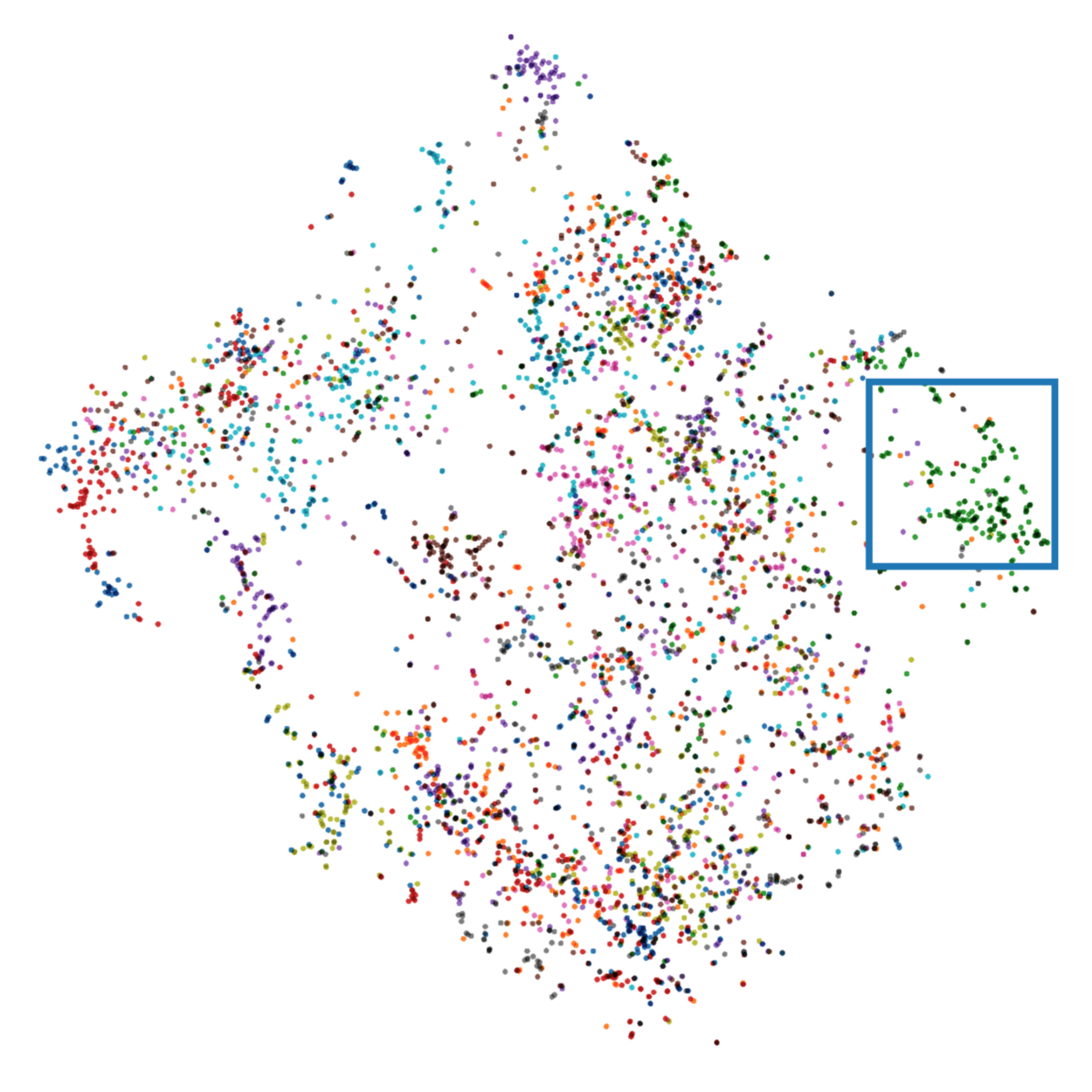}
  \end{minipage}
  \begin{minipage}{\textwidth}
    \center \includegraphics[width=\imagewidth]{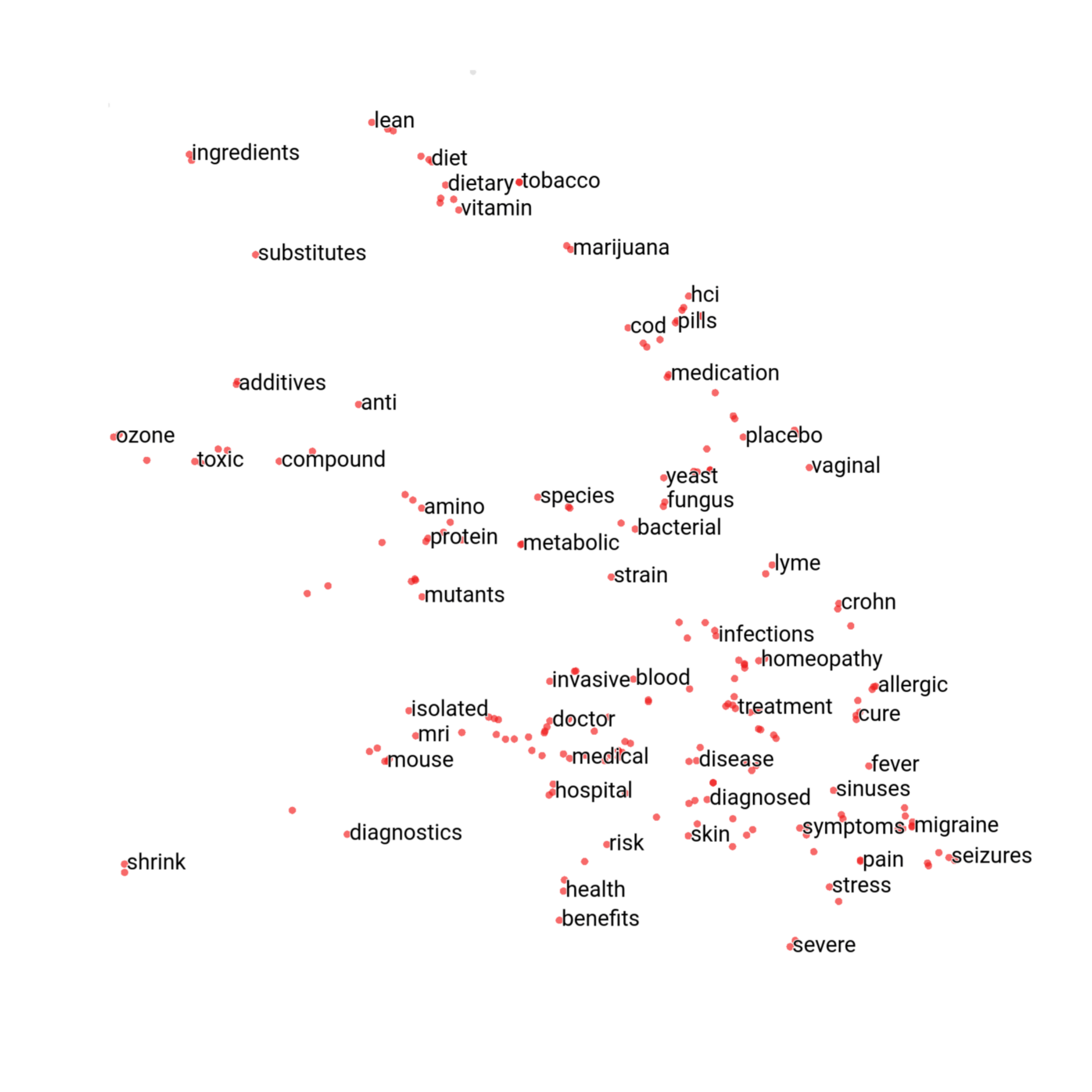}
  \end{minipage}
  \caption{Visualization (best viewed in color) of the space of
  \glove\ embeddings for \textsc{20Newsgroups} (top), and enlargement
  of the part in the blue square (bottom). Each point represents an
  embedding, and (in the top part) each colour represents one of the
  20 classes in the codeframe, and the one the term represented by the
  embedding is highly predictive of.}
  \label{fig:glove}
\end{figure}
}

Figure~\ref{fig:supervised} shows the distribution of WCEs for the
very same terms represented in~Figure~\ref{fig:glove}.  The
visualization shows almost perfect word clusters for classes (top).
This should come at no surprise, since the WCEs explicitly encode
class structure.  As a counterpart, local semantics within clusters
vanishes (bottom), given that WCEs disregard word-word interactions.
For example, for certain pairs of words (e.g., \{``attacking'',
``attacks''\}, \{``terror'',``terrorism''\}), the two words happen to
lie far from each other within the cluster for class
\texttt{talk.politics.mideast}.

\def \imagewidth {0.55\textwidth} \plot{
\begin{figure}[t]
  \begin{minipage}{\textwidth}
    \center
    \includegraphics[width=\imagewidth]{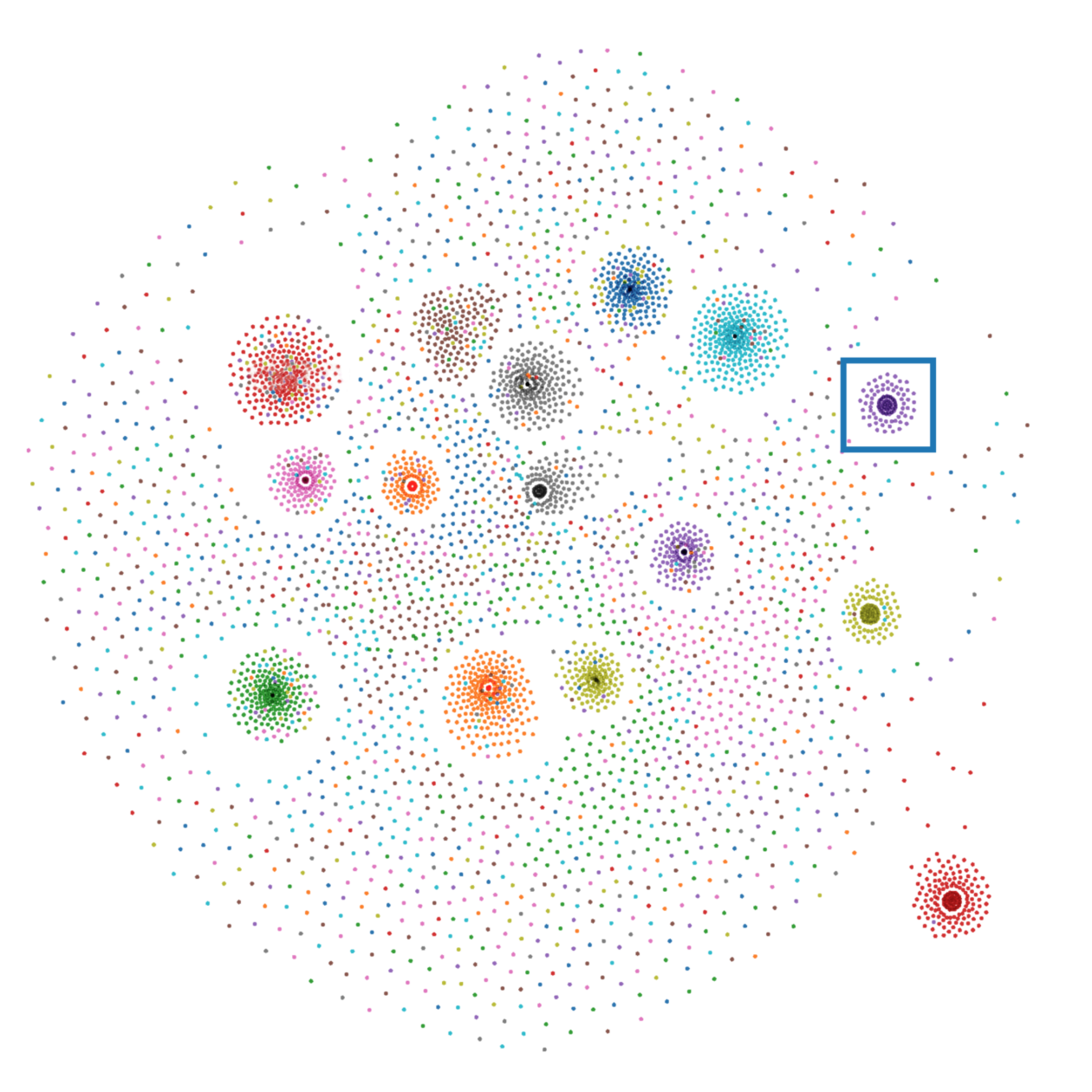}
  \end{minipage}
  \begin{minipage}{\textwidth}
    \center
    \includegraphics[width=\imagewidth]{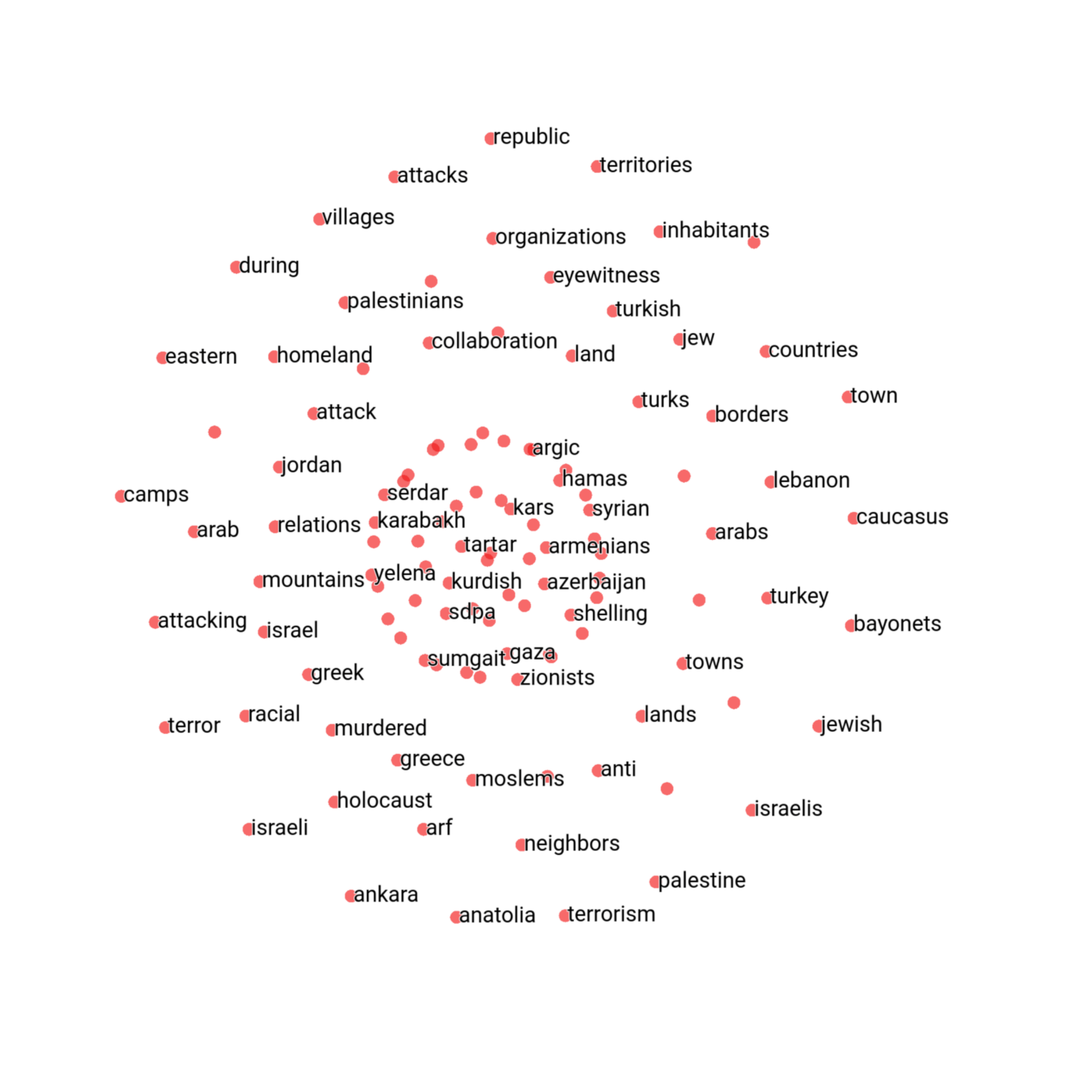}
  \end{minipage}
  \caption{Same as Figure~\ref{fig:glove}, but with WCEs instead of
  \glove\ embeddings.}
  \label{fig:supervised}
\end{figure}
}

Figure~\ref{fig:glove-supervised} shows the distribution of the
concatenation of \glove\ vectors and WCEs for the very same terms of
Figures~\ref{fig:glove} and~\ref{fig:supervised}.  This representation
brings together the best of the two worlds.  Globally, a neat class
structure emerges, as imposed by the WCEs (top).  Locally, clusters
exhibit a meaningful inner structure, thanks to \glove\ vectors
(bottom).  As an example, relevant words for class \texttt{sci.space}
organize in smaller clusters such as \{``moon'', ``earth'',
``lunar''\} and \{``allen'', ``grifin'', ``dani''\}.

\plot{
\begin{figure}[t]
  \begin{minipage}{\textwidth}
    \center
    \includegraphics[width=\imagewidth]{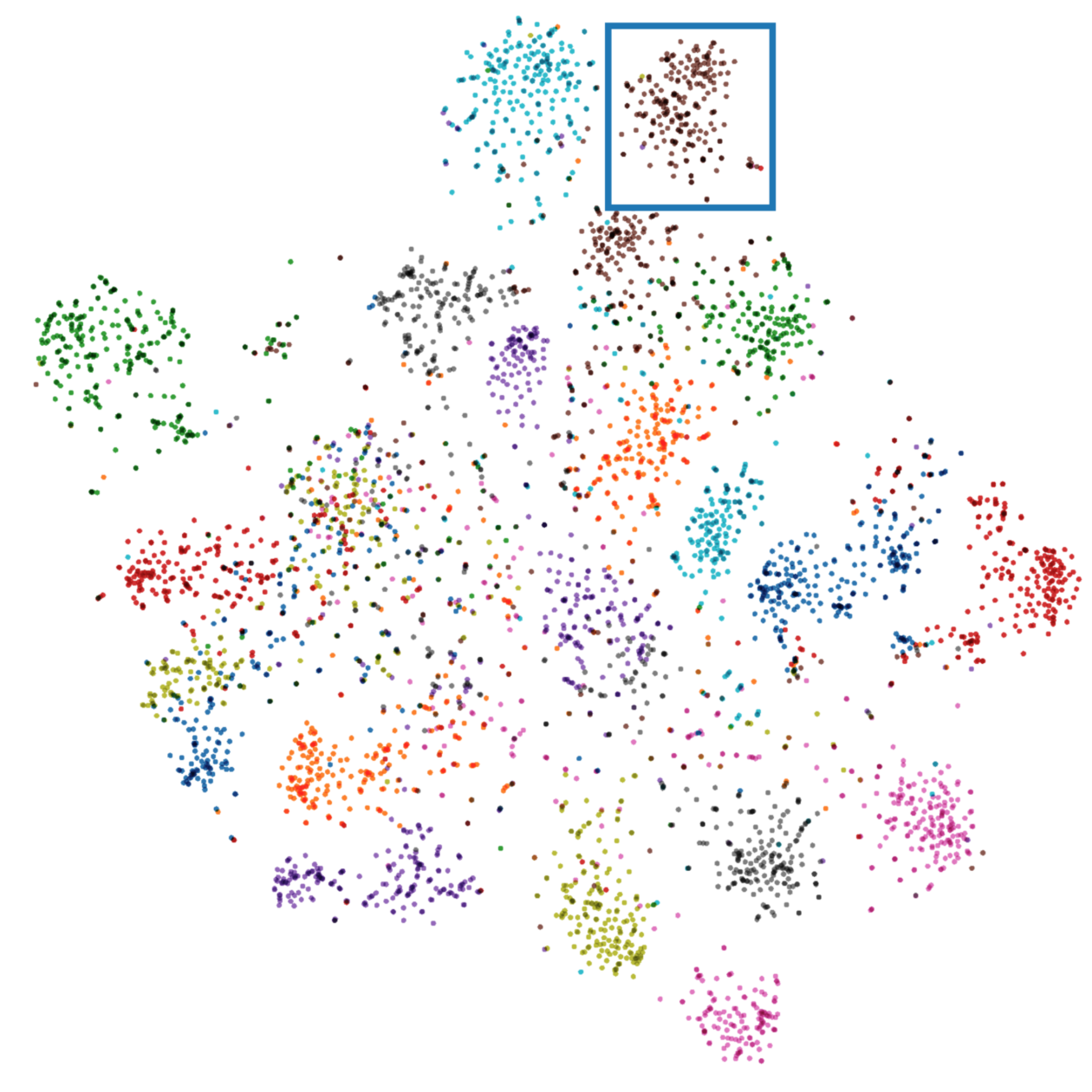}
  \end{minipage}
  \begin{minipage}{\textwidth}
    \center
    \includegraphics[width=\imagewidth]{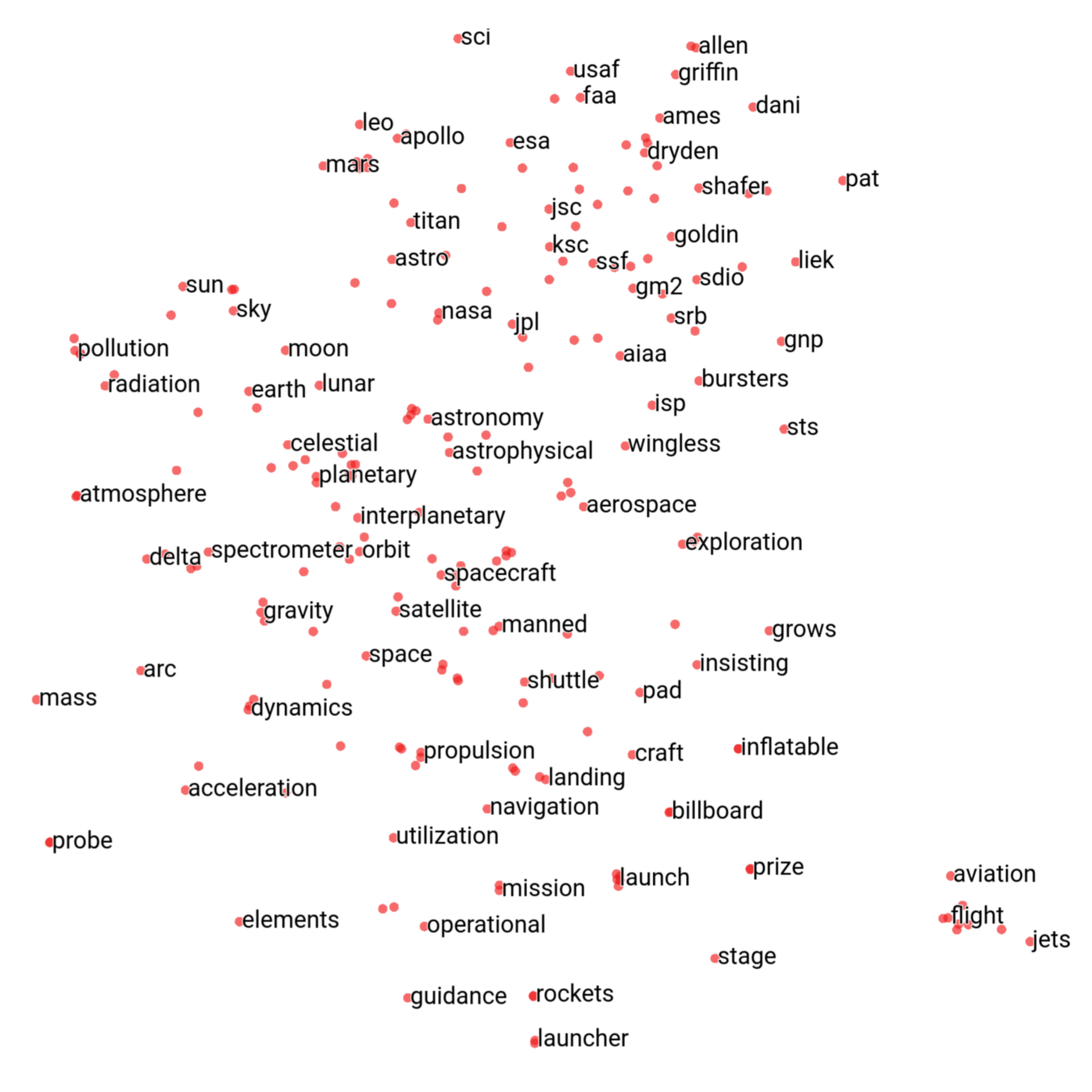}
  \end{minipage}
  \caption{Same as Figure~\ref{fig:glove}, but with \glove+WCE
  embeddings instead of \glove\ embeddings.}
  \label{fig:glove-supervised}
\end{figure}
}

% ----------------------------------------------------------------------

\subsection{Can we Learn WCEs for Out-of-Vocabulary Terms?}
\label{sec:predict}

\noindent In contrast to unsupervised word embeddings, WCEs are
inherently task-dependent, and thus heavily rely on the distribution
of the terms in the training set.  This means that, for any word
encountered at testing time that was not observed during training, the
corresponding WCE will be a vector of zeroes (or a a randomly
initialized vector), and this could harm performance.  This is yet
another manifestation of the well-known problem of
\emph{out-of-vocabulary} (OOV) \emph{terms}, which represents an
active area of research in the field of word
embeddings~\cite{Garneau:2019rc}.

Classifiers that make use of pre-trained embeddings do have a chance
to make sense of OOV terms (i.e., terms not encountered at classifier
training time).  When (as is usually the case) pre-trained embeddings
are generated from huge quantities of text, an OOV term is fairly
likely to have a corresponding pre-trained embedding.  If this is the
case, what is learnt during classifier training is thus related to
this term too, inasmuch as its pre-trained embedding is at least
partly aligned with the embeddings of some terms encountered during
classifier training.
% This problem is related to the well-known problem of
% out-of-vocabulary terms, an active area of research in the field of
% word embeddings.

In this section we cope with the problem of WCEs for OOV terms.  Our
idea is to predict the WCE for an OOV term from its pre-trained
embedding (when available), on the grounds that semantically similar
terms (as observed in general language use) can be expected to exhibit
similar class-conditional distributions.
% \fsebcomment{Qui un esempio potrebbe starci
% bene.} % to the classification problem (correlationte
% the fact that these word-label correlations are never observed
% during training).

We frame the problem of generating the WCE for an OOV term as a
multivariate regression task.\footnote{Another technique for solving
this problem is \emph{Latent Semantic
Imputation}~\cite{yao2019enhancing}.
% LSI generates embedding vectors for out-of-vocabulary terms (with
% missing representation) in a vector space (in our case: in the space
% of supervised embedding space) by transferring the representation of
% their known representation in another vector space (in our case:
% unsupervised embedding space) analyzing the neighborhood and
% applying techniques inspired by manifold learning
This methods allows filling the missing representation in a vector
space (in our case: in the space of WCEs) by analyzing the
neighborhood of the term representation in another vector space (in
our case: the space of unsupervised embeddings) via techniques
inspired by manifold learning.}  As from Section~\ref{sec:method}, let
$\mathbf{E}=[\mathbf{U} \oplus \mathbf{S}]$ be the embedding matrix,
where rows $\mathbf{e}_{i}=[\mathbf{u}_{i} \oplus \mathbf{s}_{i}]$ are
the embeddings, consisting of a concatenation of a pre-trained
embedding $\mathbf{u}_{i}$ and a WCE $\mathbf{s}_{i}$. We train a
two-layered feed-forward network (in the experiments we use 64 units
in the hidden layer, ReLU activation, and 0.5 dropout) to predict the
WCE $\mathbf{s}_{i} \in \mathbf{S}$ from the pre-trained embedding
$\mathbf{u}_{i} \in \mathbf{U}$, using the terms in the vocabulary as
the training examples. We adopt Mean Square Error (MSE) as the loss
criterion. Once trained, and once an OOV term is encountered, the
regressor is asked to generate the WCE for it, provided this term has
a pre-trained embedding. %\alexcomment{improve}

We have run experiments, using \textsc{RCV1-v2} as the dataset and all
the five learners of Section~\ref{sec:classifiertraining}, in which we
compare the accuracy of two different configurations: (i) a
configuration in which the WCEs of an OOV term is a vector of zeros
(which is the setting we have used so far), and (ii) a configuration
in which the WCEs of an OOV term has been predicted from the
pre-trained embedding of the term, using all the non-OOV
% \alexcomment{(sarebbe forse comodo definire ``In Vocabulary (IV)''?
% )}
terms as training examples.

Unfortunately, the differences in classification accuracy between the
two configurations turned out to be barely discernible; in the
interest of brevity, we thus omit to plot them out explicitly.
% We omit the plots comparing the $F_{1}^{M}$ and $F_{1}^{\mu}$ for
% the original models against a variant equipped with the regressor
% since both trends are overlapped.
The likely reason of this result is that, while \textsc{RCV1-v2}
contains no less than 384,327 OOV terms, occurring 2,073,278 times in
the test set, these represent only 2.16\% of the total number of term
occurrences, which means that their impact on classification accuracy
is minimal.

However, in a qualitative (although somehow ``anecdotal'') evaluation
we have verified that the predicted WCEs for OOV terms look
meaningful.  In order to do this, we looked for OOV terms whose
predicted WCE displays a large correlation with some class, i.e.,
whose WCE is such that the value for some of its features is high.
Some interesting examples of such OOV terms include ``astronauts'',
``invincible'', ``battlefields'', ``indiscriminate'' for class
\texttt{GDEF} (which is about armed forces, defence policy, and
defence budget); ``windfarm'' for class \texttt{GENV} (about
environment, pollution, conservation, green issues); ``pneumatic'' and
``prostatectomy'' for class \texttt{GHEA} (dealing with health and
diseases).

We also found many misspelled terms whose predicted WCE displays a
large correlation with some class, e.g., ``sellling'' and ``exchang''
for classes \texttt{C311} (domestic markets, sales and imports) and
\texttt{C312} (external markets and exports); ``emploment'' for class
\texttt{C41} (all management issues); ``manufatures'' for class
\texttt{E12} (monetary/economic policy and intervention, interest
rates), among others. This is interesting, because the ability to make
sense of misspelled terms is important for many text management
applications.  Incidentally, these findings also speak about the
ability of \glove\ to model rare terms. \fsebcomment{Questo può avere
a che fare con i ``Misspelling Oblivious Word Embeddings'' di Fabrizio
Silvestri?}

Surprisingly, we have also found new correlations for non-English
terms, like Spanish terms ``\textit{aseguradora}'' (insurer) for class
\texttt{E121} (money supply), and ``\textit{mantenimiento}''
(maintenance) for class \texttt{E313} (inventories and stocks of
manufacturing raw materials). \fsebcomment{Qui si potrebbe forse dire
due parole sul perché ci sono queste occorrenze di termini non
inglesi.}

\section{Discussion}
\label{sec:discussion}

% \noindent In this section we explore some ``advanced topics''
% concerning WCEs, including how to generate WCEs for terms that do
% not occur in the training data (Section~\ref{sec:predict}), how the
% topology of the embedding space looks like for different types of
% embeddings (pre-trained, supervised, and concatenations of
% pre-trained and supervised) (Section~\ref{sec:exploration}), and
% what are the known limitations of WCEs
% (Section~\ref{sec:limitations}).

% ----------------------------------------------------------------------

\subsection{Term Semantics: From Unsupervised to Supervised}
\label{sec:connections}

\noindent In Section~\ref{sec:WCEs} we have touched upon the
connections between WCEs and \fasttext, and argued that the two
methods are the supervised counterparts of SPPMI and \wordvec,
respectively. These are just the most recent examples of a trend, in
the field of extracting term semantics from data for text
classification purposes, that over the years has seen a move from
unsupervised to supervised techniques.

The first interesting example of this trend is that of \emph{term
clustering}. In the '90s, term clustering was advocated, among others,
as a means to implement dimensionality reduction for text
classification purposes, according to the idea that clusters of
semantically related terms, instead of individual terms, would serve
as features. While standard unsupervised techniques were initially
used~\cite{Lewis92}, the field slowly moved to using supervised ones,
such as \emph{distributional clustering by class distribution}
(DCCD)~\cite{Baker98,Bekkerman03}. While unsupervised term clustering
has the simple goal of grouping together semantically related terms,
DCCD has the goal of grouping together terms that are discriminative
for the same classes; as such, it constitutes a technique for building
special-purpose term clusters, i.e., ones that are to be used for text
classification only, and only for the specific codeframe on which they
have been trained.

The second interesting example is that of \emph{term weighting}. In
text classification, and also in other tasks such as text search and
text clustering, term weighting serves the purpose of emphasizing the
importance of terms that are deemed to be more important in describing
the semantics of the documents they occur in. While standard
unsupervised techniques were initially used~\cite{Yang94},
\emph{supervised term weighting} (STW) techniques later started to
gain prominence~\cite{SAC03b}, based on the notion that the terms
that should weigh more in representing a document are not the ones
that are rarest in the collection, but the ones which are most
correlated with the labels of interest. As in the case of DCCD, STW
techniques leverage the class labels of the training documents, and
generate document representations that have a special-purpose nature,
i.e., should be used only for the specific text classification task on
which they have been trained. The notion of STW is brought one step
further in \emph{learning to weight}~\cite{moreo2019learning}, where
the STW function is not given but is learnt from data.

% ----------------------------------------------------------------------

\subsection{Known Limitations}
\label{sec:limitations}

\noindent In this article we have focused our attention on multiclass
classification by topic.  Other classification scenarios remain
unexplored.  Two important such scenarios include (i) simple binary
classification, and (ii) classification by dimensions other than
topic, such as, e.g., sentiment classification (an active area of
research where deep learning is already showing interesting
results~\cite{zhang2018deep}).

In their current form, WCEs are not suitable for binary
classification. The reason is that, since the dimensionality of WCEs
is the number $m$ of classes in the codeframe of interest, in binary
classification the dimensionality of WCEs would be 1, which indicates
that WCEs would convey very little information to the classification
process.
% rely on their ability to scatter prior task-specific knowledge
% across the various classes.  \fsebcomment{Cercare una spiegazione un
% po' più convincente.}  The distributional representations it
% produces in binary contexts are thus inherently limited.
One possible strategy to counter this problem might be based on
increasing the number of classes artificially.  A possible approach to
do so might gain inspiration from the structural learning
framework~\cite{ando2005framework}.  This strategy would consist of
adding new classes that account for the presence or absence of certain
highly predictive terms for the task (i.e., adding to $\mathbf{Y}$ new
columns corresponding to binary versions of columns in $\mathbf{X}$
for highly predictive terms), and then computing the WCEs across them
(a similar intuition has been explored
in~\cite{moreo2016distributional}). 
%this is indeed the rationale of DCI, across pivots...
However, preliminary experiments we have conducted along this vein are
still inconclusive.
% did not show any significant improvement so far.

% General binary classification: idea of structural problems
% ~\cite{ando2005framework}, i.e., use highly predictive terms
% (according to, e.g., IG) as if they were classes to create the $Y$
% matrix (some preliminary experiments did not perform as expected).

% Word-Sentiment embeddings: What about sentiment polarity? this is a
% binary problem (might be ill-defined in our case, since one column
% might be redundant and thus the method is hardly justified).
% Possible ideas: use structural problems, use senti-wordnet as
% pre-trained sentiment priors, use sentiment embeddings as generated
% for other domains (idea of life-long learning for sentiment and
% multi-domain, possibly with many domains and PCA).

In addition to this, WCEs depend on the training set prevalences of
the classes in the codeframe. This might compromise their contribution
to tasks characterized by the presence of prior probability shift,
i.e., by the fact that the prevalence of a class in the training data
is substantially different from the prevalence of the same class in
the unseen data.  This might yield WCEs detrimental for tasks such as
text quantification~\cite{Gonzalez:2017it}, which indeed target
scenarios characterized by prior probability shift.
% This is to be confirmed.

% ----------------------------------------------------------------------

\section{Conclusions}
\label{sec:conclusions}

\noindent In this article we have presented word-class embeddings
(WCEs), i.e., distributed representations of words specifically
designed for multiclass text classification.  The hypothesis
underlying the present study is that the class-conditional
distributions of a term defines a fingerprint that might help to
refine its pre-trained representation for applications of multiclass
text classification.  The extensive empirical evaluation we have
conducted indeed confirms this hypothesis.

WCEs are easy and quick to compute.  Word-class embeddings are meant
to expand, and not modify, pre-trained word representations that model
general language usage.  Although this implies adding some parameters
to the model, we have seen that this additional cost is negligible in
practice; they contribute to the classification task far more than
they complicate the model.

We have investigated, also with the aid of a visualization tool, how
these new embeddings alter the topology of the embedding space when
WCEs are concatenated to embeddings pre-trained on generic, unlabelled
text corpora.  Our findings suggest that the new representation adds
global class structure while preserving local word semantics, and is
thus better suited for the classification task.

Unsupervised word embeddings are known to encode a mixture of the
senses of polysemous terms~\cite{camacho2018word}.  We think WCEs
indirectly help to disambiguate relevant domain-dependent terms.  The
word-class distribution might uncover the meaning of a term prevalent
in the domain of interest, thus leaving a domain-specific mark on the
resulting representation.  This information might be thought of as a
form of task-dependent word bias, which reframes the general-purpose
word meaning through the lens of the codeframe of interest.

% WCEs are easy to compute and, when concatenated with pre-trained
% embeddings, yield improvements in text classification performance
% across datasets and deep architectures.  The semantic modelled by
% WCEs is task specific.  It serves the purpose of modulating the
% semantic enclosed in pre-trained embeddings in a way that becomes
% better suited for text classification.  WCEs is the context-counting
% counterpart of \fasttext, just like PPMI stands to SGNS.

% "...the algorithm should find widespread use in TC and many areas of
% supervised learning"

% future
In future work we will try to provide solutions for the limitations
discussed in Section~\ref{sec:limitations}.  Other directions worth
exploring would be that of modelling, in sentiment classification
contexts, the sentiment prior of words across different domains (e.g.,
reviews of books, music, films, kitchen appliances, etc.), thus
producing word-sentiment embeddings with as many dimensions as there
are source domains available.  This would hopefully help in
cross-domain sentiment classification
tasks~\cite{blitzer2006domain,moreo2016distributional}.  It should
similarly be interesting to investigate the implications of this idea
on multi-task learning~\cite{caruana1997multitask}, where each task
might contribute with a dedicated task-specific embedding to the
representation for other tasks.
% Rich Caruana. 1993. Multitask learning: A knowledge-based source of
% inductive bias. In Proceedings of the Tenth International Conference
% on Machine Learning.  Semantic-embedding from WordNet and PCA of
% bag-of-synsets? This should help to improve the performance of word
% sense disambiguation (current state of the art is a bidirectional
% LSTM with \glove\ vectors -- this should be easy to test)

% ----------------------------------------------------

\section*{Acknowledgements}
  The present work has been supported by the \textsf{ARIADNEplus}
  project, funded by the European Commission (Grant 823914) under the
  H2020 Programme INFRAIA-2018-1. The authors' opinions do not
  necessarily reflect those of the European Commission.  Thanks to
  NVidia for donating the two Titan GPUs on which many of the
  experiments discussed in this paper were run.

% ----------------------------------------------------

\vskip 0.2in \bibliography{WCE}
\bibliographystyle{plain}

\end{document}